\documentclass[acmsmall]{acmart}

\AtBeginDocument{%
  \providecommand\BibTeX{{%
    \normalfont B\kern-0.5em{\scshape i\kern-0.25em b}\kern-0.8em\TeX}}}

\setcopyright{acmcopyright}

\acmJournal{TIST}

\usepackage{multirow}
\usepackage{array}
\usepackage{setspace}
\usepackage{url}
\usepackage{graphicx}
\begin{document}

\title{Conditional Text Generation for Harmonious Human-Machine Interaction}


\author{Bin Guo}
\email{guob@nwpu.edu.cn (Corresponding-author)}
\authornotemark[1]
\affiliation{%
  \institution{Northwestern Polytechnical University}
  \city{Xi'an}
  \country{P.R.China}
}

\author{Hao Wang}
\affiliation{%
  \institution{Northwestern Polytechnical University}
  \city{Xi'an}
  \country{P.R.China}
}

\author{Yasan Ding}
\affiliation{%
  \institution{Northwestern Polytechnical University}
  \city{Xi'an}
  \country{P.R.China}
}

\author{Wei Wu}
\email{wuwei@microsoft.com}
\affiliation{%
  \institution{Microsoft corporation}
}

\author{Shaoyang Hao}
\affiliation{%
 \institution{Northwestern Polytechnical University}
 \city{Xi'an}
 \country{P.R.China}}

\author{Yueqi Sun}
\affiliation{%
  \institution{Northwestern Polytechnical University}
  \city{Xi'an}
  \country{P.R.China}}

\author{Zhiwen Yu}
\affiliation{%
  \institution{Northwestern Polytechnical University}
  \city{Xi'an}
  \country{P.R.China}}

\renewcommand{\shortauthors}{B.Guo, et al.}

\begin{abstract}
In recent years, with the development of deep learning, text generation technology has undergone great changes and provided many kinds of services for human beings, such as restaurant reservation and daily communication. The automatically generated text is becoming more and more fluent so researchers begin to consider more anthropomorphic text generation technology, that is the conditional text generation, including emotional text generation, personalized text generation, and so on. Conditional Text Generation (CTG) has thus become a research hotspot. As a promising research field, we find that many efforts have been paid to exploring it. Therefore, we aim to give a comprehensive review of the new research trends of CTG. We first summary several key techniques and illustrate the technical evolution route in the field of neural text generation, based on the concept model of CTG. We further make an investigation of existing CTG fields and propose several general learning models for CTG. Finally, we discuss the open issues and promising research directions of CTG.
\end{abstract}



\maketitle

\section{Introduction}

Jorge Luis Borges\footnote{\url{https://www.goodreads.com/author/show/500.Jorge_Luis_Borges}} once described a magic Library, named ``The Library of Babel''\footnote{\url{http://www.paulrittman.com/JLBLibraryofBabel.pdf}}, where everyone could find any book he wanted. The readers cannot help but wonder who wrote these books. Are they all written by human writers? Absolutely, the answer is no. This library seems unlikely to exist, however, the development of text generation technology in recent years has made it possible. For instance, Philip M. Parker, having written and sold more than 100,000 books on Amazon, utilizes computer programs to collect massive publicly information on the Internet for automatic compilation into books\footnote{\url{https://www.kurzweilai.net/he-wrote-200-000-books-but-computers-did-some-of-the-work}}. The above scene of Parker belongs to the \textit{text-to-text} generation \cite{genest2011framework}, which takes existing textual materials as input and automatically generates the new text.

Text-to-text generation is a typical subfield of \textit{text generation} \cite{mckeown1992text}, which uses diverse types of information to enable computers to learn to express like human, including image, text and so on. According to different data sources, text generation can be divided into \textit{data-to-text}, \textit{text-to-text}, and \textit{image-to-text} generation. News generation is a typical application of data-to-text generation. For example, there was an earthquake in California on March 17, 2014, and the Los Angeles Times firstly provided detailed information about the time, location and magnitude of the quake. Actually that news article was automatically generated by a `robot reporter', which converted the incoming registered seismic data into text by filling slots in predefined templates \cite{oremus2014first}. The data-to-text generation technology fills the established template with structured data and generates the output text containing all key components, which has exerted considerable influence in the field of news media.

The application of text generation from text to text includes machine translation \cite{cho2014learning}, dialogue system \cite{shang2015neural}, text summarization \cite{rush2017neural}, reading comprehension \cite{hermann2015teaching}, etc. By understanding the original text and obtaining its semantic representation, natural language text is generated for communicating, summarizing or refining. Besides, The application fields from image to text generation include image captioning \cite{mao2014explain}, visual question answering \cite{antol2015vqa}, etc. By processing image information, the contents contained in the image can be understood to generate corresponding natural language descriptions and answers.

Deep learning contributes to the most recent advances in the text generation field. Specifically, with the help of the recurrent neural networks (RNN) \cite{elman1990finding}, attention mechanism, generative adversarial networks (GAN) \cite{goodfellow2014generative}, reinforcement learning (RL), Variational Autoencoder (VAE) \cite{kingma2013auto} and Transformer \cite{vaswani2017attention}, the generated text becomes more coherent, logical and emotionally harmonious, which is more suitable for offering assistance in every aspect of people's lives. For example, the dialogue systems, such as Microsoft XiaoIce\footnote{\url{http://www.msxiaobing.com/}}, Cortana\footnote{\url{http://www.msxiaona.cn/}} and Apple Siri\footnote{\url{https://www.apple.com/siri/}}, can not only chat with us, but also assist us to operate electronic devices. News-writing-robots have provided creative assistance for journalists, and the machine translation technology has effectively met our needs of translation.

Advancements in universal text generation technology prompt researchers to explore more anthropopathic text generation methods, such as context-based text generation \cite{jaech2018low}, personalized text generation \cite{luo2019learning}, topic-aware text generation \cite{wang2018reinforced}, emotional text generation \cite{kong2019adversarial}, knowledge-enhanced text generation \cite{young2018augmenting} and visual text generation \cite{dai2017towards}. Obviously, applying additional information in text generation may make the generated text more personified and facilitate harmonious human-machine interaction. However, new challenges are raised, summarized as follows.
\begin{itemize}
\item How to efficiently integrate the additional conditional information with traditional model structures is a big challenge.
\item Due to the scarcity of text datasets with specific conditions, training the conditional text generation models become more difficult.
\item There is no reasonable evaluation metrics of the conditional text generation, making it difficult to quantify the performance of models.
\end{itemize}

This paper aims to give an in-depth survey of the development of neural text generation models. Specifically, we mainly focus on various studies on \textit{conditional text generation} (CTG), such as context-based text generation, topic-aware text generation, and knowledge-enhanced text generation. Compared with general text generation, the conditional text generation is more in line with the needs of precision and friendly services.

To sum up, we summarize the contributions of our work as follows.
\begin{itemize}
\item Based on a brief review of current text generation techniques, we characterize the concept model of CTG and present the major human-centric services.
\item We make an investigation of several different CTG fields, including context-based text generation, personalized text generation, topic-aware text generation, emotional text generation, knowledge-enhanced text generation, visual text generation, multi-conditional text generation and pre-trained language model-based text generation. Besides, the evaluation methods and conditional datasets are also discussed. Based on the summary of existing researches, we propose several general learning models for CTG.
\item We further discuss some promising research directions of CTG, including the consideration of different types of contexts, the multi-modal data translation, lifelong learning, and so on.
\end{itemize}

The remainder of this paper is organized as follows. In Section 2, we give a brief review of key text generation techniques. In Section 3, we characterize the concept model of CTG. We then summarize the major researches of CTG in Section 4 and propose several general learning models for CTG in section 5, followed by the open issues and future research directions in Section 6. Finally, we conclude this paper in Section 7.

\section{THE KEY TECHNIQUES OF NEURAL TEXT GENERATION}
The past few decades have witnessed a huge leap forward in the text generation technology. Specifically, from the original rule-based and statistical methods to the end-to-end neural network-based methods, the overall quality of generated content is further improved. This section gives a brief review of key techniques of neural text generation.

DNN-based methods do not require manual feature extraction, and can automatically learn the continuous vector representations in three steps for the task-specific knowledge in different tasks, i.e. encoding, reasoning and decoding successively \cite{gao2019neural}. The inputs of neural network models will be firstly encoded into a vector space, where semantically related or similar concepts are close to each other. Afterwards the neural networks will reason in the vector space according to the hidden state and current input to produce the system response. Finally, the system response will be decoded to generate natural language text. Different neural network structures are usually adopted in the stages of encoding, reasoning and decoding, and all the parameters are optimized through back-propagation and gradient descent algorithms. The end-to-end learning mechanism promotes neural networks to fully mine the correlation in the data, and alleviates the requirement of characteristic engineering greatly.

The key techniques of neural text generation mainly include RNN, GAN, RL, VAE, and Transformer, which will be summarized in the following subsections. Natural language text is a typical kind of sequential data with specific relationships between contexts, and the natural sequential structure of RNN is very suitable for modeling text data. Since RNN contain internal memory, which can remember previous inputs and the current input, it makes sequence modeling much easier. The output at the current time step depends not only on the instantaneous input, but also on outputs of previous time steps, which makes it highly capable of capturing contextual information and generating sentences that satisfy syntactic structures. However, the word-by-word sequential generation process of RNN cannot learn representations of full sentences, making it tend to generate inconsistent and uninformative text because of the absence of global features like topics or high-level syntactic features. With latent variables in continuous space, VAE can capture implicit language structure and utterance-level semantics (\textit{e.g.}, topics, syntactic properties), which are served as the global representations during the decoding process. By the sampling procedure of the latent variables, VAE is capable of producing more natural, meaningful, and diversified natural language texts.

The RNN-based text generation models trained by maximizing the log-likelihood objective function are prone to the problem of exposure bias, caused by the inconsistency of the sequence input during training and testing. Therefore, GAN, another powerful deep generative model that has become a huge success in computer vision, is introduced to text generation which uses adversarial training to replace the maximum likelihood training to simulate the real data distribution and generate higher-quality text. Through the adversarial training of the generator and discriminator, the generator of GAN gains the ability to generate almost real data. However, the original GAN is only suitable for processing continuous data such as images, while text is a typical kind of discrete data, so it cannot be applied directly to text generation. Many efforts have been made to adjust the internal calculation mechanism of GAN to deal with this problem, among which the introduction of RL greatly promotes the application of GAN in text generation. By combining the reward mechanism and the policy gradient technology of RL, GAN skillfully avoids the above problem that gradient cannot back propagation when facing discrete data and achieves promising results.

VAE and GAN are the two most powerful deep generative models which can generate data with complex distribution approximate to the real data distribution from random noise with simple distribution. The distinct difference between them is the distribution metric, that is, the loss function used to measure the quality of the generated data, is different. VAE utilizes an explicit measurement method that measures the KL divergence of training data and noise by assuming that training data is generated by another distribution. GAN, on the other hand, avoids the explicit measurement of distribution difference by making the neural network learning the measurement through adversarial training. In view of the two distribution measurement criteria are not perfect measurements, VAE and GAN have their own problems to be solved.

Faced with other problems of RNN, including the inability to effectively capture long-term dependencies, the vulnerability to the problem of gradient vanishing or exploding, and the lack of parallel computing capability, the Transformer model is proposed which adopts self-attention mechanism to replace the sequential structure in RNN. The self-attention mechanism can capture the context dependency among all words in a sequence to achieve more efficient sequence modeling without distance restrictions and obtain more semantically-rich text representations. Transformer has shown excellent performance in various NLP tasks since it was proposed and has great development potential.

In summary, these key techniques' advantages and disadvantages are compared in Table~\ref{tab:technologies}.

\begin{table}
 \caption{A summary of text generation techniques}
 \label{tab:technologies}
 \begin{tabular}{|m{2.5cm}<{\centering}|m{5cm}<{\centering}|m{5cm}<{\centering}|}
    \hline
    \textbf{Technique} & \textbf{Advantages} & \textbf{Disadvantages}\\
    \hline
    RNN & Natural sequence structure is very suitable for the task of sequence modeling & Cannot effectively capture the long-distance dependence between sentences\\
    \hline
    GAN & Unsupervised learning; Generating clearer and more realistic samples than other generative models & Instable training process; Not suitable for processing discrete data, such as text\\
    \hline
    Reinforcement learning & Similar to human learning manners; Combining with GAN can subtly solve the existing problems in GAN and generate realistic text & Quite complicated training process\\
    \hline
    VAE & Leveraging the latent vectors to increase the diversity of the generated text & The latent variable ensures that the desired content is generated, regardless of its quality\\
    \hline
    Transformer & The attention mechanism can efficiently capture the long-term context information; Fast parallel computing speed & Large amount of calculation and slow training speed\\
    \hline
  \end{tabular}
 \end{table}

\subsection{RNN}
RNN is one of the most commonly used neural network models in text generation, whose natural sequence structure is suitable for the task of text sequence modeling. The recurrent structure in RNN determines that it can process textual data sequentially, of which each hidden state takes the current input and the previous hidden state into consideration. Given the input text sequence $X=(x_{1},x_{2},...,x_{n})$, the hidden state $s_{t}$ of time step ${t}$ is calculated as follows:

\begin{equation}
s_{t} = f(U\cdot x_{t} + W\cdot s_{t-1})
\end{equation}

After sequential processing, all semantic information of the given text are compressed into a fixed-length vector, the hidden state vector at the last time step, enabling the RNN model to have memory of previous content. Nevertheless, the problems of gradient vanishing or exploding still limit the application prospect of RNN. Variants of RNN model, such as long short-term memory (LSTM) and gated recurrent unit (GRU), combine the short-time and long-time memory through uniquely designed gating mechanisms, which makes them effectively solving these problems. There are three gate structures in LSTM that control the information in the cell state and selectively determine whether the information is retained or not. The forget gate determines what information in the cell state needs to be discarded according to the current input and the previous hidden state, which is calculated as follows:

\begin{equation}
f_{t} = \sigma (W_{f}\cdot [h_{t-1},x_{t}] + b_{f})
\end{equation}

The input gate determines how much new information is added to the cell state, which is calculated as follows:

\begin{equation}
i_{t} = \sigma (W_{i}\cdot [h_{t-1},x_{t}]+b_{i})
\end{equation}
\begin{equation}
\tilde{C_{t}} = tanh(W_{c}\cdot [h_{t-1},x_{t}]+b_{c})
\end{equation}

Then the cell state is updated based on the input gate and the forget gate as follows:

\begin{equation}
C_{t} = f_{t}\cdot C_{t-1}+i_{t}\cdot \tilde{C_{t}}
\end{equation}

Finally, the output gate determines the output of the LSTM unit at time step ${t}$ according to the cell state.
\begin{equation}
o_{t} = \sigma (W_{o}\cdot [h_{t-1},x_{t}]+b_{o})
\end{equation}
\begin{equation}
h_{t} = o_{t}\cdot tanh(C_{t})
\end{equation}

Numerous researchers have shown that LSTM has the ability to generate natural and realistic texts in many generation tasks. Sutskever \textit{et al.} \cite{sutskever2014sequence} firstly propose the Sequence-to-Sequence (Seq2seq) learning model, which is a generalized framework for converting one sequence to another. In this framework, a LSTM as the encoder compresses sequences into vector representations. Then another LSTM as the decoder predicts output words one by one conditioned on the hidden state, and it takes the previous output as the input to predict the next output. Since this framework has no limitation of the length of input/output sequences, it has been widely used in text generation, including machine translation \cite{cho2014learning}, text summarization \cite{rush2017neural}, and dialogue system \cite{vinyals2015neural}. Consequently, the data-driven end-to-end training based on the Seq2seq model has become the mainstream method in text generation.

Actually, traditional Seq2seq models generally have two problems. The first is that all the inputs are transformed into a vector with fixed length, which limits the ability of latent vectors to represent input information, and the second is that assigning all the input words with the same weight cannot effectively capture the key information. To solve these problems, the Attention mechanism, an widely utilized mechanism in computer vision, is introduced into NLP. Through assigning different weights to different parts of the input sequence according to the current decoding state, the Attention mechanism can extract the key components from the input, which helps generation models make more accurate judgments while reducing the computation and storage consumption. The Attention mechanism is firstly applied to the Seq2seq model to fulfill machine translation tasks \cite{bahdanau2015neural}, and now has gradually become an important part of text generation models. For example, Xing \textit{et al.} \cite{xing2018hierarchical} introduce an attention-based multi-turn response generation model to capture the most relevant content in the conversation context. The Attention mechanism makes the multi-turn dialogue more coherent and consistent.

\subsection{GAN and RL}
From the perspective of neural network optimizing, there are still some defects in RNN-based generation models. First, most RNN-based text generation models are trained by maximizing the log-likelihood objective function, which may lead to the problem of exposure bias. Second, most loss functions are calculated at the level of words, while most evaluation metrics are based on the level of sentences, which may result in the inconsistency between the optimization direction of the model and the actual requirements. In this case, researchers introduce the GAN \cite{goodfellow2014generative} into the study of text generation. GAN is composed of two parts: the generator and the discriminator. The generator produces false sample distributions similar to the real data, and the discriminator distinguishes generated samples and real samples as accurately as possible. For example, Zhang \textit{et al.} \cite{zhang2016generating} attempt to combine the LSTM and convolutional neural network (CNN) to generate realistic text using the idea of adversarial training.

However, the original GAN is only applicable to generate continuous data and has poor performance on processing discrete data because it is difficult for the gradient of the discriminator to correctly back-propagate through discrete variables. In order to solve this problem, Zhang \textit{et al.} \cite{zhang2016generating} utilize the smooth approximation algorithm to approximate the output of generator. Instead of utilizing the standard objective function of GAN, they match the feature distribution and make the word predictions `soft' in the embedding vector space to generate high-quality sentences. Researchers have also made some fine-tuning to GAN's structure to generate discrete data, e.g., the Wasserstein GAN model \cite{arjovsky2017wasserstein}.

Although the direct improvement of GAN has achieved some progress, it is still far from meeting practical requirements. Therefore, the idea of RL begins to be introduced to text generation. RL is usually a Markov decision process in which the action in each state will be rewarded (or reversely rewarded--punishment). For maximizing the expected rewards, the RL machine tries various possible actions in different states to evaluate the optimal policy according to the rewards provided by the environment. It can be seen that the reward mechanism in RL could help the GAN to deal with discrete data, which provides a new possibility for the application of GAN in text generation. For example, Yu \textit{et al.} \cite{yu2017seqgan} propose the \textit{SeqGAN} model to solve the problems of GAN in generating discrete text data. SeqGAN regards the text generation as a sequence decision procedure in RL, in which the generated sequence at each timestep represents the current state, the next word to be generated is regarded as the action to be taken and the returned reward is the discriminator's score of the generated sequence. Through gradient policy algorithms, the SeqGAN model directly avoids the differentiability problem in the generator and obtain remarkable results in generating realistic natural language text.

\subsection{VAE}
Although the traditional Seq2seq model has made great progress in text generation, it tends to produce general and safe sentences with high probability, such as `I do not know' and `I am sorry'. At the same time, the training of text generation system needs a large amount of high-quality labelled data, namely supervised training. However, in reality, most of the data is unlabelled and labelling a large amount of data is very time-consuming. The idea of unsupervised learning is introduced to solve this problem. VAE is a powerful unsupervised generative model, which contains an encoder that encodes input data into latent variables, and a decoder that decodes latent variables to reconstruct the original input data. Given the input $x$, the encoder will encode it into latent space $p_{\theta}(z|x)$, where $\theta$ is the parameters of encoder. The decoder does the opposite which finds the probability distribution of the input based on the hidden variable $p_\phi(x|z)$, where $\phi$ is the parameters of decoder. There is usually a latent hierarchical structure in natural language, and latent variables in VAE can capture and reflect such inherent language structure, so as to produce more natural and expressive natural language text.

The work of Bowman \textit{et al.} \cite{bowman2016generating} introduce a RNN-based VAE text generation model which assigns whole sentences with distributed latent vectors. By appending Gaussian prior distribution regularization on the hidden layers of the encoder, a sequence autoencoder model is constructed and the output sentence is generated word by word conditioned on the hidden vector to obtain consistency and diversity. Sequential data usually shows hierarchical structures and complicated dependencies between sub-sequences. For example, the sentence sequences and word sequences in a multi-round conversation have massive dependencies. Serban \textit{et al.} \cite{serban2017hierarchical} attach the latent variable to the hierarchical dialogue model to assign the generative model with multiple levels of variability for meaningful and diverse responses. Specifically They attach a high-dimensional latent variable to each sentence in the dialogue history, followed by generating responses conditioned on the latent variable.

\subsection{Transformer}
From the perspective of neural network training, RNN-based generation models also have some obvious shortcomings. Firstly, RNN processes the input sequence with strict linear order from forward to back, which leads to the problem of gradient vanishing or exploding due to the long back propagation path. Secondly, RNN lacks efficient parallel computing capability due to its linear propagation structure where the calculation at the next time step relies on the outputs at the previous time steps. Therefore, RNN faces the issue of low calculation efficiency in large-scale application scenarios. To address this problem, Google proposes a new sequence modeling model, the Transformer model \cite{vaswani2017attention}, which abandons the sequence structure in RNN and just contains Attention modules.

Specifically, the Transformer model is an encoder-decoder structure, only consisting of Attention modules and feed forward neural networks. The self-attention mechanism is the core of Transformer that captures the dependency among words in a sequence to obtain better semantic representations of each word. The self attention module is calculated as follows:

\begin{equation}
Attention(Q,K,V)=softmax(\frac{QK^{T}}{\sqrt{d_{k}}})V
\end{equation}

The $Q$, $K$ and $V$ are all vectors, each of which is obtained by multiplying the input vector by the weight matrix. The multi-head attention mechanism, composed of many self-attention modules to form an attention module, is proposed to further improve the ability of capturing context semantic information. After encoded by the self attention module, the output vector will be sent to the feed forward neural network, calculated as follows:

\begin{equation}
FFN(z) = max(0,ZW_{1}+b_{1})W_{2}+b_{2}
\end{equation}

Besides the self attention module and the feed forward neural network module, there is also an encoder-decoder attention module in the decoder, which has the same mechanism as the traditional Seq2seq model. Due to the parallelization of the Attention module, Transformer has powerful parallel computing capacity and broad application prospect. Since Transformer was put forward, the various models based on it have achieved excellent performance in various NLP tasks, such as BERT\cite{kenton2019bert} and GPT\cite{radford2018improving}, making significant impact on the whole research area of NLP.

\section{CONDITIONAL TEXT GENERATION}
The development of deep neural networks brings unprecedented progress to text generation. However, there are still some problems with the existing text generation technology. For example, many studies train the text generation model only based on the content of input text, ignoring many other factors. However, a real person not only considers the context, but also adjusts the content according to their own conditions (such as mood and gender) and external factors (such as weather and environment) when speaking or writing. In this paper, we take conditional text generation (CTG) as the future research direction which is the key factor to improve the quality of generated text. Specifically, it includes context-based text generation, personalized text generation, topic-aware text generation, emotional text generation, knowledge-enhanced text generation, and visual text generation. In this section, we formalize the definition of CTG and introduce the wide application fields of it.

\subsection{The Concept Model}
The CTG refers to taking certain external conditions into consideration to influence the generated results in the process of text generation. These conditions usually include \textit{context}, \textit{topic}, \textit{emotion}, \textit{external knowledge}, and so on. The general text generation methods only consider the text content factor, which makes the generated text less diverse and has a large gap with human expression. Consideration of external conditions in text generation makes it more anthropomorphic and brings better services to human beings in various fields.

We first give a formal definition of general text generation. Given the input text sequence $X=(x_{1},x_{2},...,x_{S})$, the target of general text generation model is to generate the output text sequence $Y=(y_{1},y_{2},...,y_{T})$, where $S$ and $T$ are the length of the input and output sequence respectively. The general text generation model can be defined as follows:

\begin{equation}
p(Y|X) = \prod_{t=1}^{T}p(y_{t}|X,y_{<t})
\end{equation}1

At each decoding time step, the decoder will combine the input text and the output in the previous time steps to generate the current result. Finally, a text sequence with the highest probability is generated. For example, in the machine translation system, $X$ might be a source Chinese sentence and $Y$ might be an target English sentence, while in the dialogue system, $X$ might be the query and $Y$ might be the response, without considering other additional information in the generation process. On the basis of general text generation, the CTG models fuse additional condition to generate more anthropomorphic text. We define various kinds of CTG fields as follows:

\begin{definition}
CONTEXT-BASED TEXT GENERATION: \textit{Integrating contextual information during text generation to enhance the understanding of environmental state to produce more coherent and informative text content.}
\end{definition}

The contexts of natural language refer to the situations they are generated which are the key factor to ensure consistency and smoothness of the generated text. Given a set of contexts $C = \{c_{i}\}_{i=1,2,...,K_{c}}$, each context $c_{i}$ may be a text sequence, a simple word, a sentiment score, and so on, and $K_{c}$ is the number of context types. The context-based text generation model can be defined as follows:

\begin{equation}
p(Y|X,C) = \prod_{t=1}^{T}p(y_{t}|X,C,y_{<t})
\end{equation}

Take human conversation for example, $C$ refers to the historical dialogue content in multi-rounds conversation. The daily dialogue process of human beings usually lasts for several rounds. The historical dialogue content in the multi-rounds dialogue is one kind of context information, and we will generate responses based on the historical dialogue to keep the conversation consistent. While in the advertisements writing scenario, $C$ refers to various types of information about specific product, such as its brand, function, price, etc., and may also be information about user preferences. Only by integrating multi-kinds of contextual information can more appealing advertisements be generated.

\begin{definition}
PERSONALIZED TEXT GENERATION: \textit{Assigning specific personalization characteristics to the text generation agents to produce personalized text contents which fit the given personalization characteristics.}
\end{definition}

Personalization means that everyone has characteristics different from others, which will subtly influence how and what we express ourselves. Given a set of personalized characteristics $S = \{s_{i}\}_{i=1,2,...,K_{s}}$, each $s_{i}$ represents age, gender, profession, or other characteristics, and $K_{s}$ is the number of personalized characteristics types. The personalized text generation model can be defined as follows:

\begin{equation}
p(Y|X,S) = \prod_{t=1}^{T}p(y_{t}|X,S,y_{<t})
\end{equation}

Similarly, take dialogue as an example. People of different genders and ages have different views on the same thing, so the personalized characteristics $S$ will have an impact on the dialogue content. When writing commodity description advertisements, it is necessary to combine the personalized features $S$ of users, such as age, gender and shopping preference, so as to generate descriptions more in line with users' expectations. Therefore, in order to make the text generation agent more personified, it is necessary to assign specific personalized characteristics to generate text content conforming to the personalized information.

\begin{definition}
TOPIC-AWARE TEXT GENERATION: \textit{Incorporating a specific topic in the process of text generation to make the whole text content suitable for the topic and ensure the coherence and rationality of the generated text.}
\end{definition}

Natural language text has very strong internal relevance, especially in long text. A piece of text usually aligns around a specific topic, so considering topic information can generate more coherent and meaningful text. Given a set of topic words $T = \{t_{i}\}_{i=1,2,...,K_{t}}$, each $t_{i}$ is a topic word and $K_{t}$ is the number of topic words. The topic-aware text generation model can be defined as follows:

\begin{equation}
p(Y|X,T) = \prod_{t=1}^{T}p(y_{t}|X,T,y_{<t})
\end{equation}

When we write an article, we usually expand our thinking according to a specific topic $T$, such as maternal love, to ensure the logical consistency and consistency of the whole article. In the process of foreign language translation, it is necessary to combine the topic of the whole text content, to get a more fluent and consistent translation around the central topic. Combining topic information is a key factor to ensure logical coherence and compact semantics in text generation.

\begin{definition}
EMOTIONAL TEXT GENERATION: \textit{Embodying the emotional expressions of the agents in the process of text generation, such as positive or negative, happy or sad, to adjust the content and expression style of the generated text.}
\end{definition}

Emotion is a very important attribute of natural language, and people usually have certain emotions in daily communication or writing. Different emotions have important effects on what is being expressed. For example, when we are angry, we usually say something that is not rational or hurts others. Incorporating emotion into text generation can make the generated content more personified. In dialogue systems, it has a direct and quantifiable impact on product usability and user satisfaction when considering specific emotions. Given a specific emotion category $E$, which may be anger, sadness, joy, and so on, the emotional text generation model can be defined as follows:

\begin{equation}
p(Y|X,E) = \prod_{t=1}^{T}p(y_{t}|X,E,y_{<t})
\end{equation}

\begin{definition}
KNOWLEDGE-ENHANCED TEXT GENERATION: \textit{Embracing external knowledge, such as search engine or knowledge base to provide factual basis and reference of the generated content in the text generation procedure.}
\end{definition}
Human has a wealth of prior knowledge, and can flexibly combine our own knowledge in communication, translation or writing to express ourselves. Combining external knowledge in the text generation system can make the generated text more informative, more consistent with the logic of human expression, and reduce the possibility of common sense mistakes. Given a set of knowledge facts $F = \{f_{i}\}_{i=1,2,...,K_{f}}$, each $f_{i}$ is a text sequence (also called free-text knowledge) or a knowledge triple from knowledge graph (also called structured knowledge), and $K_{f}$ is the number of knowledge facts. The knowledge-enhanced text generation model can be defined as follows:

\begin{equation}
p(Y|X,F) = \prod_{t=1}^{T}p(y_{t}|X,F,y_{<t})
\end{equation}

When we are asked a specific question, we will combine our knowledge to understand and reason the question, and find the corresponding answer to reply. For example, the question ``What is the capital of China?'' can be answered by combing geographical knowledge ``Beijing is the capital of China''. The combination of knowledge is the key factor to ensure the real humanization of text generation system. Only through the interaction and expression of knowledge with the real world, can text generation agent be truly integrated into our daily life.

\begin{definition}
VISUAL TEXT GENERATION: \textit{Integrating the semantic information in images into generated text, such as generating text descriptions according to image contents, or answering questions about given images.}
\end{definition}

Data in our life is multi-modal, including not only text, but also images, sounds, and so on. We can automatically extract the information contained in the image and translate it into understandable natural language. Images can vividly depict external events and our psychological activities, so visual text generation has a rich application prospect. Given an image $I$, the visual text generation model can be defined as follows:

\begin{equation}
p(Y|X,I) = \prod_{t=1}^{T}p(y_{t}|X,I,y_{<t})
\end{equation}

\begin{figure}[htbp]
    \centering
    \includegraphics[scale=0.4]{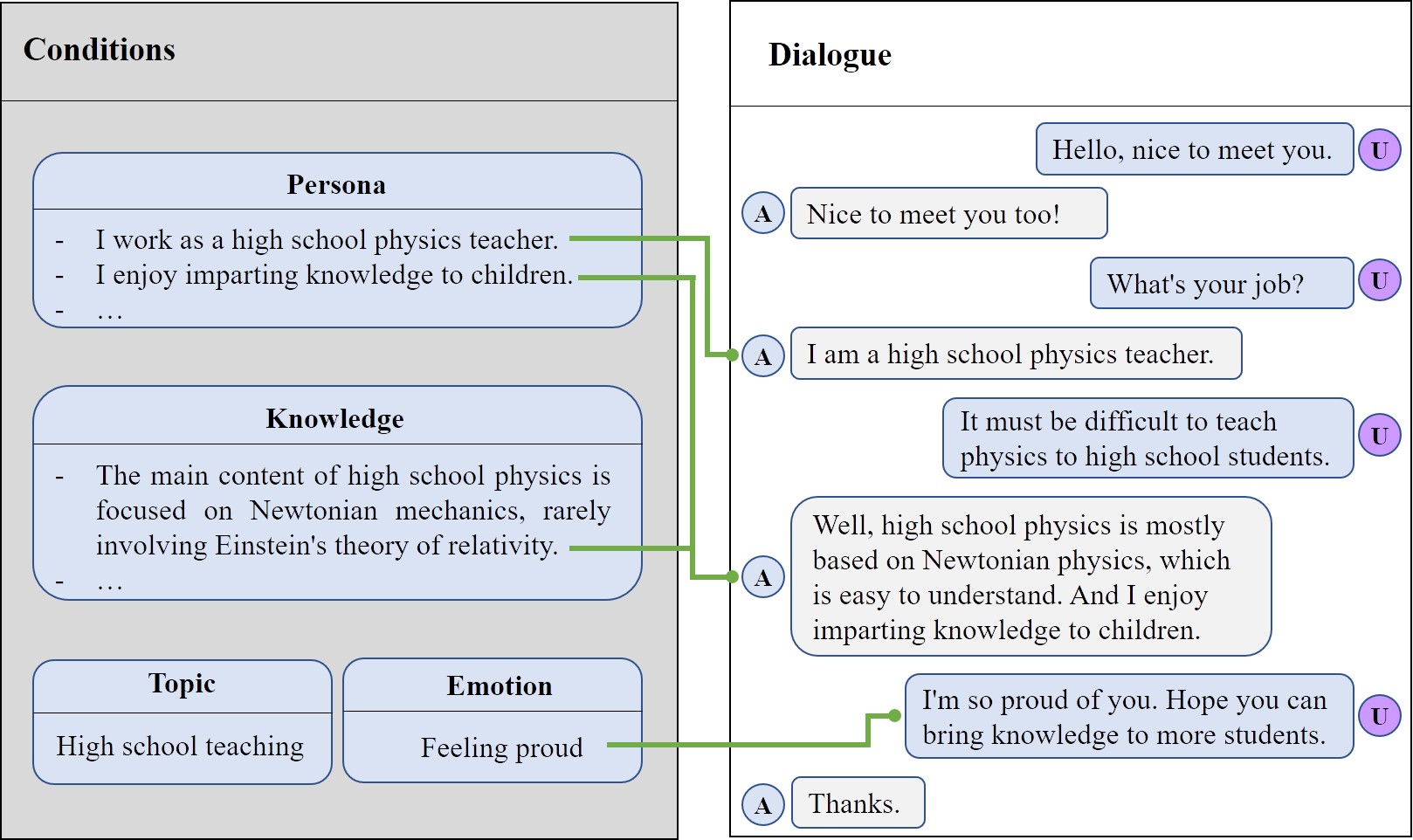}
    \caption{The example of conditional dialogue system}
    \label{img:example}
\end{figure}

Having formalized the various kinds of CTG, a real-life actual example, the dialogue system incorporating several conditions, is shown in figure~\ref{img:example}. In this example, different conditions are considered to enhance the performance of the dialogue agent. The personalized characteristics $S$ are two descriptive sentences about the work and preferences of the agent, and the knowledge facts $F$ is a text paragraph about the content of physics in high school, while the context $C$ is the historical dialogue. The whole topic $T$ of this conversation is ``High school teaching'', and the emotion $E$ inside the dialogue is ``Feeling proud''. Based on these conditions, the dialogue agent can generate more relevant, personalized, substantial, and context-consistent responses.

We will discuss the technical details of implementing various kinds of CTG fields in the following sections. At present, most text generation models are based on the encoder-decoder structure, in which the encoder transforms input sequences into semantic vector representations, and the decoder generates outputs according to the input information, such as dialogue responses, and product reviews. Applying constraints of conditions in different parts of text generation models, including encoders, decoders, and their interaction modes, has been widely studied. In the encoding stage, external conditions can be encoded by various techniques, such as RNN and Transformer, and served as the input of the decoder together with the original input to control the generation process. The decoder can be modified by weighted decoding \cite{ghazvininejad2017hafez} or other technologies to control the decoding procedure to increase or decrease the probability of words with certain conditions. Meanwhile, the attention mechanism or RL can enhance the interaction mode of the encoder and decoder, to mine the implicit and deep semantic information in conditions. In short, only considering from multiple aspects including the encoder and decoder to integrate different conditions, can CTG systems produce content with higher quality and personification to provide us with more comfortably services.

\subsection{Text Generation-based Human-Centric Services}
Text generation technology has a wide range of application scenarios in daily life. It is an ongoing effort of the academic/industry researchers to use various text generation technologies to provide human-centric services, presented as follows.

\textbf{\textit{Goal-oriented dialog systems.}} The dialogue systems are the most typical applications of text generation, which can be divided into goal-oriented and non-goal-oriented systems. Goal-oriented dialogue systems assist human to fulfil various tasks to reduce our operational burden, such as restaurant reservation, and travel time arrangement. In addition, goal-oriented dialog systems can also help companies accomplish specific businesses, such as customer transactions in a bank. According to Lauren Foye\footnote{\url{https://www.juniperresearch.com/press/press-releases/chatbots-a-game-changer-for-banking-healthcare}}, banks can automate up to 90\% of their customer interaction using dialog systems by 2022. Apple Siri\footnote{\url{https://www.apple.com/siri/}}, Microsoft Cortana\footnote{\url{https://www.microsoft.com/en-us/cortana/}}, Google Assistant\footnote{\url{https://assistant.google.com/}}, and Amazon Alexa\footnote{\url{https://developer.amazon.com/en-US/alexa/}} are all typically frequently-used goal-oriented dialogue systems in our daily life. Siri is the first virtual assistant with a voice deployed in Apple devices to assist human to operate smartphones, which was born of this desire to make our interactions with computers more human-like, while Microsoft Cortana is the virtual assistant created by Microsoft for Windows 10, Windows Mobile and all of Microsoft's integrated hardware. Google assistant is a smart personal assistant similar to Siri but can deploy on a wide range of devices, including android phones, android TV, wearable devices, etc. Take Siri as an example, we will introduce the detailed technical workflow of it.

\begin{figure}[htbp]
    \centering
    \includegraphics[scale=0.4]{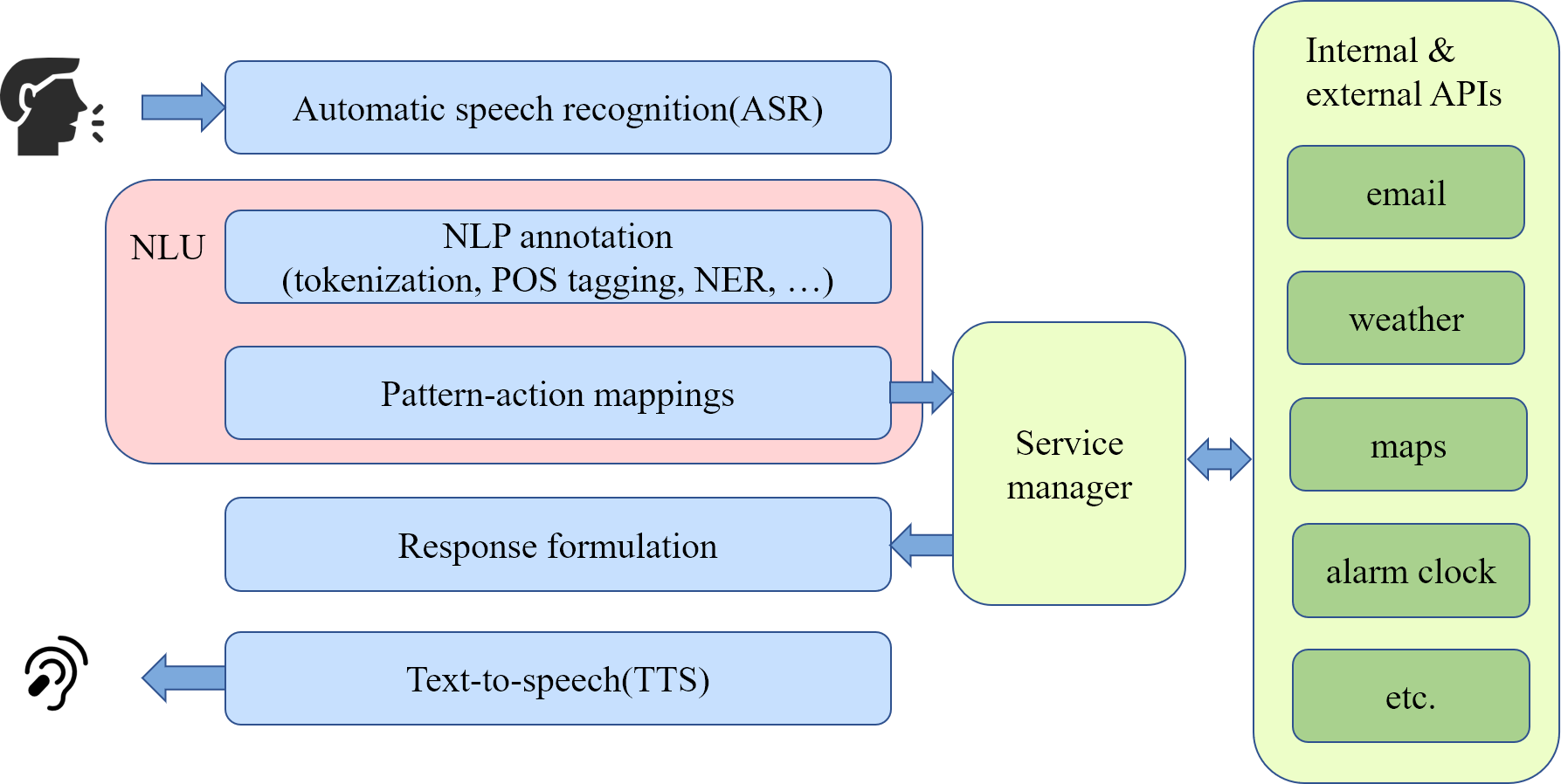}
    \caption{The overall working flow chart of Siri}
    \label{img:siri}
\end{figure}

On all IOS devices, we can say ``Hey Siri'' to invoke Siri hands-free. A very small speech recognizer runs all the time and listens for just those two words. When it detects ``Hey Siri'', the rest of Siri parses the following speech as a command or query. The overall working flow chart of Siri is shown in the figure~\ref{img:siri}.

The input command of Siri passes through four stages for processing altogether. The first stage is speech recognition, where the ``Hey Siri'' detector uses a DNN to convert the acoustic pattern of our voice at each instant into a probability distribution over speech sounds. It then uses a temporal integration process to compute a confidence score that the phrase we uttered is ``Hey Siri''. If the score is high enough, Siri wakes up. After having collected and subsequently converted our command into a file, Siri sends it to Apple servers for processing. Once in the Apple servers, our spoken words undergo different flowchart branches to arrive at a possible solution. The third stage is to understand the meaning of the command using NLP technologies. Apple servers run NLP algorithms such as tokenization and named entity recognition, on input text to understand the intent of what the user is trying to say. For instance, the NLP engines can differentiate that when a user is saying ``set an alarm for 7AM tomorrow'', the user is asking about setting an alarm and not about making a call. Finally, Siri communicates with other apps on the phone to provide the desired deliverable response to us in voice.

\textbf{\textit{Chatbots.}} The non-goal-oriented dialogue systems, also known as chatbots, can communicate with humans normally in the open domain. Instead of completing specific tasks, chatbots engage in chatty conversations with humans and perform like a real person as much as possible. Chatbots provide us with a realistic and interactive dialogue experience and establish certain emotional connections with us. In recent years, with the emergence of a large amount of dialogue data and the breakthrough of machine learning applied to dialogue AI, intelligent dialogue systems have achieved gratifying results in the academia and industry. Microsoft XiaoIce\footnote{\url{http://www.msxiaobing.com/}} is one of the most popular social chatbots in the world that has made conservations with hundreds of millions of users and successfully built long-time emotional connections with them. Zhou \textit{et al.} \cite{zhou2018design} describe the development of the Microsoft XiaoIce system to provide some guidance for chatbot researchers, which will be briefly summarized below.

\begin{figure}[htbp]
    \centering
    \includegraphics[scale=0.4]{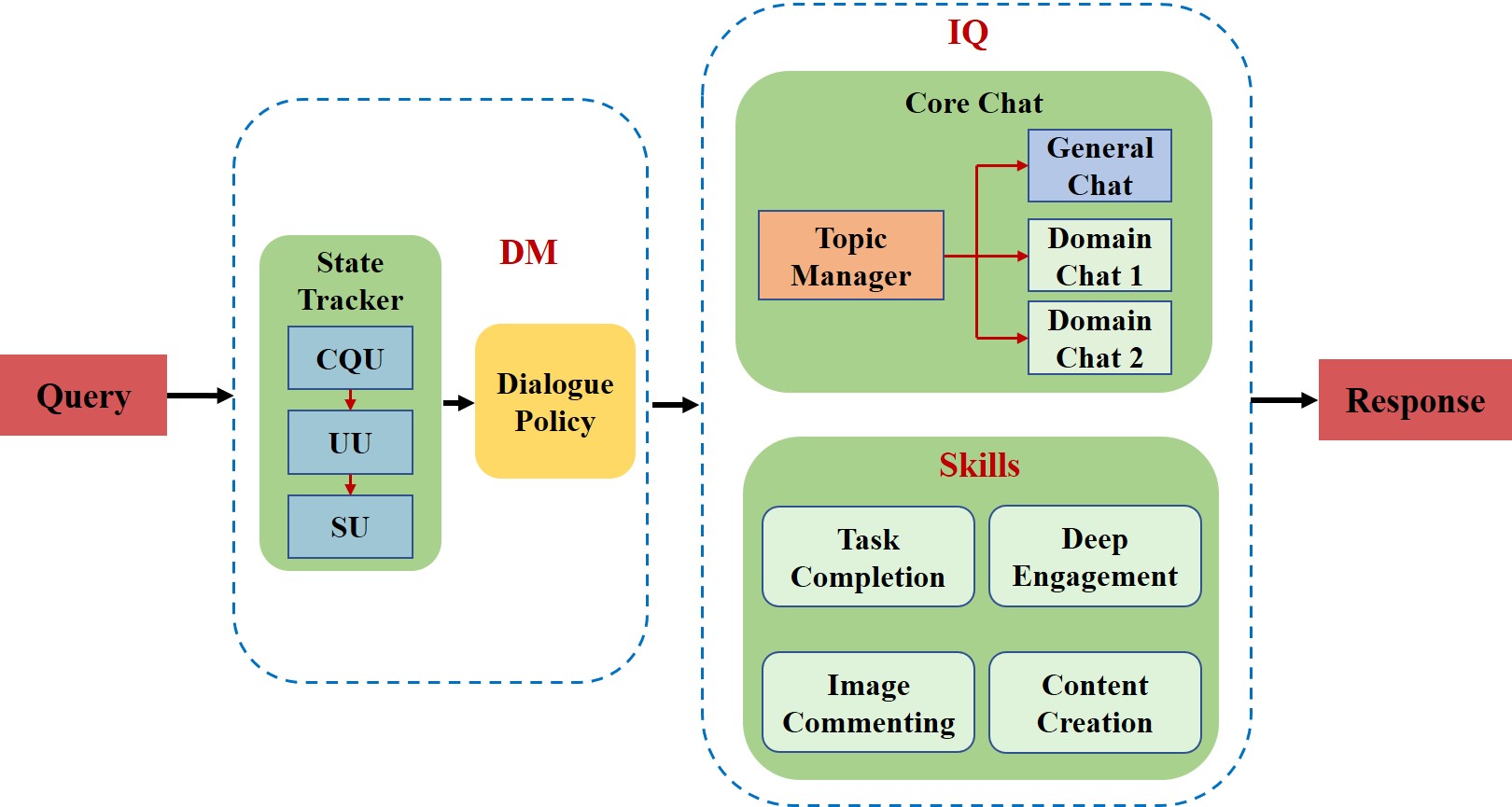}
    \caption{The overall framework of XiaoIce}
    \label{img:xiaoice}
\end{figure}

XiaoIce is based on an empathic computing framework that enables chatbots to recognize human emotions and states, understand users' intentions, and respond dynamically to users' needs. Integration of IQ, EQ, and Personality is core to XiaoIce's system design. IQ capacities include knowledge and memory modeling, image and natural language understanding, reasoning, generating, and predicting. These are the foundations for developing conversational skills that allow chatbots to meet the specific needs of users and help the user accomplish specific tasks. EQ refers to the need for chatbots to be emotionally intelligent enough to produce socially attractive responses (\textit{e.g.}, having a sense of humor, comforting, etc.), and to be able to decide to drive a conversation to a new topic when it comes to a standstill, or to actively listen when the user engages in the conversation. Personality is defined as the characteristic set of behaviors, cognitions and emotional patterns that form an individual's distinctive character. Social chatbots need to hold a consistent personality and set the right expectations for the user during a conversation to gain users' long-term confidence and trust. The overall framework of XiaoIce is shown in the figure~\ref{img:xiaoice}.

After receiving the information from users, the system will use Dialogue Management (\textit{DM}) module to manage the whole dialogue process, in which the Global State Tracker is responsible for updating the system status. The Global State Tracker utilizes Contextual Query Understanding (\textit{CQU}), User Understanding (\textit{UU}), and System Understanding (\textit{SU}) module respectively to integrate the dialogue context, users' characteristics, and system state for accurate system state capturing. Then the Dialogue Policy module decides the following dialogue strategy according to the updated dialogue status, that is, whether the query is to be answered by the Core Chat or a certain skill.

The Core Chat is designed for open domain conversation, which is divided into General Chat and Domain Chat. General Chat mainly answers general questions, while Domain Chat focuses on answering professional questions in the particular domain. The main realization technique of chitchat is the combination of retrieval model and sorting model, in which the retrieval model generates candidate response sets, and the sorting model sorts the candidate responses to output the response with the highest score. In addition to the Core Chat, XiaoIce also has a number of constantly updated skills, including Image Commenting aiming to generate comments based on user input images, Content Creation accomplishing creative work such as poetry creation and story generation.

\textbf{\textit{Question Answering. }} In addition to the dialogue systems, the Question Answering (QA) system, providing corresponding answers to users' different questions, is another typical application of text generation. The QA system needs to find relevant content through search engines or knowledge bases to organize the corresponding answers which may relate to commonsense facts or details about specific events. Dehghani \textit{et al.} \cite{dehghani2019learning} propose a QA model, called \textit{TraCRNet}, to achieve the goal of open-domain query answering. The TraCRNet model reasons to correctly answer the question by utilizing information of multiple documents extracted from a search engine which includes not only the high-ranked web pages but also the low-ranked web pages that are not directly related to the question.

\begin{figure}[htbp]
    \centering
    \includegraphics[scale=0.5]{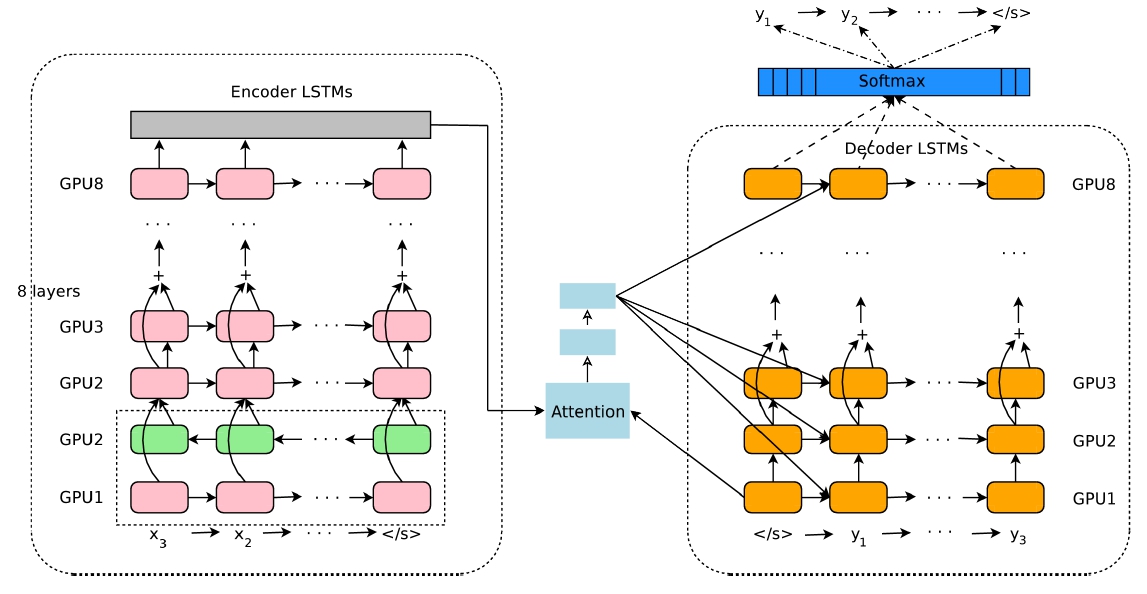}
    \caption{The model architecture of GNMT}
    \label{img:google}
\end{figure}

\textbf{\textit{Machine translation. }} With the development of economic globalization, the world has become a small village where people from all over the world can communicate with each other via the Internet. As a result, translation becomes an essential requirement for better communication with each other. Machine translation systems also provide researchers with the opportunity to exchange ideas with researchers around the world, enabling scientific researches to develop more rapidly and vigorously. Google translate\footnote{\url{https://translate.google.com/}} is a free translation service offered by Google. It provides instant translation between 80 languages and supports the translation of words, sentences, and web pages between any two languages. According to statistics, Google translate translates over 100 billion words every day, which is one of the most popular translation software in the world. The Google Translation team puts forward the Google Neural Machine Translation system (\textit{GNMT}) \cite{wu2016google} for the first time in 2016, which consists of a deep LSTM network with 8 encoder and 8 decoder layers using residual connections as well as attention mechanism. The model architecture of GNMT is shown in the figure~\ref{img:google}.

The GNMT is a typical sequence-to-sequence learning framework with the attention mechanism, composed of an encoder, a decoder, and an attention network. The encoder encodes the input sentence into vector representations, and the decoder generates one word at a time according to the current states. The attention mechanism allows the decoder to focus on the important information in the source sentence to build the connection of the encoder and decoder. With a large amount of paired translation data for training, GNMT can achieve excellent translation performance. In the meanwhile, to solve problems in neural network model training, Google proposes a lot of training tricks to improve the performance of the model. The low-precision arithmetic is employed in the inference computations to accelerate the final translation speed, the length-normalization procedure and coverage penalty are employed in the beam search technique to encourage the output sentence to cover all the words in the source sentence. The words in the input and output are divide into a limited set of common sub-word units (``wordpieces'') to balance the flexibility of ``character''-delimited models and the efficiency of ``word''-delimited models for improving handling of rare words.

After the GNMT system, Google proposes the Transformer model for the machine translation task, which completely abandons the common network structure in RNN, and only adopts the attention mechanism to carry out sequence modeling. After that, Transformer has become a standard structure for many NLP tasks, giving a big boost to the development of the NLP field.

\textbf{\textit{Product review and advertisement generation.}} Due to the large number of product reviews in the shopping websites, writing reviews for products may puzzle customers and waste their time. Fortunately, based on the information of a given product and the rating of the review, reviews can be automatically generated by the review generation technology, providing references for other customers. Ni \textit{et al.} \cite{ni2018personalized} build an assistant system for helping users to write reviews. This model expands the contents of input phrases and conforms to users' personalized aspect preferences to generate diverse and smooth product reviews. At the same time, writing specific advertisements for vast products is also a time-consuming task, particularly when we want to generate personalized ads for each consumer. The personalized advertisement generation technology can automatically generate well-suited product description according to different selling points and user preferences/traits. It not only provides great convenience for consumers, but also for sellers. Chen \textit{et al.} \cite{chen2019towards} propose a personalized product description generation model by leveraging neural networks combined with the knowledge base.

\textbf{\textit{Text summarization.}} In recent years, the volume of text data from various sources has exploded. Text summarization systems help people quickly understand the main content of information and improve the efficiency of information acquisition. According to Radef \textit{et al.} \cite{radev2002introduction}, a summary is defined as ``\textit{a text that is produced from one or more texts, that conveys important information in the original text(s), and that is no longer than half of the original text(s) and usually, significantly less than that}''. The purpose of a text summarization system is to produce a concise and fluent summary of the source articles while retaining key information content and overall meaning. Nallapati \textit{et al.} \cite{nallapati2016abstractive} firstly introduce the attentional encoder-decoder architecture into text summarization system, and achieve the state-of-the-art performance, which provides the direction for the follow-up research works.

\textbf{\textit{Data storytelling.}} There is a lot of structured data in real life. Computers are good at analyzing structured data, while humans are more inclined to read complete stories rather than a jumble of data. Therefore, how to make use of structured data to create more personified stories is a key research direction of text generation. Data storytelling is the key technology to achieve this goal, whose typical applications include news report generation, company report generation, sports event report generation, and so on. News report generation systems can produce complete, meaningful, and reasonable news according to the time, place, cause, process, and other factors of the specific news event, while sports report generation systems can generate detailed game reports according to the situation of sports events. Narrative Science \footnote{\url{https://narrativescience.com/}} is a text report generation company whose purpose is to give life to data. It is dedicated to the research of automatic text generation technology, which can automatically generate a high-quality text after a planning process based on the key information in the data and its expression in the machine.

\textbf{\textit{Image captioning and visual QA.}} Text information is just one way for human to obtain information, while we are more likely exposed to image information in real life. Image captioning technology can automatically generate corresponding text description according to the content of images, so as to facilitate readers to better understand image contents. Feng \textit{et al.} \cite{feng2019unsupervised} train an unsupervised image captioning model to generate image captions without the paired image-sentence datasets. Visual QA is another interactive point between text generation and image understanding. By understanding image content and corresponding questions at the same time, visual QA technology can generate answers of questions related to the images. Li \textit{et al.} \cite{li2019visual} convert visual QA question into a machine reading comprehension problem combined with the large-scale external knowledge base to realize the knowledge-based visual QA.

With the rapid development of NLP technology, it is ubiquitous to find the usage of text generation technology in our daily life. Nevertheless, text generation has been far from mature. For example, it is still easy to find out whether we talk to a real person or a chatbot, which means that an obvious gap still exists between robots and human beings. CTG is the key factor to solve this problem, which aims to generate the high quality and anthropopathic textual content. In this paper, we will study different types of conditional text generation fields.

\section{MAJOR RESEARCH AREAS}
After introducing the definition and application scenarios of CTG, in this section, we make a detailed investigation of different CTG fields, including context-based text generation, personalized text generation, topic-aware text generation, emotional text generation, knowledge-enhanced text generation and visual text generation. Furthermore, summarize the multi-conditional text generation and pre-trained language model-based CTG works.

\subsection{Context-based Text Generation}
In many applications of text generation, the context information is the key factor to realize the coherence and smoothness of the generated text. The context means the situations in which natural languages are generated. In dialogue systems, the context usually refers to the dialogue history that has taken place in multi-rounds dialogues. The ability to consider previous utterances is the core to build active and engaging dialogue systems \cite{chen2017survey}. Meanwhile, in review generation systems, the context refers to the time, emotions, sentiments and other factors. The context information provides clues to the generation of natural language \cite{tang2016context}. Therefore, in order to generate high quality text, it is necessary to consider the context information in CTG. We give a brief summary of context-based text generation methods in Table~\ref{tab:context-based}.

\begin{table}
 \caption{A summary of context-based text generation methods}
 \label{tab:context-based}
 \begin{tabular}{|m{2.5cm}<{\centering}|m{4cm}<{\centering}|m{6cm}<{\centering}|}
    \hline
    \textbf{Work} & \textbf{Method} & \textbf{Description}\\
    \hline
    Sordoni \textit{et al.} \cite{sordoni2015neural} & Context embedding & Embedding all the words and phrases in the dialogue history into continuous representations as additional inputs of the decoding stage\\
    \hline
    Voita \textit{et al.} \cite{voita2018context} & Context embedding & Encoding the source and the context sentence separately and learning the context-aware representation of the source sentence through the attention mechanism\\
    \hline
    Serban \textit{et al.} \cite{serban2016building}& Hierarchical context embedding & Embedding the word sequences in each context sentences at the low-level and embedding the sentence sequences in the historical dialogues at the top-level to efficiently capture the context information\\
    \hline
    Xing \textit{et al.} \cite{xing2018hierarchical} & Hierarchical context embedding& Leveraging the attention mechanism to extend the HRED model\\
    \hline
    Jaech \textit{et al.} \cite{jaech2018low} & Context Adaption & Using the context information (dialogue history) to transform the weights of recurrent units in RNN to effectively capture high-dimensional context\\
    \hline
  \end{tabular}
 \end{table}

\textbf{\textit{Context embedding.}}
The most simple and effective way to combine context information is to embed it directly to get the vector representations as a part of the decoder input, so that context information can be considered in the generated text. Among the numerous tasks of context-based text generation, the utilization of context in dialogue systems attracts great attention of researchers. For instance, Sordoni \textit{et al.} \cite{sordoni2015neural} embed all words and phrases in the dialogue history into vector representations. Dialogue history information is encoded into vectors, which are decoded by another RNN to produce context-aware responses. In addition to the dialogue system, other applications also need to add different types of context information. Tang \textit{et al.} \cite{tang2016context} define the \textit{context} as the information or situations that may influence the output content. The encoder encodes the contexts information (\textit{e.g.}, a sentiment score, a product id or a user id) into continuous semantic representations and concatenates them as the input of the decoder.  During decoding, the context information are attended through a gating mechanism to generate context-sensitive product reviews. Similarly, Clark \textit{et al.} \cite{clark2018neural} propose a text generation model in stories, which treats entity representations extracted from dialogue history as context. By encoding the historical conversations together with the entity representations as context, the model can better determine which entities or words to be mentioned next.

When considering the context information in machine translation systems, multiple sentences can be treated as a whole, and relevant information between sentences can be captured, which not only prevents errors in the case of ambiguity, but also improves the consistency of translation. Voita \textit{et al.} \cite{voita2018context} propose a context-aware machine translation model that can control and analyze the flow of information from context to the translation model. The source sentence needed to be translated and the context sentence are encoded separately to get the context-aware representation of the source sentence through the attention mechanism. It is identified that pronoun is the key information captured by the model. In the field of machine translation, Kang \textit{et al.} \cite{kang2020dynamic} proposed to select contextual sentences dynamically for each source sentence to be translated. The Context Scorer module is used to score each context sentence based on the currently translated source sentence and incorporate important context sentences into the translation module.

\textbf{\textit{Hierarchical context embedding}}
Instead of embedding context information directly, hierarchical context embedding method divides the embedding process into two steps to capture information in context more effectively. Concretely, the first step is to embed the word-level information, and the second step is to embed the sentence-level information. Serban \textit{et al.} \cite{serban2016building} use Hierarchical Recurrent Encoder-Decoder (\textit{HRED}) model to hierarchically encode the dialogue history and guide the generation of replies. In particular, the word sequences in each context sentence are encoded at the low-level, while the sentence sequences in the historical dialogues are encoded at the top-level. Xing \textit{et al.} \cite{xing2018hierarchical} leverage the attention mechanism to extend the HRED model. By incorporating attention mechanism at the words and sentences level respectively, the model captures the most important parts in the context. Zhang \textit{et al.} \cite{zhang2018less} observe that we can have more smooth conversations without much context information in the multi-user dialogue, and produce a tree-based hierarchical multi-user dialogue model, which builds a tree structure consisting of many branches for multi-user conversations to select exact context sequences. Besides, Tian \textit{et al.} \cite{tian2017make} conduct a comprehensive survey on existing context-aware conversational models and find that compared with the non-hierarchical model, the hierarchical model is more capable for capturing context information.

\textbf{\textit{Context adaption.}} The context adaption method changes the model itself rather than regarding context as an extra input of CTG model. For example, Jaech \textit{et al.} \cite{jaech2018low} utilize the context information to transform weights of the recurrent layer in RNN. In particular, they utilize a low-rank decomposition algorithm to control the degree of parameter sharing in context, which performs well on high-dimensional and sparse context.

\textbf{\textit{VAE-based methods.}} How to ensure the coherence of context in the generation of long-form text (\textit{e.g.}, a single or multiple paragraphs) is a challenging problem. Both high-level abstract features (\textit{e.g.}, topics, sentiments, etc.) and low-level fine-grained features (\textit{e.g.}, specific word choices) should be considered to generate globally-coherent long text sequences. Traditional RNN models tend to generate repetitive and inconsistent long-form text due to the poor feature extraction ability, while VAE models based on deep latent variables can capture these high-lever features to generate coherent and high-quality long text. For example, Shen \textit{et al.} \cite{shen2019towards} propose a VAE-based multi-level network structure for CTG, which contains a multi-level decoder to capture coherent long-term structure inherent in long-form text by generating high-level intermediate representations of input sentences. Meanwhile, multiple stochastic layers are used between VAE's encoder and decoder to generate more semantically-rich latent variables for producing more coherent and less repetitive long text. Shao \textit{et al.} \cite{shao2019long} firstly design a sequence of groups, and subsequently generates each sentence conditioned on the planning result and previously generated context sentences. A hierarchical latent structure containing global planning and local sequence latent variables is used to improve the diversity of the generated text.

\subsection{Personalized Text Generation}
Various human characteristics significantly impact interpersonal communication and writing styles. In other words, personalization plays a key role in enhancing the quality of CTG model. In dialogue systems, personalization is vital for creating truly smart dialogue agents which can be seamlessly incorporated into the lives of human beings. In product review generation systems, personalization ensures the generated product review depends not only on the attributes of the product, but also on the preferences of specific users, endowing the authenticity for the generated reviews. Many efforts for personalized text generation are conducted, as summarized in Table ~\ref{tab:personalized}, and we will discuss them in details below.

\begin{table}
 \caption{A summary of personalized text generation methods}
 \label{tab:personalized}
 \begin{tabular}{|m{2.5cm}<{\centering}|m{4cm}<{\centering}|m{6cm}<{\centering}|}
    \hline
    \textbf{Work} & \textbf{Method} & \textbf{Description}\\
    \hline
    Li \textit{et al.} \cite{li2016persona} & RNN + Speaker model & The speaker model encodes each individual speaker into a vector to capture characteristics; Generating personal responses matching a specific user\\
    \hline
    Luan \textit{et al.} \cite{luan2017multi} & Autoencoder + Multi-task learning & Training a response generation model on a small personalized dialogue data, and then training an autoencoder model with non-conversational data; Sharing parameters of the two models to obtain the personalized dialogue model\\
    \hline
    Yang \textit{et al.} \cite{yang2017personalized} & Transfer learning + Pretrain and fine-tuned & Respectively using massive generic dialogue data and a small-scale personalized dialogue data to pre-trained and fine-tune the dialogue model to generate personalized responses\\
    \hline
    Yang \textit{et al.} \cite{yang2018investigating} & Reinforcement Learning + Persona embedding & Embedding user-specific information into vector representation; RL mechanism optimizes three rewards -- topic coherent, informative and grammatical, to generate more personalized responses\\
    \hline
  \end{tabular}
 \end{table}

\textbf{\textit{Personalized feature embedding.}} The simplest method to achieve personalized text generation is  to embed the personalized characteristics of different users. Li \textit{et al.} \cite{li2016persona} present a speaker model which encodes user profile into vectors so as to capture personalized features and guide the response generation during the decode stage. Instead of encoding personalized features into vector representations directly, Herzig \textit{et al.} \cite{herzig2017neural} use an additional neural network to capture the high-level personalized information based on the personality traits. The additional layer implicitly influences the decoding hidden state to ensure that the personalized features are integrated into the generated text. Li \textit{et al.} \cite{li2019towards} propose the User-aware Sequence Network (\textit{USN}), to generate a summary for a user's review according to his preference on different aspects or writing style. The user-aware encoder selects the user-concerned information in a review, and the user-aware decoder combines user characteristic and user-specific language habits into word generation. Zheng \textit{et al.} \cite{zheng2019personalized} propose a trait fusion module to capture the persona information of each speaker. Each persona trait is encoded into vector representation and all traits are merged to get the integrated persona vectors. The persona aware attention mechanism controls the attention weights of the context vector and the persona-aware bias estimates the word generation distribution. Luo \textit{et al.}  \cite{luo2019learning} build a personalized goal-oriented dialog system for the restaurant reservation task. The profile model encodes user profiles into vector representations and storages conversation history from similar users. The preference model captures user preferences over knowledge base entities and combines with the profile model to enhance the performance in terms of task completion and user satisfaction. To improve the personality consistence of the generated dialogue responses, Song \textit{et al.} \cite{song2020generate} propose the generate-delete-rewrite mechanism to delete inconsistent words from a generated response prototype and further rewrite it to a personality-consistent one.

\textbf{\textit{Multi-task and transfer learning.}} The personalized text datasets are so scarce that the above models are difficult to perform very well. Some researchers attempt to enhance the performance of personalized text generation by transfer learning and multi-task learning models. For instance, the work of Luan \textit{et al.} \cite{luan2017multi} trains a dialogue model to predict responses given previous contexts and an autoencoder model with large volumes of non-conversational personal data to model the role-specific characteristics of different users. Through the multi-task learning mechanism which shares the decoder parameters of the two models, these models can capture speaker roles, expressive styles and domain expertise characteristic of the targeted user and generate personalized responses without heavy recourse to each speaker's conversational data. Yang \textit{et al.} \cite{yang2017personalized} propose a domain adaptation-based personalized dialogue model. They respectively use massive generic dialogue data and a small-scale personalized dialogue data to pre-trained and fine-tune the dialogue model, and apply the policy gradient algorithm to improve the personalized and informative features of generated responses. Similarly, Zhang \textit{et al.} \cite{zhang2019neural} put forward the Learning to Start (\textit{LTS}) model to optimize the quality of responses, which divides the training process into initialization (modeling the responding style of human) and adaptation (generating personalized responses) for generating relevant and diverse responses.

\textbf{\textit{GAN and RL models.}} Our writing style can be perceived by his specific word usage manners, which means different language habits can reflect our personalized characteristics. Yuan \textit{et al.} \cite{yuan2019personalized} propose a personalized sentence generation model based on GAN. In the training procedure, the frequently used words are incorporated as the input sources. Then the sentence structure is constrained to generate sentences similar to the original sentences of the same author. RL can control the quality of generated content through different policies or rewards, so researchers consider incorporating it to implement personalized text generation. Yang \textit{et al.} \cite{yang2018investigating} present the attention-based hierarchical encoder-decoder architecture via RL to realize personalized dialogue generation, which defines three types of reward mechanisms, including \textit{topic coherence}, \textit{mutual information}, and \textit{language model} to force the text generation model to generate topic-relevance and coherent dialogue responses.

\textbf{\textit{VAE-based methods.}} The above embedding based personalized text generation methods learn user information from training data and cannot discover the common properties among users. User-level features can also be depicted through latent variables, so VAE-related models are introduced into personalized text generation. Wasserstein Auto-Encoders (\textit{WAE}) is a typical VAE variation, which conducts adversarial training on latent variables, instead of assuming that latent variables subject to a simple Gaussian distribution to fit the real data distribution and improve the generation ability of VAE. Chan \textit{et al.} \cite{chan2019modeling} embed and mix the user-level and sentence-level information into multimodal latent distributions. The mixed distribution is then regarded as the prior distribution of WAE, and extended to the Gaussian Mixture Distributions to guide the decoder to generate personalized responses for different users. To generate product tips with personalized features of users, Li \textit{et al.} \cite{li2019persona} present the Persona-Aware Tips Generation model (\textit{PATG}), which employs adversarial variational auto-encoders to model the persona information of different users. The persona information is distilled from all historical tips and reviews of a target user and expressed by the latent variables in VAE. An external memory-based Pointer Network is also deployed to conduct the memory reading to retrieve more accurate persona information.

\subsection{Topic-aware Text Generation}
Topic information is indispensable in our daily communication, reading or writing. We usually have a conversation around a specific topic and usually identify the topic of an article before we read or write it. In this subsection, we give a review of topic-aware text generation studies, as presented in Table~\ref{tab:topic-aware}.

\begin{table}
 \caption{A summary of topic-aware text generation methods}
 \label{tab:topic-aware}
 \begin{tabular}{|m{2.5cm}<{\centering}|m{4cm}<{\centering}|m{6cm}<{\centering}|}
    \hline
    \textbf{Work} & \textbf{Method} & \textbf{Description}\\
    \hline
    Xing \textit{et al.} \cite{xing2017topic} & RNN + LDA + Topic embedding & Utilizing LDA to get topic information, and embedding the topic words into the vector; Generating more informative, and topic relevant responses\\
    \hline
    Dziri \textit{et al.} \cite{dziri2019augmenting}& HRED + LDA + Topic embedding & Combining topic and context information to produce not only contextual but also topic-aware responses\\
    \hline
    Wang \textit{et al.} \cite{wang2018reinforced} & CNN + LDA + RL & Using LDA to get topic information, CNN to capture the dialogue information, and RL to optimize the model with specific evaluation metric; Generating coherent, diverse, and informative text summaries\\
    \hline
    Feng \textit{et al.} \cite{feng2018topic} & RNN + Topic embedding & Assigning each topic with different weight to maintain a multi-topic coverage vector and updating them in the decoding process in order\\
    \hline
  \end{tabular}
 \end{table}

\textbf{\textit{Topic extracting and embedding.}} It is a common idea to extract topic information from existing text and embed it into vector representations to guide text generation. Xing \textit{et al.} \cite{xing2017topic} propose a topic-aware Seq2seq (\textit{TA-Seq2Seq}) model to generate informative and interesting responses for chatbots. TA-Seq2Seq incorporates topic information of the dialogue history extracted by the pre-trained \textit{LDA} model with the input sentence, and utilizes the joint attention mechanism to guide the generation process. Choudhary \textit{et al.} \cite{choudhary2017domain} observe that topic information can be divided into multiple domains (\textit{e.g.}, games, sports or movies) to provide the fine-grained guidance for the generator. They adopt domain classifiers to capture domain information from the dialogue history for generating domain-relevant responses. Feng \textit{et al.} \cite{feng2018topic} develop a multi-topic-aware LSTM (MTA-LSTM) model to generate a paragraph-level text under target multiple topic words. In the MTA-LSTM model, each topic will be assigned with different weights to maintain a multi-topic coverage vector, which is updated in the decoding process in order. Then the vector will guide the generator to generate the topic-aware text with an attention module. A long article usually spans many topics, while a simple text summary usually cannot cover all topics. To generate text summaries of specific topics of interest to users, Krishna \textit{et al.} \cite{krishna2018generating} propose to generate multi summarizations for a given article according to different topics. With an article and a topic of interest as input, the proposed pointer-generator network will pay higher attention to the relevant parts of the topic in the input article to generate topic-tuned summarizations. Dziri \textit{et al.} \cite{dziri2019augmenting} introduce a Topical Hierarchical Recurrent Encoder Decoder (\textit{THRED}) model to generate contextual and topic-aware responses. THRED hierarchically encode the dialogue history in the word and sentence level respectively and capture topic information from dialogue context using a pre-trained LDA model. In the decoder, the generation probability is biased towards generating topic words by adding an extra probability to the original generation probability.

\textbf{\textit{CNN-based methods.}} The above studies mostly employ the RNN model in the task of topic-aware text generation. In addition to RNN, several other models are also applied to this task, which also achieve remarkable results. For instance, Wang \textit{et al.} \cite{wang2018reinforced} propose a topic-aware convolutional Seq2seq (\textit{ConvS2S}) model, which leverages the joint attention and biased probability generation mechanism for incorporating topic information. Word and topic embeddings of the source sequence are encoded by the associated convolutional blocks. Then the joint attention mechanism attends to words and topics according to the decoder states, and the biased probability generation is performed to generate coherent, diverse, and informative summaries,

\textbf{\textit{VAE-based methods.}} The latent variables in VAE can capture features implicitly in natural language, which is useful for providing high-level guidance to text generation. If latent variables correspond to a specific topic, the probability distribution of generated words will be narrowed down, thus improving the rationality of generated content. Consequently, the VAE is widely used in topic-aware text generation. For instance, Wang \textit{et al.} \cite{wang2019topic} propose a topic-guided variational autoencoder (\textit{TGVAE}) method to generate natural language text under the guidance of the designated topic. Specifically, TGVAE generates a Gaussian mixture model (\textit{GMM}) for latent variables as the prior distribution which is parametrized by a neural topic module responsible for capturing long-range semantic information in the whole document. Each mixture component corresponds to a specific topic, which guides to generate semantically-meaningful sentences under the given topic. Gao \textit{et al.} \cite{gao2019topic} propose a neural variational language model to study the topic-level Gaussian distributions in latent space. They utilize CNN to get the vector representations of input sentences and predict the Gaussian distribution of topics using the full connection layer. By sampling the topic distribution, the proposed model can generate diversified sentences conditioned on given topics. In previous works of text generation based on VAE, the distribution of latent variables is usually assumed as Gaussian distribution, which makes it difficult to distinguish which part of latent variables controls the structure and which part controls the semantics of natural language. In order to solve this problem, Li \textit{et al.} \cite{li2019topic} develop the \textit{TATGM} model, which adopts a sequential VAE to learn the structural features of text and a topic model to extract the semantic features of text to generate different expressions of the same structure in different topics. TATGM's topic model generates text based on the Gaussian distribution of latent variables, which ensures the capture of textual semantic information. At the same time, the encoder acts as a discriminator to force the decoder to generate the text with similar semantics.

\subsection{Emotional Text Generation}
Natural language is full of emotions, and emotional words are more likely to stimulate the interest of readers. Additionally, people adjust their speaking style and content according to their own and other people's emotional changes in daily communication. Due to the necessity of integrating emotional information, researchers pay attention to incorporating emotional information into the generated text in order to provide users with better experience, as summarized in Table~\ref{tab:emotional}.

\begin{table}
 \caption{A summary of emotional text generation methods}
 \label{tab:emotional}
 \begin{tabular}{|m{2.5cm}<{\centering}|m{4cm}<{\centering}|m{6cm}<{\centering}|}
    \hline
    \textbf{Work} & \textbf{Method} & \textbf{Description}\\
    \hline
    Zhou \textit{et al.} \cite{zhou2018emotional} & GRU + Emotional embedding & Emotion category embedding captures emotional information and the internal emotion memory balances the grammaticality and the expression degree of emotions\\
    \hline
    Fu \textit{et al.} \cite{fu2018style} & GRU + Multi-task learning & Multi-decoder Seq2seq module generates outputs with different styles and style embedding module augments the encoded representations\\
    \hline
    Kong \textit{et al.} \cite{kong2019adversarial} & Conditional GAN (CGAN) + Sentiment control & The generator generates sentimental responses based on a sentiment label and the discriminator distinguishes the generated replies and real replies\\
    \hline
    Li \textit{et al.} \cite{li2019reinforcement} & RL + Emotional editor & The emotional editor selects the template sentence based on the topic and emotion, and RL forces the model to enhance the coherence and emotion expression of generated responses\\
    \hline
  \end{tabular}
 \end{table}

\textbf{\textit{Emotion extraction and embedding.}} One common way to generate emotional text is to extract emotional information from input text and embed it into vector representations as the input of the decoder. Asghar \textit{et al.} \cite{asghar2018affective} propose an LSTM-based emotional dialogue generation model with three designed mechanisms to incorporate affective/emotional aspects into generated responses. The affective word embeddings introduce an external cognitively engineered affective dictionary to augment traditional word embeddings with affective vectors. The affective loss functions minimize the Euclidean distance between the affective embeddings of inputs and responses, and maximize the affective content of responses to explicitly train an affect-aware model. The affectively diverse beam search injects affective dissimilarity across the beam groups based on affective word embeddings to promote the generation of emotionally rich responses. Zhou \textit{et al.} \cite{zhou2018emotional} produce the Emotional Chatting Machine (\textit{ECM}) to generate grammatically correct, context-relevant and emotionally consistent responses. The ECM leverages emotion category embedding for capturing high-level abstraction of emotion expressions, an internal emotion state for balancing grammaticality and emotion dynamically, and an external emotion memory to help generate more explicit and unambiguous emotional expressions. Majumder \textit{et al.} \cite{majumder2020mime} consider that emotional responses often mimic the emotion (positive or negative) of the user. The emotion stochastic sampling and emotion mimicry mechanism are proposed to encode context and emotions to generate appropriate and empathetic responses.

\textbf{\textit{Emotion transferring.}} In addition to embed emotional information directly, the transfer from one emotion to another is a promising way to generate emotional text. Fu \textit{et al.} \cite{fu2018style} achieve the goal of transferring the emotion of reviews from positive/negative to negative/positive through multi-task learning and adversarial training. They leverage a style embedding module to augment the language style representations and a multi-decoder Seq2seq model to respectively generate text with different styles. Luo \textit{et al.} \cite{luo2019towards} propose the \textit{Seq2SentiSeq} model, which adopts the Gaussian Kernel layer to incorporate the numeric sentiment intensity value into the decoder, so as to finely control the sentiment intensity of generated text. At the same time, the cycle reinforcement learning algorithm controls the process of model training which balances both sentiment transformation and content preservation through the elaborately designed rewards to tackle the problem of lacking parallel data.

\textbf{\textit{GAN and RL models.}} Several works confirm that GAN and RL have significant effects on emotional text generation. For example, Wang \textit{et al.} \cite{wang2018sentigan} propose the SentiGAN with multiple generators and one multi-class discriminator to enhance the sentiment accuracy and quality of generated texts. In the SentiGAN, multiple generators are trained simultaneously to generate texts with different sentiment labels, such as positive or negative, and the multi-class discriminator makes each generator focus on generating its own examples of a specific sentiment label accurately. Kong \textit{et al.} \cite{kong2019adversarial} introduce a conditional GAN (\textit{CGAN})-based sentiment-controlled dialogue generation model. The generator of CGAN produces sentimental responses under the given dialogue history and sentiment label, while the discriminator identifies the quality of generated responses through checking whether the items (\textit{e.g.}, dialogue history, sentiment label, and dialogue response) belongs to the real data distribution. Li \textit{et al.} \cite{li2019reinforcement} propose the emotional editor module to select the template sentences according to emotion and topic information in the dialogue history, and introduce RL to promote the quality of generated responses from three points: emotion, topic and coherence.

\textbf{\textit{VAE-based methods.}} Utilizing latent variables of VAE to control the sentiment of generated text has also been explored by researchers. For example, Hu \textit{et al.} \cite{hu2017toward} combine VAE and holistic attribute discriminators for effective imposition of semantic structures. They allocates one dimension of the latent representation to encode ``positive'' and ``negative'' semantics, to capture a salient attribute independent with other features. The global discriminators facilitate effective imposition of latent code semantics to guide the discrete text generator learning. Chen \textit{et al.} \cite{chen2019sentiment} endow the poetry generator with the ability to express the specific sentiment (\textit{e.g.}, negative and positive), to improve the semantics and diversity of generated poems. Since sentiments are often strongly coupled with semantics in poetry, the authors make latent variables conditioned on both sentiment and text content to capture generalized sentiment-related semantics. Besides, a temporal sequence module captures sentiment transition patterns among different lines of the poetry to generate diverse poems under the control of semantic-level and line-level sentiments.

\subsection{Knowledge-enhanced Text Generation}
Nowadays, most text generation systems take advantage of deep neural network models to generate fluent, semantic and consistent text. However, there is still a big gap between such machine-generated text and human expression, that is human will combine their knowledge in speaking or writing, while most text generation systems fail to achieve this. By combining sufficient knowledge, such as commonsense knowledge and information about specific objects/events, CTG systems can generate more logical, credible, and informative text. The external knowledge includes structured knowledge graph (KG), which is composed of knowledge triples with the form of $<head ,relation,tile >$, and unstructured knowledge base (KB), which is composed of natural language text about specific concepts. There are many ways to combine external knowledge in CTG systems, as summarized in Table~\ref{tab:knowledge}, and we will introduce them in details below.

\begin{table}
 \caption{A summary of knowledge-enhanced text generation methods}
 \label{tab:knowledge}
 \begin{tabular}{|m{2.5cm}<{\centering}|m{4cm}<{\centering}|m{6cm}<{\centering}|}
    \hline
    \textbf{Work} & \textbf{Method} & \textbf{Description}\\
    \hline
    Ghazvininejad \textit{et al.} \cite{ghazvininejad2018knowledge} & Keyword matching + Facts embedding & The Facts Encoder module leverages an external memory for embedding the conversation-related facts to generate content-rich responses\\
    \hline
    Dinan \textit{et al.} \cite{dinan2018wizard} & Transformer + Memory Network & Memory Network retrieves knowledge about the dialogue from the memory and Transformer encodes and decodes the text representations to generate responses; Conducting knowledgeable discussions on open-domain topics\\
    \hline
    Zhou \textit{et al.} \cite{zhou2018commonsense} & Knowledge graph attention& Graph attention mechanisms integrate commonsense information from the knowledge based on the dialogue history; Generating more appropriate and informative responses\\
    \hline
    Mazumder \textit{et al.} \cite{mazumder2018towards} & Lifelong learning + Open-world knowledge base completion & Obtaining new knowledge by asking users when facing unknown concepts and then inferencing to grow knowledge over time\\
    \hline
  \end{tabular}
 \end{table}

\textbf{\textit{Unstructured KB enhanced models.}}
Unstructured KBs contain abundant knowledge with textual form. How to extract the knowledge related to input and combine it into the generation stage are main directions of current research. The common way to extract knowledge from unstructured KB is keyword matching with each word in the input as keywords. For instance, Ghazvininejad \textit{et al.} \cite{ghazvininejad2018knowledge} extract relevant knowledge facts from knowledge base using keyword matching and encode them into vectors to provide factual evidence for the dialogue response generation. In order to extract relevant knowledge more effectively, Ren \textit{et al.} \cite{ren2019thinking} propose to use the global perspective for selecting appropriate knowledge. A topic transition vector is obtained from the context to express global information and then used to guide the local knowledge extraction process for generating informative and fluent text. Zhao \textit{et al.} \cite{Zhao2020Low-Resource} separate parameters relying on knowledge-grounded dialogues in the whole model to get better results with insufficient knowledge-grounded dialogue data. In the decoder stage, the \textit{Language Model} generates common words, the \textit{Context Processor} module generates context words, and the \textit{Knowledge Processor} module generates words from knowledge document by a hierarchical attention mechanism. The decoding manager fuses the generation probability of three modules, and dynamically switches the decoding mode according to decoding states, to generate context-relevant, informative and reasonable responses.

After extracting the relevant textual knowledge, the next most important thing is to understand its semantics and integrate it into the text generation process. Young \textit{et al.} \cite{young2018augmenting} transform the extracted knowledge triples into a sequence tokens and encode them into vector representations using LSTM. The context vectors and knowledge vectors are concatenated to calculate match scores with different responses to select the most appropriate response. Wang \textit{et al.} \cite{wang2019augmenting} build a technical-oriented dialogue system to communicate with people about Ubuntu-relevant questions. The knowledge text descriptions are embed into vectors by word embedding average or BERT model which are concatenated with traditional word embeddings to enhance the understanding of the technical term in dialogue history. A knowledge reader attentively read knowledge embeddings and retrieve the semantic information at each decoding stage to generate informative responses. In addition to RNN, Dinan \textit{et al.} \cite{dinan2018wizard} combine \textit{Transformer} and \textit{Memory Network} to build an open-domain knowledge-based dialogue system. The Memory Network retrieves related knowledge from the Internet according to the input as the knowledge memory. Each sentence in the memory
is independently encoded with a Transformer encoder, and the same Transformer is used to encode the dialogue context. The standard dot-product attention between the memory candidates and the dialogue context is performed to select knowledge sentences to be used which are served as the input of the decoder to generate knowledgeable responses.

Due to the powerful performance of Transformer, more and more researchers begin to use it for knowledge understanding. Zhao \textit{et al.} \cite{zhao2019document} introduce the hierarchical interaction between the context and external document knowledge to capture the most important parts in the document and context using the multi-head attention module in Transformer for selecting the most appropriate response. Li \textit{et al.} \cite{li2019incremental} generate vector representations of external knowledge using the multi-head attention mechanism and then incorporates them to encode knowledge utterances span in the multi-turn dialogue. The decoder firstly generates contextual coherence responses attending on the context information and then refines them by attending on the knowledge vectors to generate more informative response. In knowledge-enhanced dialogue systems, there can be one-to-many relations between the dialogue context and the knowledge, which makes the knowledge selection is diverse and difficult. Kim \textit{et al.} \cite{Kim2020Sequential} propose to keep track of the prior and posterior distribution over knowledge using laten variables to improve the knowledge selection accuracy. Through sequentially modeling the history of knowledge selection in previous turns, the scope of probable knowledge candidates at current turn is reduced. The posterior distribution over knowledge leverages the response information to select knowledge more accurate.

\textbf{\textit{Structured KG enhanced models.}} Knowledge graph is a kind of structured knowledge base, which describes physical entities and their connections accurately. As its name suggests, knowledge graph is also helpful to build knowledge-enhanced text generation systems. Wang \textit{et al.} \cite{wang2019knowledge} propose an entity linking module to decide the optimal entity in the input question for selecting knowledge triples from external KG, which will be encoded as common words by the LSTM. The similarity scores of the question and relation candidates are calculated to select the most appropriate triples for question answering. The TranE algorithm \cite{bordes2013translating} is proposed to transform structured triples into low dimensional vector representations and is widely used in knowledge-enhanced text generation systems. Moussallem \textit{et al.} \cite{moussallem2019augmenting} link knowledge facts based on the translated document, encode them into vectors by TransE, and concatenate them with the internal vectors of NMT embeddings as the decoder input to enhance the quality of generated translations. Gune \textit{et al.} \cite{gunel2019mind} extract entities from external KG and adopt TransE to get vectors of them. The knowledge vectors are then fed into the separate multi-head attention channel to generate coherent text summaries. The attention mechanism is also applied in learning knowledge form knowledge graphs, which is called graph attention. Guan \textit{et al.} \cite{guan2019story} present an incremental encoding schema to mine hidden information in the story context and graph, and a contextual attention mechanism to encode knowledge graph into vectors. The multi-source attention mechanism is used to comprehensively understand the content of stories to generate reasonable and consistent story endings. Zhou \textit{et al.} \cite{zhou2018commonsense} produce a knowledge-based dialogue model (\textit{CCM}) that leverages two graph attention mechanisms to promote dialogue understanding and knowledgeable responses generating. The \textit{static graph attention} module encodes the graphs relevant to the dialogue history and concatenate graph vectors with input vectors to enhance the semantic information of the input. The \textit{dynamic graph attention} module attentively reads all the knowledge graphs and all triples in each graph based on decoder states to adaptively choose a generic word or an entity from the retrieved graphs for word generation.

To generate text with more entities, Moon \textit{et al.} \cite{moon2019opendialkg} propose the \textit{DialKG Walker} model that learns the symbolic transitions of dialog contexts as structured traversals over KG, and the \textit{graph decoder} that attends on viable KG paths to predict the most relevant entities in the KG, by associating these entities with the dialogue context and entities mentioned in the previous turn. Koncel \textit{et al.} \cite{koncel2019text} study the problem of generating paragraphs with multiple sentences given only a short title. The \textit{Graph Transformer} model computes the hidden representations of each node in a graph by attending over its neighbors following a self-attention strategy to leverage the relational structure of knowledge graph. The decoder attends on encodings of the knowledge graph and document title using the decoder hidden state to generate informative texts. Chen \textit{et al.} \cite{chen2019enhancing} propose a data-to-text generation model, which extracts entities appear in the data field and links them to Wikidata as external knowledge to form the temporary memory. The dual attention mechanism is applied to generate words conditioned on both input table information and background knowledge fact information. To consider the dialogue context in the knowledge retrieve process, Wu \textit{et al.} \cite{wu2020diverse} design a Felicitous Fact mechanism to help the model focus on the knowledge facts that are highly relevant to the context.

GCN \cite{kipf2016semi} is an extension of CNN in the graph domain, which can effectively learn the structural information of nodes and edges in the knowledge graph. De \textit{et al.} \cite{de2019question} regard the question answering problem as the graph inference problem. Nodes in this graph correspond to named entities in a document whereas edges encode relations between them (\textit{e.g.}, cross and within-document coreference links or simply co-occurrence in a document).  The entity graph relates mentions to entities within and across documents, the document encoder obtains representations of mentions in context, and the relational GCN propagates information through the entity graph to generate correct answers.

\textbf{\textit{Continuous learning models.}} Although existing studies introduce some real-world knowledge to CTG systems, the knowledge is usually fixed and cannot be expanded or updated. Continuous learning in the interactive surroundings is an important capability of human beings. We keep on learning and updating our knowledge according to our experiences in the daily life, which should be considered as an important factor when building humanoid text generation systems. Mazumder \textit{et al.} \cite{mazumder2018towards} build a knowledge learning model, namely lifelong interactive learning and inference (\textit{LiLi}), enabling chatbots to interactively and continuously learn new knowledge when communicating with users. By mimicking humans to acquire knowledge, Lili enquires ask users for related items when facing unknown concepts and then infers to expand knowledge over time.

\subsection{Visual Text Generation}
Since people usually gather information from images, visual text generation is also an important research direction in text generation. Two of the most important applications are image captioning and visual QA. A summary of visual text generation methods is given in Table~\ref{tab:visual}.

\begin{table}
 \caption{A summary of visual text generation methods}
 \label{tab:visual}
 \begin{tabular}{|m{2.5cm}<{\centering}|m{4cm}<{\centering}|m{6cm}<{\centering}|}
    \hline
    \textbf{Work} & \textbf{Method} & \textbf{Description}\\
    \hline
    Vinyals \textit{et al.} \cite{vinyals2015show} & RNN + CNN & Encoder CNN captures information in images, and decoder RNN generates neural language descriptions based on their features\\
    \hline
    Malinowski \textit{et al.} \cite{malinowski2015ask} & LSTM + CNN & CNN and LSTM respectively encodes the image and the question into vectors to capture the semantic information, and then another LSTM generates corresponding answers\\
    \hline
    Dai \textit{et al.} \cite{dai2017towards} & Conditional GAN (CGAN) & CNN captures information in an image, and LSTM generates the relevant descriptions; the discriminator evaluates the quality of generated descriptions\\
    \hline
    Das \textit{et al.} \cite{das2017visual}& Memory Network & Embedding the image, historical dialogue and given question respectively to consider the image and dialogue context information in conversation\\
    \hline
  \end{tabular}
 \end{table}

\textbf{\textit{Combining CNN and RNN.}} In order to capture the information in images and generate natural language text, combining CNN and RNN is the main solution for visual text generation. Vinyals \textit{et al.} \cite{vinyals2015show} achieve the goal of automatically viewing an image and generating the reasonable description utilizing the encoder-decoder structure. The encoder CNN captures information in the image, and the decoder RNN generates the text description. Due to the heavy loss of image information causing by the high-dimension structure of CNN, Xu \textit{et al.} \cite{xu2015show} propose an image caption model utilizing attention mechanism to extract the most important information in images to generate more accurate and detailed image description. Malinowski \textit{et al.} \cite{malinowski2015ask} combine CNN with LSTM to answer questions about the given image. They utilize the CNN to capture the related information in the image about questions, and an LSTM to generate answers based on latent representations of the image and question. Zhu \textit{et al.} \cite{zhu2016visual7w} build a semantic relationship between text descriptions and regions in the image by object-level grounding to generate answers of questions correspond with specific image regions.

\textbf{\textit{Memory Network-based models.}} Instead of simple single-round visual QA, Das \textit{et al.} \cite{das2017visual} implement a visual dialogue system to communicate with users in multiple rounds about a given image. They put forward the task of \textit{Visual Dialogue} and publish a large-scale Visual Dialogue dataset called VisDial\footnote{\url{https://visualdialog.org/}}. Three novel encoders are designed for the visual dialogue task, in which the \textit{Late Fusion} module encodes the image, historical dialogue and the given question respectively, the \textit{Hierarchical Recurrent Encoder} module encodes the dialogue history in the sentence level and the \textit{Memory Network} module stores the former QA pair as the ``fact'' to offer factual basis for the latter responses generation.

\textbf{\textit{GAN-based models.}} Different from the above works, Dai \textit{et al.} \cite{dai2017towards} leverage the conditional GAN (\textit{CGAN}) model to generate high-quality image descriptions in three aspects, including naturalness, semantic relevance, and diversity.  They jointly learn a generator to produce descriptions conditioned on images and an evaluator to assess how well a description fits the visual content with the criteria of natural and semantically relevant.

\textbf{\textit{VAE-based methods.}} In image captioning, it is important to ensure the lexical and syntactic diversity of generated captions. Chen \textit{et al.} \cite{chen2019variational}propose the Variational Multi-modal Inferring tree (\textit{VarMI-tree}) to model the lexical and syntactic diversities by inferring their latent variables in an approximate posterior inference guided by a visual semantic prior. Conditioned on the visual features and the latent variables, diverse captions of given images are generated. Previous works usually generate latent variables for entire input sentences, ignoring information about the substructures in the sentences. Aneja \textit{et al.} \cite{aneja2019sequential} develop the \textit{SeqCVAE} model, which learns a latent space for every word to capture the `intention' about how to complete the sentence. The data-dependent transition model captures the `intention', a representation of the remaining part of the sentence by encoding them with a backward LSTM, to generate more diverse captions.

\subsection{Multi-conditional text generation}
We have summarized existing works of CTG under a single condition, but in practice, these conditions often act simultaneously on the text generation system to produce more reasonable and anthropomorphic content. For example, in daily conversations, generation models usually consider the context information, personality characteristics of the interlocutor, and abundant external knowledge to generate reasonable responses. Therefore, considering the hybrid of different conditions is essential for improving the quality of CTG systems.

Emotion is an inherent attribute of natural language. Explicit emotion modeling and combining other conditions can improve the naturalness and humanness of the generated content. In order to realize emotional text generation, it is necessary to first detect the emotions contained in the textual content. Based on the observation that people usually express emotions rely on conversation contexts and external knowledge, Zhong \textit{et al.} \cite{zhong2019knowledge} propose to interpret the contextual utterances and leverage the external commonsense knowledge to enhance the emotion detection performance. The hierarchical self-attention and cross-attention modules are used to abstract contextual information and the context-aware affective graph attention is used to leverage knowledge to facilitate the understanding of context and emotion detection in conversations. Kao \textit{et al.} \cite{kao2019model} develop the chatbot that will change emotions according to the conversation context with users. Through the sentiment recognition model, the dialogue agent can provide the robot's emotions as feedback while talking with a user. Peng \textit{et al.} \cite{peng2019topic} believe that topics, like emotions, are important factors in dialogue systems, which can ensure the semantic coherence of the whole conversation. They develop the Twitter Latent Dirichlet Allocation (\textit{LDA}) model to detect the topic words of the input sentences as the prior knowledge, and the dynamic emotional attention mechanism to obtain the content and affective information related to the input texts and additional topics. In order to generate recommended reason text of specific items for users in recommender systems, Bai \textit{et al.} \cite{bai2020fusing} fuse aspect sentiment and external knowledge for recommended reasons generation. The fine-tuned BERT model is applied to get the aspect and aspect sentiment polarity from the reviews. The aspect fusion module fuses aspects and the item title, and the knowledge fusion module fuses relevant knowledge by the bi-directional self-attention module to generate personalized and content-rich recommended reasons. Zhong \textit{et al.} suggest that persona plays an important role in empathetic conversations, and first present a novel large-scale multi-domain dataset for persona-based empathetic conversations \cite{zhong2020towards}. Based on this dataset, they propose an efficient BERT-based response selection model, CoBERT, using multi-hop co-attention to learn higher-level interactive matching.

External knowledge can provide guidance for any CTG system to improve the system's understanding of conditions and generate more informative texts. For instance, Chen \textit{et al.} \cite{chen2019towards} introduce a Knowledge Based Personalized (\textit{KOBE}) product description generation model which fuses product aspects, user categories, and knowledge base to generate informative and personalized product descriptions. The self-attention modules in Transformer are used to encode the product attributes, the relevant knowledge, and the specific user categories into semantic vector representations, and perform deep semantic interaction to capture semantic features for the decoder. Yang \textit{et al.} \cite{yang2019enhancing} fuse external knowledge into topic-to-essay generation systems to provide background information for essay generation. The memory-augmented neural model selects knowledge concepts and then stores them into a memory matrix. The decoder then attends the memory to guide the text generation and updates it according to the decoder states to generate topic-consistent and informative essays.

The generation of natural language text is usually influenced by many factors, so the combination of multiple conditions is a promising research trend of CTG systems. By considering the appropriate context, combining accurate knowledge, expressing specific emotions, and conforming to unique personalized characteristics, text generation systems can generate more anthropomorphic texts.

\subsection{Pre-trained language model-based CTG}
The idea of pre-training has been widely explored in NLP. By pre-training the model on large-scale text corpus to initialize most of the network parameters which learn universal knowledge about syntactic and semantic information of neural language text, and fine-tuning the model using a small amount of specific downstream task data, excellent performance can be achieved. The pre-trained language models are first introduced into NLP for word embedding \cite{mikolov2013distributed}. Using a large amount of data to train the LSTM language model in an unsupervised way, the contextual word vector of each word can be obtained to demonstrate strong results across discriminative natural language understanding (\textit{NLU}) tasks \cite{mccann2017learned} \cite{peters2018deep}.

Recent pre-trained language models based on large Transformer architectures prove the ability of both big models and big data to improve language representation and generation performance. Transformer is gradually replacing the mainstream position of LSTM in NLP. Among numerous works, the BERT model \cite{kenton2019bert} and the GPT model \cite{radford2018improving} receive the most attention. BERT learns bidirectional representations of massive textual data by conditioning on both the forward and backward sequential contexts. Just adding a specific output layer rather than adjusting the model's structure, the pre-trained BERT model can be fine-tuned to achieve the state-of-the-art performance in many NLU tasks, such as text classification. Subsequently, a lot of work is done to optimize the pre-training process of BERT to further improve the ability of language representation, among which the most typical work includes XLnet \cite{yang2019xlnet}, RoBERTa \cite{liu2019roberta}, and ELECTRA \cite{clark2020electra}.

In terms of unconditional natural language generation, the GPT models are regarded as starting points for pre-trained natural language generation due to its form of standard language model for text generation tasks. GPT adopts the typical pre-training and fine-tuning training framework, with Transformer decoder as the feature extractor. In the pre-training stage, the training task is the unidirectional language model to encode language knowledge into the decoder, while in the fine-tuning stage, parameters of the pre-training model are fine-tuned according to specific tasks. Considering that supervised fine-tuning of the pre-training model with data of specific domains will reduce the generalization ability of the model, GPT-2 \cite{radford2019language} and GPT-3 \cite{brown2020language} remove the supervised fine-tuning stage of GPT and directly use massive training samples for zero-shot training. For all NLP tasks, GPT-3 with 175 billion parameters shows impressive performance without any gradient updating or fine-tuning.

Designing a pre-trained model for natural language generation tasks which often adopt Seq2seq frameworks with attention mechanism, is highly potential and critical. In addition to the GPT series that only contain Transformer decoders, researchers also explore the Seq2seq pre-training for unconditional text generation to jointly train encoders and decoders for better generation performance, including MASS \cite{song2019mass}, UniLM \cite{dong2019unified}, BART \cite{lewis2019bart}, T5 \cite{raffel2019exploring} and so on. For example, MASS combines the pre-training of encoder and decoder to reconstruct a sentence fragment, which masks a piece of tokens of input sentences randomly in the encoder and predicts the masked tokens in the decoder. The joint training process improves the ability of feature extraction and language modeling, which can achieve promising generative performance with zero-shot or few-shot fine-tuning on task-specific data. More and more researches show that pre-trained language models are of great value in text generation tasks.

After exploring the great potential of pre-trained language models for text generation, CTG powered by pre-trained language models has become a promising research direction. Based on the powerful generation ability of pre-trained language models, text generation models can be better at controlling the expression of conditions and generating more personified text under specific conditions. The most straightforward way to incorporate pre-trained language models to CTG systems is to modify the model architecture for extra conditional inputs or condition-specific fine-tuning. Mao \textit{et al.} \cite{mao2019improving} perform intermediate fine-tuning on the story data to adapt the pre-trained GPT-2 model to the domain of stories, and then fine-tune on the target story generation dataset with a multi-task learning objective to generate grammatical and stylistic consistency stories. An auxiliary training signal is used to provide common sense grounding for generated stories, which constrains the model to rank sensible text with lower perplexity. Keskar \textit{et al.} \cite{keskar2019ctrl} propose to add a mention, namely control codes specific to each type of text (\textit{e.g.} ``books'' for novel-type texts), to the input text and include one at the beginning of each text during the pre-training phase. By this means, the Conditional Transformer Language (\textit{CTRL}) model learns the relationship between the control codes and the text that follows to determine the generated text under the desired condition controlled by the specific code. Considering that large pre-trained transformer models are sensitive to large parameter changes during fine-tuning, Ziegler \textit{et al.} \cite{ziegler2019encoder} choose to adapt the pre-trained language model to arbitrary conditional inputs. The pseudo self-attention module learns a task-specific encoder which injects learned encoder conditioning directly into the pre-trained self-attention of the model. Because of the arbitrary length of the input sentence, these additional conditional inputs can be injected into the pre-trained model without changing the model architecture to affect the generated text. Chen \textit{et al.} \cite{chen2020distilling} leverage the idea of model distilling for better text generation performance. The Conditional Masked Language Modeling (\textit{C-MLM}) task is proposed to enable pre-trained BERT with additional conditional input by randomly masking tokens only in the target sequence. In the knowledge distillation stage, the generated sequences of word logits by the \textit{teacher} BERT model contains information from both backward and forward contexts, providing sequence-level global guidance. This probability distribution is soft targets for the \textit{student} text generation model to mimic, which contains more useful and fine-grained conditional information. To leverage the redundant external knowledge under capacity constraint, Zhao \textit{et al.} \cite{zhao2020knowledge} propose a pre-trained language model-based response generation model with a knowledge selection module, which formalize knowledge selection as a sequence prediction process. The knowledge selection and response generation module are jointly optimized with unlabeled dialogues to endow a generative model with both rich knowledge and good generalization ability.

Above CTG works based on pre-training language models need to adjust the model structure or fine-tune the model with the data under specific conditions, entailing the significant cost of retraining. The Plug and Play Language Model (\textit{PPLM}) \cite{dathathri2019plug} allows anyone to flexibly plug in one or more simple attribute models representing the desired control objective into a large, unconditional LM. Instead of training a large language model from scratch, PPLM trains smaller attribute models to influence the generated results of the existing ones, such as GPT-2. Attribute models are responsible for estimating the probability that a text sequence $x$ with a specific attribute $\alpha$ (\textit{e.g.} Positive or Negative). By maximizing the probability of the generated sequence $x$ with the desired attribute $\alpha$, the generated sentence has the pre-defined conditions. Considering that PPLM still requires large amounts of labeled texts to effectively balance generation fluency and proper conditioning, Carbone \textit{et al.} \cite{carbone2020etc} leverage topic models to enhance PPLM with an unlabeled collection of documents. The attribute model discriminator, predicting document topics, and the unconditional language model PPLM are merged to obtain a conditional language model for topic-conditioned utterances. To equip the dialogue model with multiple skills (\textit{e.g.}, emphatic response, weather information, etc.) without retraining the whole dialogue model, Madotto \textit{et al.} \cite{madotto2020adapter} propose the Adapter-Bot, which triggers different skills via different Adapters trained independently. The backbone of the Adapter-Bot is a pre-trained conversational model such as DialoGPT \cite{zhang2019dialogpt}, providing the ability of response generation. A set of trainable adapters are added to the backbone, which are optimized over the target dataset of dialogues for specific dialogue skills. Using the trained dialogue manager to select the right dialogue skill under the dialogue story, Adapter-Bot shows high-level control over the chatbot.

To generate text under more precise conditions in the word-level and phrase-level rather than just high-level conditions such as topic and sentiment, Chan \textit{et al.} \cite{chan2020cocon} propose the Content-Conditioner (\textit{CoCon}) model to control the generated text under the guide of target text content at a fine-grained level. The encoder and decoder of the model are pre-trained by GPT-2, and the CoCon block incorporates the target content into the encoded text representation before passing the content-conditioned representations into the decoder for generation. By self-supervised training without labeling data, CoCon can produce high-quality text with content. Zhang \textit{et al.} \cite{zhang2020pointer} explore the problem of generating text from a given set of lexical constraints. Given lexical constraints, the proposed \textit{POINTER} model generates high-level words (\textit{e.g.}, verbs and adjectives) as the keywords constrains, then inserts other words of finer granularity around the keywords iteratively until the whole sentence is generated. The training objective of POINTER is to generate a complete text sequence with a set of keywords as constraints, which is similar to the masked language modeling (\textit{MLM}) objective in BERT, so pre-trained BERT is used to initialize the model training to boost the generation performance.

The latent variables in the VAE model can capture higher-level sentence representations, such as topics, semantics, and patterns, to facilitate CTG. In order to combine VAE's powerful ability in the pre-training language model, Li \textit{et al.} \cite{li2020optimus} propose the OPTIMUS model, the first large-scale pre-trained deep latent variable models. OPTIMUS includes two stages: pre-training and fine-tuning. In the pre-training stage, the sentence-level (variational) auto-encoder objectives are trained on large text corpus to construct a universal latent space for reorganizing sentences. In the fine-tuning stage, by representing labeled specific tasks' data in the pre-trained latent space, OPTIMUS fine-tunes the latent space by updating all/part of the parameters to adapt to the downstream tasks. Specially, for the stylized response generation task, through embedding the history, response, and style-reference into the joint latent space to fine-tune OPTIMUS, the generated responses are closer to the desired text style.

With sufficient pre-training on huge text corpora, large pre-trained language models have shown impressive language understanding and language generation ability. Researches of CTG leveraging pre-trained language models have also attracted more and more attention, as summarized above. How to integrate conditional information into pre-training language models more effectively and reduce the cost of retraining is the important future research direction.

\subsection{Decoding Strategies for CTG}
We have made a detailed investigation of different CTG fields above, such as context-based text generation and topic-aware text generation. Meanwhile, the hybrid of different conditions and pre-trained language models applied in CTG systems are also been discussed. In summary, previous works focus on modifying the encoder, or the interaction mode between encoders and decoders to fuse conditional information into the text generation process. In the decoding stage, different decoding strategies can have a huge impact on the quality of the generated text. Researchers have proposed several decoding strategies to improve the quality of generated contents, \textit{e.g.}, reducing repetitive words or phrases. Through straightforwardly restricting the probability distribution of different words during the decoding stage under specific conditions, high-quality content with specific conditions can be generated by CTG models. In this section, we will introduce several universal decoding strategies, such as Beam Search and Top-k Sampling, and then summarize some improved decoding strategies proposed for CTG, including Weighted Decoding and Unlikelihood Training.

\textbf{\textit{Beam Search.}}
Due to the large number of words in the vocabulary, the number of possible sequences in generation is enormous, so researchers propose some heuristics to reduce the searching space thus making the generation practical. Beam Search is a most commonly used decoding strategy in text generation models recently. It approximately maximizes the likelihood of the whole generated sequence given a hyper-parameter $\beta$, known as beam size. In the first decoding time step, $\beta$ words with the highest conditional probability are selected as the first words of the candidate output sequence. For each subsequent time step, additional words will be attached to $\beta$ output sequences of the last time step, in which the $\beta$ sequences with the highest conditional probability will be selected. The optimal generated sequences are determined from the final $\beta$ candidates with the highest conditional probability. Beam Search greatly reduces the time and space requirements for searching in the text generation process by limiting the beam size. However, since only $\beta$ sequences are selected at each time step, the final generated content may not be optimal.

To overcome shortcomings of Beam Search, many improved methods have been proposed. For example, Vijayakumar \textit{et al.} \cite{vijayakumar2016diverse} propose the Group diverse Beam Search to increase the variety of generated texts, which divides the beam into groups and utilizes a group dissimilarity penalty to reduce the similarity between different search groups. Similarly, Li \textit{et al.} \cite{li2016simple} propose the Sibling diverse Beam Search that contains a penalty proportional to the rank of a candidate token to encourage preserving hypotheses from diverse sources within the beam.

\textbf{\textit{Top-k and Top-p sampling.}}
The goal of sampling-based decoding strategies is to reduce repetitions and increase the diversity of the generated content, by utilizing stochastic decisions in the generation process. The Top-k sampling \cite{fan2018hierarchical} samples from the next-token distribution after having filtered this distribution to keep only the top $k$ tokens, while the Top-p sampling \cite{holtzman2019curious}, also known as nucleus sampling, samples from the top tokens with a cumulative probability just above a threshold $p$.

The above decoding strategies are aiming at making the distribution of generated words similar to the distributions of words in human-generated texts at a higher level. In the meanwhile, many studies focus on decoding strategies serving for specific conditions, to make CTG systems more effectively fuse the conditional information, which will be discussed below.

\textbf{\textit{Weighted Decoding.}}
Weighted Decoding \cite{ghazvininejad2017hafez} is a typical decoding strategy that increases or decreases the probability of words under certain conditions. Ghazvininejad \textit{et al.} \cite{ghazvininejad2017hafez} propose an interactive poetry generation system which enables users to edit and polish generated poems by adjusting from different aspects (\textit{e.g.},sentiment, alliteration, etc.). In the $t^{th}$ decoding step, a partial hypothesis $y_{<t}=y_{1},...,y_{t-1}$ is expanded by calculating the generation score for next word $w$:

\begin{equation}
score(w,y_{<t};x)=score(y_{<t};x)+logP_{RNN}(w|y_{<t},x)+\sum_{i}w_{i}*f_{i}(w;y_{<t};x)
\end{equation}

$logP_{RNN}(w|y_{<t})$ represents the generative log-probability of the word $w$ and $score(y_{<t};x)$ is the accumulated generation score of all generated words. The $f_{i}(w;y_{<t};x)$ refers to decoding features with corresponding weights $w_{i}$. A decoding feature assigns a real value to the generation probability of word $w$, to control the weights of generated words under different conditions. Weighted Decoding considers 8 types of features to control the generated content, including topical words, emotions, and so on. Instead of manually designing calculation formulas of decoding features as additional items of the decoding objective function, Holtzman \textit{et al.} \cite{holtzman2018learning} train a number of discriminative models to construct a more powerful generator under different aspects of conditions, each of which encodes an aspect of high-quality generation to enhance the generation performance of the original RNN generator. The decoding objective function is formalized as follows:

\begin{equation}
f_{\lambda}(x,y)=log(P_{lm}(y|x))+\sum_{k}\lambda _{k}s_{k}(x,y)
\end{equation}

This objective is composed of the traditional RNN language model probability $log(P_{lm}(y|x))$ and additional scores $s_{k}(x,y)$ calculated by designed discriminators with learned mixture coefficients $\lambda _{k}$. Four discriminators are proposed to discriminate between good and bad generations (\textit{e.g.}, Repetition Model for avoiding word repetitions, Relevance Model for guaranteeing contextual relevance of the generated content, etc.) and are interpolated in the objective function as log probabilities. The weight coefficient of each discriminator is optimized to minimize the difference between the scores assigned to the gold continuation and the continuation predicted by the current model. Similarly, Baheti \textit{et al.} \cite{baheti2018generating} incorporate topic and semantic similarity constraints into the decoding objective of dialogue systems to encourage the generation of more topic-relevant and content words in responses. In order to match the distribution over topics in generated responses and input, the \textit{HMM-LDA} model is applied to estimate topic distributions over words in responses and inputs so as to compute their topical similarity. The semantic similarity constraints are designed to encourage generated responses to be semantically similar to inputs.

Weighted Decoding is a useful technique in conditional text generation, which can force the expected conditional features to appear in the generated text by assigning them a high generation probability. However, when the weight of the specific feature is too large, Weighted Decoding risks going off-distribution, thus generating unanticipated words \cite{see2019makes}.

\textbf{\textit{Unlikelihood Training.}}
The standard approach of training neural text generation models is to maximize log-likelihood and approximately decoding the most likely sequence, which is known to have fatal defects. The likelihood training will force the model to generate common words with high frequency, making the generated text dull, and to repeat themselves at word and sentence level. In order to solve these problems, Welleck \textit{et al.} \cite{welleck2019neural} optimize the \textit{unlikelihood training} objective for training text generation models more efficiently. The key idea of unlikelihood training is to decrease the generation probability of certain tokens, making incorrect repeating tokens less likely and frequent tokens less likely, thus forcing repetitions to have low probability and improving the quality of generated text.

Considering that existing dialogue systems tend to generate frequent words and repetition words, Li \textit{et al.} \cite{li2019don} first incorporate unlikelihood training into dialogue systems to regularize generated outputs to match human-written distributions. Three different biases needed to be mitigated are considered, including repetition and copying, vocabulary usage, and contradictions. The first two biases are mainly responsible for reducing the occurrence of repeated words and high-frequency words, and the bias of contradictions is designed to assign low probability to inconsistent and contradictory utterances. Through dividing training samples into positive and negative coherent behavior, the likelihood training is performed on coherent data, and the unlikelihood objective is applied to the incoherent data to reduce the probability of generating the context incoherent response.

Through unlikelihood training, text generation models can force unlikely generations to be assigned lower probabilities, such as repetitive and dull text generation, to improve the overall quality of generated sentences. In summary, unlikelihood training has great potential for researches of CTG systems. By controlling the probability of specific conditional features expressed in the generated content, the efficient fusion of conditional information, and higher-quality conditional text generation can be achieved.

\subsection{Conditional datasets}
The training of CTG models needs the support of a large number of conditional text data, such as emotional text data or personalized text data, but this kind of data is relatively scarce. In order to protect researchers from data scarcity, many high-quality conditional text datasets have been released. We give a brief summary of conditional text datasets in Table~\ref{tab:datasets}.

\begin{table}
 \caption{A review of conditional text datasets}
 \label{tab:datasets}
 \begin{tabular}{|m{4.3cm}<{\centering}|m{2.7cm}<{\centering}|m{6cm}<{\centering}|}
    \hline
    \textbf{Dataset} & \textbf{Type} & \textbf{Description}\\
    \hline
    Cornell Movie Dialogs \cite{danescu2011chameleons} & Context-based dataset &  A multi-turn dialogue dataset extracted from raw movie scripts\\
    \hline
    Ubuntu Dialogue Corpus \cite{lowe2015ubuntu} & Context-based dataset & A multi-turn dialogue dataset extracted from the Ubuntu chat logs\\
    \hline
    PERSONA CHAT \cite{zhang2018personalizing} & Personalized dataset & A personalized dialogue dataset where two parts of every conversation are given a group of profile information\\
    \hline
    Humeau \textit{et al.} \cite{mazare2018training}& Personalized dataset & A profile-based dialogue dataset; Extracting personalized characteristics from users' posts in REDDIT\\
    \hline
    TaoDescribe \cite{chen2019towards} & Personalized dataset & A personalized product description dataset including product basic information, in which each pair is labeled with knowledge and user category attributes\\
    \hline
    Empathetic Dialogues \cite{rashkin2019towards} & Emotional dataset & An emotional dialogue dataset where each conversation is under a given emotion\\
    \hline
    EmotionLines \cite{chen2019emotionlines} & Emotional dataset & An emotional dialogue dataset in which all utterances are labeled with emotions\\
    \hline
    CMU DoG \cite{zhou2018dataset} & Knowledge-based dataset & A document grounded conversation dataset where each conversation are about contents of a specified document\\
    \hline
    Wizard of Wikipedia \cite{dinan2018wizard} & Knowledge-based dataset & A knowledge-grounded dataset with conversations directly grounded with knowledge retrieved from Wikipedia\\
    \hline
    Zhou \textit{et al.} \cite{zhou2018commonsense} & Knowledge-based dataset & A commonsense conversation dataset containing one-turn post-response pairs with corresponding commonsense knowledge graphs\\
    \hline
    VisDial \cite{das2017visual} & Visual dataset & A Visual Dialogue dataset where all queries and answers are based on the given image\\
    \hline
  \end{tabular}
 \end{table}

\textbf{\textit{Context-based datasets.}} Cornell Movie Dialogs\footnote{\url{http://www.cs.cornell.edu/~cristian/Cornell_Movie-Dialogs_Corpus.html}} \cite{danescu2011chameleons} is a large-scale multi-turn dialogue dataset providing contextual information during the conversation, which contains a large metadata-rich collection of fictional conversations extracted from raw movie scripts. Ubuntu Dialogue Corpus\footnote{\url{https://github.com/rkadlec/ubuntu-ranking-dataset-creator}} \cite{lowe2015ubuntu} is another large multi-turn dialogue dataset containing almost one million conversations extracted from the Ubuntu chat logs that is a great help for training context-sensitive technical dialogue systems.

\textbf{\textit{Personalized datasets.}} Zhang \textit{et al.} \cite{zhang2018personalizing} present a high-quality personalized dialogue dataset named PERSONA CHAT\footnote{\url{https://github.com/facebookresearch/ParlAI/tree/master/projects/personachat}}. In each dialogue, two parts of the conversation are given a group of profile information, and the whole dialogue process is conducted around these personalized characteristics. Humeau \textit{et al.} \cite{mazare2018training} build an authoritative profile-based dialogue dataset using conversations collecting from REDDIT\footnote{\url{https://www.reddit.com/r/datasets/comments/3bxlg7/}}. The personalized characteristics are extracted from users' social posts, providing a new opportunity of personalized text generation for later researchers. Chen \textit{et al.} \cite{chen2019towards} provide a personalized and knowledge-based product description dataset named TaoDescribe, collecting from Taobao\footnote{\url{https://www.taobao.com/}}, a large Chinese shopping website.  There are more than two million pairs basic information and descriptions of products , in which each pair is labeled with knowledge and user category attributes.

\textbf{\textit{Emotional datasets.}} Rashkin \textit{et al.} \cite{rashkin2019towards} publish an emotional dialogue dataset called \textit{Empathetic Dialogues}\footnote{\url{https://github.com/facebookresearch/EmpatheticDialogues}}, including an extensive set of emotions and every speaker in it feels with a given emotion during conversations. Chen \textit{et al.} \cite{chen2019emotionlines} publish another high-quality emotional dialogue dataset collecting from telescripts and dialogues in Facebook, named \textit{EmotionLines}\footnote{\url{http://doraemon.iis.sinica.edu.tw/emotionlines/index.html}}. All utterances in it are labeled with specific emotion according to textual contents to guide the emotional dialogue response generation.

\textbf{\textit{Knowledge-based datasets.}} CMU DoG\footnote{\url{https://github.com/festvox/datasets-CMU DoG}} \cite{zhou2018dataset} is a document grounded conversation dataset where each conversation is followed by specified documents about popular movies extracted from Wikipedia articles. Wizard of Wikipedia\footnote{\url{https://parl.ai/projects/wizard_of_wikipedia/}} \cite{dinan2018wizard} is another open-domain dataset whose conversations are directly grounded with knowledge retrieved from Wikipedia. Zhou \textit{et al.} \cite{zhou2018commonsense} present a commonsense conversation dataset\footnote{\url{http://coai.cs.tsinghua.
edu.cn/hml/dataset/\#commonsense}} containing one-turn post-response pairs with the corresponding commonsense knowledge graphs. Each pair in it is associated with some knowledge graphs retrieved from ConceptNet\footnote{\url{https://conceptnet.io}}, a typical structured knowledge graph.

\textbf{\textit{Visual datasets.}} VisDial\footnote{\url{https://visualdialog.org/data}} \cite{das2017visual}
is a large-scale visual dialogue dataset where all queries and answers are based on the given image. Researchers can utilize it to research the visual text generation.

\subsection{Evaluation Methods Towards CTG}
We have given a detailed investigation of different CTG fields and summarized some commonly used conditional text datasets. Another key issue faced by researchers is how to evaluate the performance of these models and make fair and meaningful comparisons among them. Only with the help of reasonable evaluation metrics, can researchers accurately evaluate the performance of the designed models, to draw correct conclusions and promote applications of these models in real life. Natural language generation technology has made great progress in recent years, but the reasonable evaluation of the generated text is still a challenging problem to be solved. At present, researchers have not formed a unified theory on how to effectively evaluate text generation systems \cite{van2019best}, and the lack of reasonable quantitative evaluation metrics will prevent the development of this field. There are mainly two kinds of evaluation metrics at the present stage, namely automated evaluation metrics and human evaluation metrics, which will be summarized below.

\subsubsection{Automatic Evaluation Metrics}
Automated machine evaluation is an intuitive and convenient method to evaluate the quality of text generation systems. The quality of the generated text is determined by uniquely designed formulas that compare the difference between the generated text and the ground-truth text. Among various evaluation metrics of text generation, the \textit{n}-gram based metrics are the most widely studied which calculate the word overlap under the \textit{n}-gram language unit between the generated text and ground-truth text. They are first applied in machine translation systems by calculating the degree of word overlap between the translated text and the target human-written references and then are widely used in the evaluation of many text generation systems. Typical evaluation metrics based on \textit{n}-gram are summarized below.

\textbf{\textit{BLEU.}} BLEU \cite{papineni2002bleu} is the harmonic mean of \textit{n}-gram precisions of the generated texts with respect to ground-truth reference sentences, where $n\epsilon \{1,2,3,4\}$. The \textit{n}-gram precisions refer to the proportion of the generated text that matches any \textit{n}-gram unit in the reference sentences. Repeated \textit{n}-gram matches are clipped to the maximum number of times the \textit{n}-gram occurs in any single reference. BLEU contains many variants, such as SentBLEU \cite{lin2004orange}, $\Delta$ BLEU \cite{galley2015deltableu}, NIST \cite{doddington2002automatic}, and so on. NIST is a typical variant of BLEU, which improves BLEU's evaluation accuracy by assigning higher weights to low-frequency \textit{n}-gram (\textit{e.g.}, more informative) and imposing length penalties.

\textbf{\textit{ROUGE-L.}} Rouge-L \cite{lin2004looking} is calculated based on the length of the longest common subsequences (\textit{LCS}) between the generated texts and the target texts, where the common subsequence needs to include the same words in the same order. Then the \textit{F}-measure is calculated based on the maximum precision and recall of reference texts to obtain the final ROUGE-L score, where the accuracy and recall are calculated by dividing the length of \textit{LCS} by the length of the generated text and the reference text respectively.

\textbf{\textit{METEOR.}} METEOR \cite{banerjee2005meteor} calculates the precision and recall of unigrams between generated texts and reference texts. In addition to the exact word matching, fuzzy matching is adopted in the calculation process based on the stem analysis and WordNet synonym. The matching degree is calculated with multiple reference texts and the best-matching one is selected as the final METEOR score.

\textbf{\textit{CIDEr.}} CIDEr \cite{vedantam2015cider} is an evaluation metric designed for the image captioning task and can also be used for other text generation tasks. The CIDEr score is calculated by the average cosine similarity between the generated texts and the reference texts on the level of \textit{n}-grams, where the importance of individual \textit{n}-grams is calculated by the Term Frequency Inverse Document Frequency (\textit{TF-IDF}) measure.

The above evaluation metrics based on \textit{n}-gram have been widely used in various text generation tasks, including machine translation, dialogue system, text summarization, etc. However, numerous studies have shown that the \textit{n}-gram matching cannot capture semantic information and effectively evaluate text generation models. In some specific applications with little variation, such as machine translation and question answering, metrics based on word-overlap of \textit{n}-grams have a higher evaluation accuracy. However, when reference texts have high diversity, such as dialogue systems and text summarization systems, it is difficult to make effective evaluation using these metrics.

In addition to evaluating the relationship between the generated text and the reference text, the characteristics of the natural language itself can also be used to evaluate the quality of text generation systems.

\textbf{\textit{Perplexity.}} Perplexity \cite{jelinek1977perplexity} is a metric used to evaluate the quality of language models, which measures the average number of uncertain words when predicting words. The smaller the number is, the better the language model performance is. The fluency and diversity of generated text can be evaluated based on perplexity.

\subsubsection{Human Evaluation Metrics}
Most of the automatic evaluation metrics can only measure the quality of text generation models based on the similarity degree between generated texts and reference texts, but they cannot reflect the correctness, informativeness, naturalness, and other internal characteristics of the generated content. Therefore, human intelligence is introduced into the evaluation of text generation systems to provide more reasonable and effective evaluation. Human evaluation takes the form of the Turing test, which makes humans determine whether the quality of the machine-generated text is high enough to distinguish it from real data.

Human evaluation is done either on the overall quality of the generated text or at the finer-grained level, such as fluency, naturalness, informativeness, persona consistency, etc. Because the conditions considered in CTG systems (such as topics, knowledge, and personalized features) are difficult to evaluate using automatic evaluation metrics, human evaluation is almost the only and most effective way to evaluate CTG systems at the current research stage. For example, in knowledge-enhanced text generation systems, informativeness is a metric to measure whether the system effectively combines external knowledge. Only when external knowledge is correctly and reasonably integrated into the generation process, can more information-rich texts be generated. In personalized text generation systems, the persona consistency metric is used to evaluate whether the generated text conforms to the persona characteristics assigned to the text generation agent.

Human evaluation is the most effective gold standard for evaluating CTG systems. However, due to the subjectivity of human, there is often a high degree of variation in the process of human evaluation \cite{van2019best}. In addition, human evaluation is usually expensive, which requires a large amount of manpower to implement the evaluation process, and is not repeatable and costly. How to design reasonable evaluation metrics to make the evaluation of CTG systems more reasonable is still a challenging problem.

\section{GENERAL LEARNING MODELS FOR CTG}
We make a detailed investigation of various CTG fields under different conditions in above sections, which shows that combining different conditions can make the generated text more appropriate, informative and anthropomorphic. In this section, we attempt to address the CTG by distilling and presenting three different schemas of conditional generation models, including explicit conditional modeling, implicit conditional modeling and conditional knowledge transferring.

\begin{figure}[htbp]
    \centering
    \includegraphics[scale=0.3]{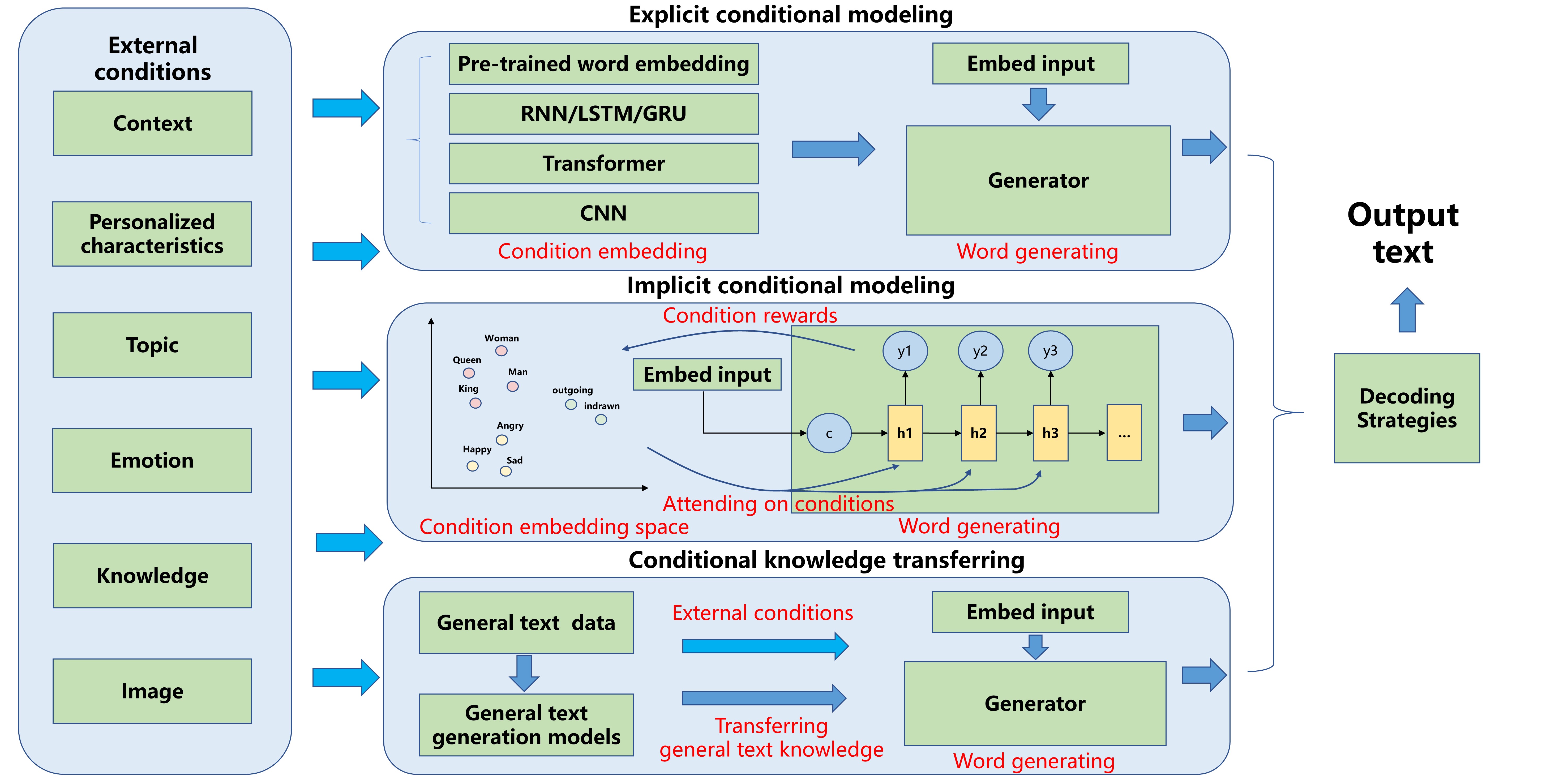}
    \caption{General learning models for CTG}
    \label{img:general}
\end{figure}

The explicit conditional modeling directly regards external conditions as a part of input, allowing the generator to process the condition information in the same way as the input information. This method is simple and efficient, which makes CTG systems easy to fuse the information of conditions from a global perspective, so as to generate results that are more consistent with external conditions. The implicit conditional modeling does not directly take conditions as input, but utilize specific algorithms, such as attention mechanism and RL, in the word generation stage to mine the implicit and deep semantic information in conditions, so as to make CTG systems more sensitive to conditions. Considering that the lack of conditional data makes it hard to fully train deep learning models, the conditional knowledge transferring extracts
general text knowledge from a large amount of text data and transfers it to CTG systems to improve the performance of CTG systems. The three general learning models mainly focus on how to fuse conditional information into the language generation process. Meanwhile, different decoding strategies can be applied to CTG systems to further control the appearance of certain conditions. Three general learning models for CTG are shown in figure~\ref{img:general}.

\subsection{Explicit conditional modeling}
In CTG systems, different conditions are in the form of natural language text (except for the image in the visual text generation), so taking them explicitly as a part of the input of the text generation system to enhance the input information is a straightforward method, known as explicit conditional modeling. In the same way that text generation systems process input, the condition will be transformed into vector representations by word embedding methods when the condition is a single word, or encoded by an additional neural networks after word embedding when the condition is a text sequence. Then vector representations of conditions will be used as the additional input information in the decoding stage to guide the text generation. Explicit conditional modeling is a simple, straightforward and proven method to integrate external conditions in CTG systems, without increasing the complexity of the model.

For example, in context-based text generation systems, Sordoni \textit{et al.} \cite{sordoni2015neural} regard the dialogue history as the context information and embed all words and phrases in the dialogue history into vector representations which are then decoded by another RNN to promote context-aware responses. In personalized text generation systems, Li \textit{et al.} \cite{li2016persona} propose persona-based models to solve the problem of speaker consistency in dialogue systems. The \textit{Speaker Model} maps each individual speaker into a vector by encoding speaker-specific information, such as age, gender, and dialect, that may influence his/her speaking content and style. Then speaker vectors are injected into decoder hidden layers to generate personalized dialogue responses. For knowledge-enhanced text generation systems, Ghazvininejad \textit{et al.} \cite{ghazvininejad2018knowledge} propose a knowledge-grounded dialogue model to produce contenful and informative dialogue responses. Given the dialogue history, the relevant knowledge facts are extracted from knowledge base using words in the dialogue history as keywords. Then the \textit{Facts Encoder} module adopts Memory Network to encode facts into vector representations based the user input and dialogue history. This module enables the dialogue system to deeper exploit inter lexical dependencies between different parts of facts and the input for effectively fusing knowledge into generation process. Then both the encoded dialogue history and knowledge facts are fed into the decoder to generate context relevant and informative responses.

\subsection{Implicit conditional modeling}
In order to capture the deeper semantic information contained in the condition, the condition information is not directly fed into the decoder after embedding, but interact with the decoding state at a deeper level using additional algorithms, known as implicit conditional modeling. The attention mechanism is a typical implicit condition modeling method, which can dynamically capture the important part of the condition according to decoder states, so as to realize implicit condition modeling and make the condition information guide the text generation process more effectively. The RL mechanism can give different rewards to actions of different states so as to give feedback on whether specific conditions are reflected in the generated results. Therefore, RL is suitable for implicit conditional modeling. By penalizing results that do not reflect conditional information, the RL mechanism forces the model to consider more conditional information in generation.

For example, Zheng \textit{et al.} \cite{zheng2019personalized} propose the persona aware attention mechanism to control the attention weights of context vectors under integrated persona vectors and the persona-aware bias to estimate the generation distribution for personalized dialogue responses generation. Zhou \textit{et al.} \cite{zhou2018commonsense} produce two graph attention mechanisms, \textit{static graph attention} and \textit{dynamic graph attention} respectively, to promote dialogue understanding and knowledgeable responses generating. The model extracts relevant knowledge graphs using the entities in the input as keywords and then encode graphs into static vector representations by the \textit{static graph attention} module, which considers all nodes and relations between nodes in a graph to encode more structural semantic information. In the decoder stage, the \textit{dynamic graph attention} attends the knowledge graphs and knowledge triples in each graph to efficiently integrate information in knowledge graphs for informative response generation.

Li \textit{et al.} \cite{li2019reinforcement} propose a RL-based dialogue model combined with an emotional editor module to generate customizable emotional responses. The emotional editor selects the template sentence according to the given topic and emotion to provide references for the generation process. The RL mechanism constrains the quality of generated responses from three points: emotion, topic and coherence, to generate emotional, topic-relevant and meaningful dialogue responses. The multi-task learning is also introduced to enhance the model discrimination to learn the coherence, topic, and emotion of a reply.

\subsection{Conditional knowledge transferring}
Compared with general text generation, CTG lacks available conditional text datasets. Although researchers have released several conditional text datasets, it is generally not enough to train a well-performing CTG system. By extracting general text knowledge from a large amount of text data and transferring it to CTG systems, well-performing CTG systems can be trained in the absence of conditional data, known as conditional knowledge transferring.

For personalized text generation, Luan \textit{et al.} \cite{luan2017multi} train a dialogue model to predict responses given previous contexts and an autoencoder model with large volumes of non-conversational personal data to model the role-specific characteristics of different users. Through the multi-task learning mechanism which shares the decoder parameters of the two models, these models can capture speaker roles, expressive styles and domain expertise characteristic of the targeted user and generate personalized responses without heavy recourse to each speaker's conversational data. The work of Yang \textit{et al.} \cite{yang2017personalized} utilizes domain adaptation mechanism to generate personalized dialogue responses. The general response generation model with an attention LSTM encoder-decoder architecture is pre-trained on the large-scale general dialogue data without user-specific information. Then the model is fine-tuned with a small amount of personalized dialogue data by a dual learning mechanism to generate personalized responses. In the training process, three rewards are proposed to evaluate the quality of generation results, that are personalization, informativeness, and grammaticality and the policy gradient method is adopted to generate highly rewarded responses.

\section{OPEN ISSUES AND FUTURE TRENDS}
Although many advanced technologies have been applied to text generation and some remarkable achievements have been made, there are still many remaining issues. In this section, we put forward some key issues and point out some future development trends of text generation.

\subsection{Different Types of Contexts}
Context information can provide the current situation, state, environment and other information for text generation systems to realize more accurate simulation of human expression, so it is very important in CTG systems. For example, in multi-turn dialogue systems, context information usually refers to historical dialogues and can make the conversation more coherent. Sordoni \textit{et al.} \cite{sordoni2015neural} encode the dialogue history into vector representations and feed them into the decoder to generate consistent responses. In machine translation systems,  Voita \textit{et al.} \cite{voita2018context} propose a context-aware machine translation model, which can control and analyze the flow of information from context to the translation model. By encoding the context information, the model can more accurately generate the correct translation text. These works achieve relatively excellent results. However, context information contains much more than those considered in existing studies. Existing researches only consider the dialogue history, predefined external information or other types context, but our expression may be affected by various factors, such as hot events, emotions, time, weather, etc. These external context information may be explicit or implicit, so how to extract them and properly represent them is a major challenge. A humanoid CTG system should be able to effectively integrate various contexts and generate reasonable text, which will be the future research direction.

\subsection{Multi-modal Data Translation and Domain Adaptation}
In addition to text data, there are various types of data, such as voice, image and video. Human can efficiently extract useful information from various types of data and convert them into corresponding text representation, such as describing the content of a painting and summarizing the content of a movie. Researchers carry out many works on text generation with multi-modal data as inputs, such as generating description/caption for a given image \cite{lu2017knowing}, conducting QA with images \cite{song2018pixels}, and communicating based on the content of a given image \cite{das2017visual}. These studies usually utilize CNN to extract relevant information in images, and generate corresponding text using common models in text generation. The existing deep learning models can only process one type of data, so handling multi-modal data requires to combine multiple models, which may bring a large computational cost. Moreover, the semantic space of different types of data may be different, which brings great difficulties for the fusion of multi-modal data. How to incorporate different types of data and develop unified models for multi-source data processing are two huge challenges. Much more efforts should be conducted on generating informative texts with multi-modal data sources.

At the same time, in many tasks of CTG, such as personalized text generation and emotional text generation, the available training data is very scarce. Most text data are totally general data and do not contain personalized or emotional characteristics, which cannot meet the requirements under specific conditions. Transfer learning is a promising way to address this problem. By learning general knowledge of natural language from massive common text data, and then transferring it to a specific domain training with a small-scale conditional text data, the model can not only master the general knowledge of the source domain, but also learn the specific needs of the target domain, to make up for the scarcity of data. Yang \textit{et al.} \cite{yang2017personalized} use the idea of domain adaptation in transfer learning to address the issue of lacking personalized dialogue data. Through fine-tuning the general dialogue model with the small size personalized dialogue data, the model can effectively generate personalized dialogue responses. Transfer learning is a rapidly developing technology in deep learning, and integrating diverse transfer learning models with scarce usable data for CTG is also a promising research direction.

\subsection{Long Text Generation}
Long text has a wide range of application areas, including writing compositions, translating articles, writing reports, etc. However, the current technology has some bottlenecks in processing long text because of the long-distance dependence existing in the natural language. Humans have the ability to extract the key information (\textit{e.g.}, contexts and topics) from long text, which is difficult for machines. Researchers have conducted much efforts on improving the models' ability of generating long text. For example, the LSTM and GRU model is produced to address the issue that the original RNN model cannot capture the long-distance dependence using the gating mechanism. Guo \textit{et al.} \cite{guo2018long} propose the \textit{LeakGAN} model to generate long texts. The feature extracted by discriminator is used as a stepwise guidance signal to guide the generator to produce high-quality text. At the same time, the hierarchical reinforcement learning provides more information to the discriminator for generating long text. How to effectively model and capture the long-term dependency in long text is a major challenge in the research. For text generation technology truly behaving like humans, it needs the ability to freely generate long or short texts, which still has much to be investigated.

\subsection{Lifelong learning}
Lifelong learning is an important ability of human beings. We continuously learn new knowledge, expand and update our knowledge base through various data sources in the physical world to adapt to the rapidly changing environment. Existing text generation models are usually trained on fixed datasets and have no ability to expand and dynamically updated according to the changes of the external environment. In order to make text generation models more personified, lifelong learning is a necessary ability. Combining external knowledge base is the first step to realize lifelong learning. There have been many text generation researches, focusing on the dialogue systems \cite{zhou2018commonsense} \cite{dinan2018wizard}, combined with knowledge bases. However, most of them are based on fixed knowledge base in which the knowledge does not keep updating in real time, so the model still does not have the ability of continuous learning.  Therefore, the dynamic evolution of knowledge base is very important. Mazumder \textit{et al.} \cite{mazumder2018towards} propose a lifelong learning model which can update its own knowledge base by asking users. This is a groundbreaking exploration which provides a direction for the lifelong learning text generation models. However, the ability to actively gain knowledge from the environment rather than simply asking the user is very important. How to find and learn the most effective information from the changeable external environment and achieve efficient lifelong learning is a very important research direction in CTG.

\subsection{Knowledge extraction and fusion from crowdsourced data}

With the rapid development and popularization of social networks, more and more crowdsourced data appear on the Web, such as the Q\&A community Quora\footnote{\url{https://www.quora.com/}} and Zhihu\footnote{\url{https://www.zhihu.com/}}. These data are the embodiment of human intelligence and can be used as the source of external knowledge of text generation systems. However, existing knowledge-enhanced text generation systems are mainly based on the structured knowledge graph or unstructured knowledge base built in advance \cite{li2019incremental}\cite{moussallem2019augmenting}, which cannot perform real-time knowledge selection and fusion from crowdsourced data and usually cover several specific domains. Crowd intelligence data covers a wide range of domains and dynamically updates itself in real time. However, the noise and heterogeneity of crowdsourced data prevent its applications in text generation systems. Therefore, how to mine suitable information from crowdsourced intelligence data in real time, enhance the semantic understanding, and fuse information properly in text generation, are promising research directions of CTG.

\begin{figure}[htbp]
    \centering
    \includegraphics[scale=0.4]{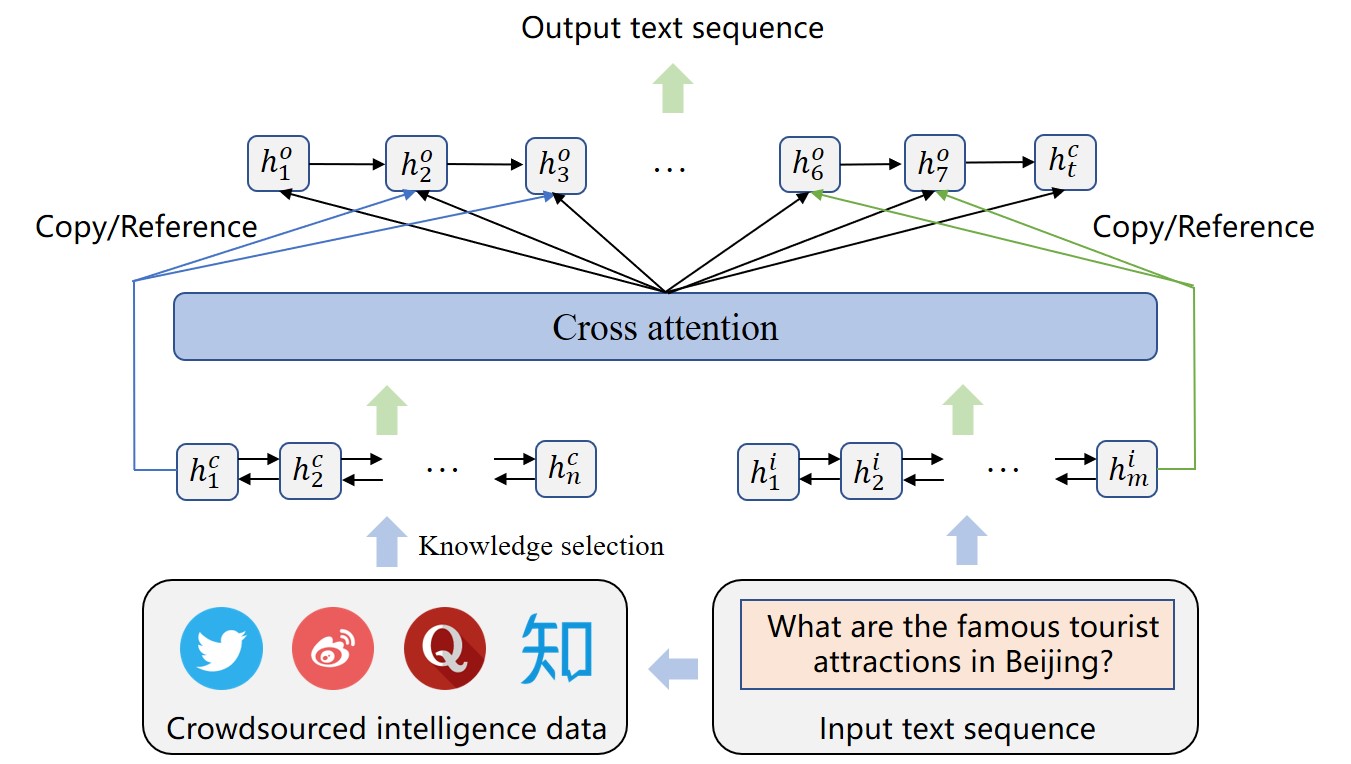}
    \caption{The concept map of our proposed model}
    \label{img:algo}
\end{figure}

To solve the problem of incorporating crowdsourced intelligence data in CTG, we have proposed a preliminary idea, as shown in figure~\ref{img:algo}. In the encoding stage, the input text will interact with the crowdsourced data and select relevant knowledge as the external knowledge source. Then the knowledge and the input text are encoded separately, and the cross-attention mechanism is used to obtain fusion vector representations of the input information. In the decoding stage, generation and copy/reference mechanisms are considered at the same time. The decoder will dynamically select whether to generate words from the vocabulary or copy words from input sources according to the current decoding state, to achieve more explicit use of crowdsourced knowledge.

\section{Conclusion}
We have made a systematic review of the research trends of conditional text generation (CTG). In this paper, we first give an introduction to the field of text generation. We then give a brief review of key techniques in the field of text generation and further give the formal definitions of different fields of CTG. Finally, we investigate the research status of various CTG fields and propose several general learning models for CTG. Though there has been a big research progress in CTG, it is still at the early stage and numerous open research issues and promising research directions should be studied, such as long text generation, multimodal data translation, and lifelong learning.

\begin{acks}
This work was partially supported by the The National Science Fund for Distinguished Young Scholars (62025205), and the National Natural Science Foundation of China (No. 61772428, 61725205)
\end{acks}

\bibliographystyle{ACM-Reference-Format}
\bibliography{CTG_Survey_Ref}


\begin{thebibliography}{167}


\ifx \showCODEN    \undefined \def \showCODEN     #1{\unskip}     \fi
\ifx \showDOI      \undefined \def \showDOI       #1{#1}\fi
\ifx \showISBNx    \undefined \def \showISBNx     #1{\unskip}     \fi
\ifx \showISBNxiii \undefined \def \showISBNxiii  #1{\unskip}     \fi
\ifx \showISSN     \undefined \def \showISSN      #1{\unskip}     \fi
\ifx \showLCCN     \undefined \def \showLCCN      #1{\unskip}     \fi
\ifx \shownote     \undefined \def \shownote      #1{#1}          \fi
\ifx \showarticletitle \undefined \def \showarticletitle #1{#1}   \fi
\ifx \showURL      \undefined \def \showURL       {\relax}        \fi
\providecommand\bibfield[2]{#2}
\providecommand\bibinfo[2]{#2}
\providecommand\natexlab[1]{#1}
\providecommand\showeprint[2][]{arXiv:#2}

\bibitem[\protect\citeauthoryear{Aneja, Agrawal, Batra, and Schwing}{Aneja
  et~al\mbox{.}}{2019}]%
        {aneja2019sequential}
\bibfield{author}{\bibinfo{person}{Jyoti Aneja}, \bibinfo{person}{Harsh
  Agrawal}, \bibinfo{person}{Dhruv Batra}, {and} \bibinfo{person}{Alexander
  Schwing}.} \bibinfo{year}{2019}\natexlab{}.
\newblock \showarticletitle{Sequential latent spaces for modeling the intention
  during diverse image captioning}. In \bibinfo{booktitle}{\emph{Proceedings of
  the IEEE International Conference on Computer Vision}}.
  \bibinfo{pages}{4261--4270}.
\newblock


\bibitem[\protect\citeauthoryear{Antol, Agrawal, Lu, Mitchell, Batra,
  Lawrence~Zitnick, and Parikh}{Antol et~al\mbox{.}}{2015}]%
        {antol2015vqa}
\bibfield{author}{\bibinfo{person}{Stanislaw Antol}, \bibinfo{person}{Aishwarya
  Agrawal}, \bibinfo{person}{Jiasen Lu}, \bibinfo{person}{Margaret Mitchell},
  \bibinfo{person}{Dhruv Batra}, \bibinfo{person}{C Lawrence~Zitnick}, {and}
  \bibinfo{person}{Devi Parikh}.} \bibinfo{year}{2015}\natexlab{}.
\newblock \showarticletitle{Vqa: Visual question answering}. In
  \bibinfo{booktitle}{\emph{Proceedings of the IEEE international conference on
  computer vision}}. \bibinfo{pages}{2425--2433}.
\newblock


\bibitem[\protect\citeauthoryear{Arjovsky, Chintala, and Bottou}{Arjovsky
  et~al\mbox{.}}{2017}]%
        {arjovsky2017wasserstein}
\bibfield{author}{\bibinfo{person}{Martin Arjovsky}, \bibinfo{person}{Soumith
  Chintala}, {and} \bibinfo{person}{L{\'e}on Bottou}.}
  \bibinfo{year}{2017}\natexlab{}.
\newblock \showarticletitle{Wasserstein Generative Adversarial Networks}. In
  \bibinfo{booktitle}{\emph{International Conference on Machine Learning}}.
  \bibinfo{pages}{214--223}.
\newblock


\bibitem[\protect\citeauthoryear{Asghar, Poupart, Hoey, Jiang, and Mou}{Asghar
  et~al\mbox{.}}{2018}]%
        {asghar2018affective}
\bibfield{author}{\bibinfo{person}{Nabiha Asghar}, \bibinfo{person}{Pascal
  Poupart}, \bibinfo{person}{Jesse Hoey}, \bibinfo{person}{Xin Jiang}, {and}
  \bibinfo{person}{Lili Mou}.} \bibinfo{year}{2018}\natexlab{}.
\newblock \showarticletitle{Affective neural response generation}. In
  \bibinfo{booktitle}{\emph{European Conference on Information Retrieval}}.
  Springer, \bibinfo{pages}{154--166}.
\newblock


\bibitem[\protect\citeauthoryear{Bahdanau, Cho, and Bengio}{Bahdanau
  et~al\mbox{.}}{2015}]%
        {bahdanau2015neural}
\bibfield{author}{\bibinfo{person}{Dzmitry Bahdanau},
  \bibinfo{person}{Kyunghyun Cho}, {and} \bibinfo{person}{Yoshua Bengio}.}
  \bibinfo{year}{2015}\natexlab{}.
\newblock \showarticletitle{Neural machine translation by jointly learning to
  align and translate}. In \bibinfo{booktitle}{\emph{3rd International
  Conference on Learning Representations, ICLR 2015}}.
\newblock


\bibitem[\protect\citeauthoryear{Baheti, Ritter, Li, and Dolan}{Baheti
  et~al\mbox{.}}{2018}]%
        {baheti2018generating}
\bibfield{author}{\bibinfo{person}{Ashutosh Baheti}, \bibinfo{person}{Alan
  Ritter}, \bibinfo{person}{Jiwei Li}, {and} \bibinfo{person}{Bill Dolan}.}
  \bibinfo{year}{2018}\natexlab{}.
\newblock \showarticletitle{Generating More Interesting Responses in Neural
  Conversation Models with Distributional Constraints}. In
  \bibinfo{booktitle}{\emph{Proceedings of the 2018 Conference on Empirical
  Methods in Natural Language Processing}}. \bibinfo{pages}{3970--3980}.
\newblock


\bibitem[\protect\citeauthoryear{Bai, Xia, and Xia}{Bai et~al\mbox{.}}{2020}]%
        {bai2020fusing}
\bibfield{author}{\bibinfo{person}{Peng Bai}, \bibinfo{person}{Yang Xia}, {and}
  \bibinfo{person}{Yongsheng Xia}.} \bibinfo{year}{2020}\natexlab{}.
\newblock \showarticletitle{Fusing Knowledge and Aspect Sentiment for
  Explainable Recommendation}.
\newblock \bibinfo{journal}{\emph{IEEE Access}}  \bibinfo{volume}{8}
  (\bibinfo{year}{2020}), \bibinfo{pages}{137150--137160}.
\newblock


\bibitem[\protect\citeauthoryear{Banerjee and Lavie}{Banerjee and
  Lavie}{2005}]%
        {banerjee2005meteor}
\bibfield{author}{\bibinfo{person}{Satanjeev Banerjee} {and}
  \bibinfo{person}{Alon Lavie}.} \bibinfo{year}{2005}\natexlab{}.
\newblock \showarticletitle{METEOR: An automatic metric for MT evaluation with
  improved correlation with human judgments}. In
  \bibinfo{booktitle}{\emph{Proceedings of the acl workshop on intrinsic and
  extrinsic evaluation measures for machine translation and/or summarization}}.
  \bibinfo{pages}{65--72}.
\newblock


\bibitem[\protect\citeauthoryear{Bordes, Usunier, Garcia-Duran, Weston, and
  Yakhnenko}{Bordes et~al\mbox{.}}{2013}]%
        {bordes2013translating}
\bibfield{author}{\bibinfo{person}{Antoine Bordes}, \bibinfo{person}{Nicolas
  Usunier}, \bibinfo{person}{Alberto Garcia-Duran}, \bibinfo{person}{Jason
  Weston}, {and} \bibinfo{person}{Oksana Yakhnenko}.}
  \bibinfo{year}{2013}\natexlab{}.
\newblock \showarticletitle{Translating embeddings for modeling
  multi-relational data}. In \bibinfo{booktitle}{\emph{Advances in neural
  information processing systems}}. \bibinfo{pages}{2787--2795}.
\newblock


\bibitem[\protect\citeauthoryear{Bowman, Vilnis, Vinyals, Dai, Jozefowicz, and
  Bengio}{Bowman et~al\mbox{.}}{2016}]%
        {bowman2016generating}
\bibfield{author}{\bibinfo{person}{Samuel Bowman}, \bibinfo{person}{Luke
  Vilnis}, \bibinfo{person}{Oriol Vinyals}, \bibinfo{person}{Andrew Dai},
  \bibinfo{person}{Rafal Jozefowicz}, {and} \bibinfo{person}{Samy Bengio}.}
  \bibinfo{year}{2016}\natexlab{}.
\newblock \showarticletitle{Generating Sentences from a Continuous Space}. In
  \bibinfo{booktitle}{\emph{Proceedings of The 20th SIGNLL Conference on
  Computational Natural Language Learning}}. \bibinfo{pages}{10--21}.
\newblock


\bibitem[\protect\citeauthoryear{Brown, Mann, Ryder, Subbiah, Kaplan, Dhariwal,
  Neelakantan, Shyam, Sastry, Askell, et~al\mbox{.}}{Brown
  et~al\mbox{.}}{2020}]%
        {brown2020language}
\bibfield{author}{\bibinfo{person}{Tom~B Brown}, \bibinfo{person}{Benjamin
  Mann}, \bibinfo{person}{Nick Ryder}, \bibinfo{person}{Melanie Subbiah},
  \bibinfo{person}{Jared Kaplan}, \bibinfo{person}{Prafulla Dhariwal},
  \bibinfo{person}{Arvind Neelakantan}, \bibinfo{person}{Pranav Shyam},
  \bibinfo{person}{Girish Sastry}, \bibinfo{person}{Amanda Askell},
  {et~al\mbox{.}}} \bibinfo{year}{2020}\natexlab{}.
\newblock \showarticletitle{Language models are few-shot learners}.
\newblock \bibinfo{journal}{\emph{arXiv preprint arXiv:2005.14165}}
  (\bibinfo{year}{2020}).
\newblock


\bibitem[\protect\citeauthoryear{Carbone and Sarti}{Carbone and Sarti}{2020}]%
        {carbone2020etc}
\bibfield{author}{\bibinfo{person}{Ginevra Carbone} {and}
  \bibinfo{person}{Gabriele Sarti}.} \bibinfo{year}{2020}\natexlab{}.
\newblock \showarticletitle{ETC-NLG: End-to-end Topic-Conditioned Natural
  Language Generation}.
\newblock \bibinfo{journal}{\emph{arXiv preprint arXiv:2008.10875}}
  (\bibinfo{year}{2020}).
\newblock


\bibitem[\protect\citeauthoryear{Chan, Ong, Pung, Zhang, and Fu}{Chan
  et~al\mbox{.}}{2020}]%
        {chan2020cocon}
\bibfield{author}{\bibinfo{person}{Alvin Chan}, \bibinfo{person}{Yew-Soon Ong},
  \bibinfo{person}{Bill Pung}, \bibinfo{person}{Aston Zhang}, {and}
  \bibinfo{person}{Jie Fu}.} \bibinfo{year}{2020}\natexlab{}.
\newblock \showarticletitle{CoCon: A Self-Supervised Approach for Controlled
  Text Generation}.
\newblock \bibinfo{journal}{\emph{arXiv preprint arXiv:2006.03535}}
  (\bibinfo{year}{2020}).
\newblock


\bibitem[\protect\citeauthoryear{Chan, Li, Yang, Chen, Hu, Zhao, and Yan}{Chan
  et~al\mbox{.}}{2019}]%
        {chan2019modeling}
\bibfield{author}{\bibinfo{person}{Zhangming Chan}, \bibinfo{person}{Juntao
  Li}, \bibinfo{person}{Xiaopeng Yang}, \bibinfo{person}{Xiuying Chen},
  \bibinfo{person}{Wenpeng Hu}, \bibinfo{person}{Dongyan Zhao}, {and}
  \bibinfo{person}{Rui Yan}.} \bibinfo{year}{2019}\natexlab{}.
\newblock \showarticletitle{Modeling personalization in continuous space for
  response generation via augmented wasserstein autoencoders}. In
  \bibinfo{booktitle}{\emph{Proceedings of the 2019 Conference on Empirical
  Methods in Natural Language Processing and the 9th International Joint
  Conference on Natural Language Processing (EMNLP-IJCNLP)}}.
  \bibinfo{pages}{1931--1940}.
\newblock


\bibitem[\protect\citeauthoryear{Chen, Ji, Ji, Sun, Zhang, Ge, Wu, Huang, and
  Wang}{Chen et~al\mbox{.}}{2019b}]%
        {chen2019variational}
\bibfield{author}{\bibinfo{person}{Fuhai Chen}, \bibinfo{person}{Rongrong Ji},
  \bibinfo{person}{Jiayi Ji}, \bibinfo{person}{Xiaoshuai Sun},
  \bibinfo{person}{Baochang Zhang}, \bibinfo{person}{Xuri Ge},
  \bibinfo{person}{Yongjian Wu}, \bibinfo{person}{Feiyue Huang}, {and}
  \bibinfo{person}{Yan Wang}.} \bibinfo{year}{2019}\natexlab{b}.
\newblock \showarticletitle{Variational Structured Semantic Inference for
  Diverse Image Captioning}. In \bibinfo{booktitle}{\emph{Advances in Neural
  Information Processing Systems}}. \bibinfo{pages}{1931--1941}.
\newblock


\bibitem[\protect\citeauthoryear{Chen, Liu, Yin, and Tang}{Chen
  et~al\mbox{.}}{2017}]%
        {chen2017survey}
\bibfield{author}{\bibinfo{person}{Hongshen Chen}, \bibinfo{person}{Xiaorui
  Liu}, \bibinfo{person}{Dawei Yin}, {and} \bibinfo{person}{Jiliang Tang}.}
  \bibinfo{year}{2017}\natexlab{}.
\newblock \showarticletitle{A survey on dialogue systems: Recent advances and
  new frontiers}.
\newblock \bibinfo{journal}{\emph{Acm Sigkdd Explorations Newsletter}}
  \bibinfo{volume}{19}, \bibinfo{number}{2} (\bibinfo{year}{2017}),
  \bibinfo{pages}{25--35}.
\newblock


\bibitem[\protect\citeauthoryear{Chen, Yi, Sun, Li, Yang, and Guo}{Chen
  et~al\mbox{.}}{2019e}]%
        {chen2019sentiment}
\bibfield{author}{\bibinfo{person}{Huimin Chen}, \bibinfo{person}{Xiaoyuan Yi},
  \bibinfo{person}{Maosong Sun}, \bibinfo{person}{Wenhao Li},
  \bibinfo{person}{Cheng Yang}, {and} \bibinfo{person}{Zhipeng Guo}.}
  \bibinfo{year}{2019}\natexlab{e}.
\newblock \showarticletitle{Sentiment-controllable Chinese poetry generation}.
  In \bibinfo{booktitle}{\emph{Proceedings of the 28th International Joint
  Conference on Artificial Intelligence}}. AAAI Press,
  \bibinfo{pages}{4925--4931}.
\newblock


\bibitem[\protect\citeauthoryear{Chen, Lin, Zhang, Yang, Zhou, and Tang}{Chen
  et~al\mbox{.}}{2019c}]%
        {chen2019towards}
\bibfield{author}{\bibinfo{person}{Qibin Chen}, \bibinfo{person}{Junyang Lin},
  \bibinfo{person}{Yichang Zhang}, \bibinfo{person}{Hongxia Yang},
  \bibinfo{person}{Jingren Zhou}, {and} \bibinfo{person}{Jie Tang}.}
  \bibinfo{year}{2019}\natexlab{c}.
\newblock \showarticletitle{Towards Knowledge-Based Personalized Product
  Description Generation in E-commerce}. In
  \bibinfo{booktitle}{\emph{Proceedings of the 25th ACM SIGKDD International
  Conference on Knowledge Discovery \& Data Mining}}. ACM,
  \bibinfo{pages}{3040--3050}.
\newblock


\bibitem[\protect\citeauthoryear{Chen, Wang, Feng, Jiang, Qin, and Lin}{Chen
  et~al\mbox{.}}{2019d}]%
        {chen2019enhancing}
\bibfield{author}{\bibinfo{person}{Shuang Chen}, \bibinfo{person}{Jinpeng
  Wang}, \bibinfo{person}{Xiaocheng Feng}, \bibinfo{person}{Feng Jiang},
  \bibinfo{person}{Bing Qin}, {and} \bibinfo{person}{Chin-Yew Lin}.}
  \bibinfo{year}{2019}\natexlab{d}.
\newblock \showarticletitle{Enhancing Neural Data-To-Text Generation Models
  with External Background Knowledge}. In \bibinfo{booktitle}{\emph{Proceedings
  of the 2019 Conference on Empirical Methods in Natural Language Processing
  and the 9th International Joint Conference on Natural Language Processing
  (EMNLP-IJCNLP)}}. \bibinfo{pages}{3013--3023}.
\newblock


\bibitem[\protect\citeauthoryear{Chen, Hsu, Kuo, Huang, and Ku}{Chen
  et~al\mbox{.}}{2019a}]%
        {chen2019emotionlines}
\bibfield{author}{\bibinfo{person}{Sheng~Yeh Chen}, \bibinfo{person}{Chao~Chun
  Hsu}, \bibinfo{person}{Chuan~Chun Kuo}, \bibinfo{person}{Kenneth Huang},
  {and} \bibinfo{person}{Lun~Wei Ku}.} \bibinfo{year}{2019}\natexlab{a}.
\newblock \showarticletitle{Emotionlines: An emotion corpus of multi-party
  conversations}. In \bibinfo{booktitle}{\emph{11th International Conference on
  Language Resources and Evaluation, LREC 2018}}. European Language Resources
  Association (ELRA), \bibinfo{pages}{1597--1601}.
\newblock


\bibitem[\protect\citeauthoryear{Chen, Gan, Cheng, Liu, and Liu}{Chen
  et~al\mbox{.}}{2020}]%
        {chen2020distilling}
\bibfield{author}{\bibinfo{person}{Yen-Chun Chen}, \bibinfo{person}{Zhe Gan},
  \bibinfo{person}{Yu Cheng}, \bibinfo{person}{Jingzhou Liu}, {and}
  \bibinfo{person}{Jingjing Liu}.} \bibinfo{year}{2020}\natexlab{}.
\newblock \showarticletitle{Distilling Knowledge Learned in BERT for Text
  Generation}. In \bibinfo{booktitle}{\emph{Proceedings of the 58th Annual
  Meeting of the Association for Computational Linguistics}}.
  \bibinfo{pages}{7893--7905}.
\newblock


\bibitem[\protect\citeauthoryear{Cho, van Merri{\"e}nboer, Gulcehre, Bahdanau,
  Bougares, Schwenk, and Bengio}{Cho et~al\mbox{.}}{2014}]%
        {cho2014learning}
\bibfield{author}{\bibinfo{person}{Kyunghyun Cho}, \bibinfo{person}{Bart van
  Merri{\"e}nboer}, \bibinfo{person}{Caglar Gulcehre}, \bibinfo{person}{Dzmitry
  Bahdanau}, \bibinfo{person}{Fethi Bougares}, \bibinfo{person}{Holger
  Schwenk}, {and} \bibinfo{person}{Yoshua Bengio}.}
  \bibinfo{year}{2014}\natexlab{}.
\newblock \showarticletitle{Learning Phrase Representations using RNN
  Encoder--Decoder for Statistical Machine Translation}. In
  \bibinfo{booktitle}{\emph{Proceedings of the 2014 Conference on Empirical
  Methods in Natural Language Processing (EMNLP)}}.
  \bibinfo{pages}{1724--1734}.
\newblock


\bibitem[\protect\citeauthoryear{Choudhary, Srivastava, Ungar, and
  Sedoc}{Choudhary et~al\mbox{.}}{2017}]%
        {choudhary2017domain}
\bibfield{author}{\bibinfo{person}{Sajal Choudhary}, \bibinfo{person}{Prerna
  Srivastava}, \bibinfo{person}{Lyle Ungar}, {and} \bibinfo{person}{Jo{\~a}o
  Sedoc}.} \bibinfo{year}{2017}\natexlab{}.
\newblock \showarticletitle{Domain aware neural dialog system}.
\newblock \bibinfo{journal}{\emph{arXiv preprint arXiv:1708.00897}}
  (\bibinfo{year}{2017}).
\newblock


\bibitem[\protect\citeauthoryear{Clark, Ji, and Smith}{Clark
  et~al\mbox{.}}{2018}]%
        {clark2018neural}
\bibfield{author}{\bibinfo{person}{Elizabeth Clark}, \bibinfo{person}{Yangfeng
  Ji}, {and} \bibinfo{person}{Noah~A Smith}.} \bibinfo{year}{2018}\natexlab{}.
\newblock \showarticletitle{Neural text generation in stories using entity
  representations as context}. In \bibinfo{booktitle}{\emph{Proceedings of the
  2018 Conference of the North American Chapter of the Association for
  Computational Linguistics: Human Language Technologies, Volume 1 (Long
  Papers)}}. \bibinfo{pages}{2250--2260}.
\newblock


\bibitem[\protect\citeauthoryear{Clark, Luong, Le, and Manning}{Clark
  et~al\mbox{.}}{2020}]%
        {clark2020electra}
\bibfield{author}{\bibinfo{person}{Kevin Clark}, \bibinfo{person}{Minh-Thang
  Luong}, \bibinfo{person}{Quoc~V Le}, {and} \bibinfo{person}{Christopher~D
  Manning}.} \bibinfo{year}{2020}\natexlab{}.
\newblock \showarticletitle{Electra: Pre-training text encoders as
  discriminators rather than generators}.
\newblock \bibinfo{journal}{\emph{arXiv preprint arXiv:2003.10555}}
  (\bibinfo{year}{2020}).
\newblock


\bibitem[\protect\citeauthoryear{Dai, Fidler, Urtasun, and Lin}{Dai
  et~al\mbox{.}}{2017}]%
        {dai2017towards}
\bibfield{author}{\bibinfo{person}{Bo Dai}, \bibinfo{person}{Sanja Fidler},
  \bibinfo{person}{Raquel Urtasun}, {and} \bibinfo{person}{Dahua Lin}.}
  \bibinfo{year}{2017}\natexlab{}.
\newblock \showarticletitle{Towards diverse and natural image descriptions via
  a conditional gan}. In \bibinfo{booktitle}{\emph{Proceedings of the IEEE
  International Conference on Computer Vision}}. \bibinfo{pages}{2970--2979}.
\newblock


\bibitem[\protect\citeauthoryear{Danescu-Niculescu-Mizil and
  Lee}{Danescu-Niculescu-Mizil and Lee}{2011}]%
        {danescu2011chameleons}
\bibfield{author}{\bibinfo{person}{Cristian Danescu-Niculescu-Mizil} {and}
  \bibinfo{person}{Lillian Lee}.} \bibinfo{year}{2011}\natexlab{}.
\newblock \showarticletitle{Chameleons in imagined conversations: A new
  approach to understanding coordination of linguistic style in dialogs}. In
  \bibinfo{booktitle}{\emph{Proceedings of the 2nd workshop on cognitive
  modeling and computational linguistics}}. Association for Computational
  Linguistics, \bibinfo{pages}{76--87}.
\newblock


\bibitem[\protect\citeauthoryear{Das, Kottur, Gupta, Singh, Yadav, Moura,
  Parikh, and Batra}{Das et~al\mbox{.}}{2017}]%
        {das2017visual}
\bibfield{author}{\bibinfo{person}{Abhishek Das}, \bibinfo{person}{Satwik
  Kottur}, \bibinfo{person}{Khushi Gupta}, \bibinfo{person}{Avi Singh},
  \bibinfo{person}{Deshraj Yadav}, \bibinfo{person}{Jos{\'e}~MF Moura},
  \bibinfo{person}{Devi Parikh}, {and} \bibinfo{person}{Dhruv Batra}.}
  \bibinfo{year}{2017}\natexlab{}.
\newblock \showarticletitle{Visual dialog}. In
  \bibinfo{booktitle}{\emph{Proceedings of the IEEE Conference on Computer
  Vision and Pattern Recognition}}. \bibinfo{pages}{326--335}.
\newblock


\bibitem[\protect\citeauthoryear{Dathathri, Madotto, Lan, Hung, Frank, Molino,
  Yosinski, and Liu}{Dathathri et~al\mbox{.}}{2019}]%
        {dathathri2019plug}
\bibfield{author}{\bibinfo{person}{Sumanth Dathathri}, \bibinfo{person}{Andrea
  Madotto}, \bibinfo{person}{Janice Lan}, \bibinfo{person}{Jane Hung},
  \bibinfo{person}{Eric Frank}, \bibinfo{person}{Piero Molino},
  \bibinfo{person}{Jason Yosinski}, {and} \bibinfo{person}{Rosanne Liu}.}
  \bibinfo{year}{2019}\natexlab{}.
\newblock \showarticletitle{Plug and Play Language Models: A Simple Approach to
  Controlled Text Generation}. In \bibinfo{booktitle}{\emph{International
  Conference on Learning Representations}}.
\newblock


\bibitem[\protect\citeauthoryear{De~Cao, Aziz, and Titov}{De~Cao
  et~al\mbox{.}}{2019}]%
        {de2019question}
\bibfield{author}{\bibinfo{person}{Nicola De~Cao}, \bibinfo{person}{Wilker
  Aziz}, {and} \bibinfo{person}{Ivan Titov}.} \bibinfo{year}{2019}\natexlab{}.
\newblock \showarticletitle{Question Answering by Reasoning Across Documents
  with Graph Convolutional Networks}. In \bibinfo{booktitle}{\emph{Proceedings
  of NAACL-HLT}}. \bibinfo{pages}{2306--2317}.
\newblock


\bibitem[\protect\citeauthoryear{Dehghani, Azarbonyad, Kamps, and
  de~Rijke}{Dehghani et~al\mbox{.}}{2019}]%
        {dehghani2019learning}
\bibfield{author}{\bibinfo{person}{Mostafa Dehghani}, \bibinfo{person}{Hosein
  Azarbonyad}, \bibinfo{person}{Jaap Kamps}, {and} \bibinfo{person}{Maarten de
  Rijke}.} \bibinfo{year}{2019}\natexlab{}.
\newblock \showarticletitle{Learning to Transform, Combine, and Reason in
  Open-Domain Question Answering.}. In \bibinfo{booktitle}{\emph{WSDM}}.
  \bibinfo{pages}{681--689}.
\newblock


\bibitem[\protect\citeauthoryear{Dinan, Roller, Shuster, Fan, Auli, and
  Weston}{Dinan et~al\mbox{.}}{2018}]%
        {dinan2018wizard}
\bibfield{author}{\bibinfo{person}{Emily Dinan}, \bibinfo{person}{Stephen
  Roller}, \bibinfo{person}{Kurt Shuster}, \bibinfo{person}{Angela Fan},
  \bibinfo{person}{Michael Auli}, {and} \bibinfo{person}{Jason Weston}.}
  \bibinfo{year}{2018}\natexlab{}.
\newblock \showarticletitle{Wizard of Wikipedia: Knowledge-Powered
  Conversational Agents}. In \bibinfo{booktitle}{\emph{International Conference
  on Learning Representations}}.
\newblock


\bibitem[\protect\citeauthoryear{Doddington}{Doddington}{2002}]%
        {doddington2002automatic}
\bibfield{author}{\bibinfo{person}{George Doddington}.}
  \bibinfo{year}{2002}\natexlab{}.
\newblock \showarticletitle{Automatic evaluation of machine translation quality
  using n-gram co-occurrence statistics}. In
  \bibinfo{booktitle}{\emph{Proceedings of the second international conference
  on Human Language Technology Research}}. \bibinfo{pages}{138--145}.
\newblock


\bibitem[\protect\citeauthoryear{Dong, Yang, Wang, Wei, Liu, Wang, Gao, Zhou,
  and Hon}{Dong et~al\mbox{.}}{2019}]%
        {dong2019unified}
\bibfield{author}{\bibinfo{person}{Li Dong}, \bibinfo{person}{Nan Yang},
  \bibinfo{person}{Wenhui Wang}, \bibinfo{person}{Furu Wei},
  \bibinfo{person}{Xiaodong Liu}, \bibinfo{person}{Yu Wang},
  \bibinfo{person}{Jianfeng Gao}, \bibinfo{person}{Ming Zhou}, {and}
  \bibinfo{person}{Hsiao-Wuen Hon}.} \bibinfo{year}{2019}\natexlab{}.
\newblock \showarticletitle{Unified language model pre-training for natural
  language understanding and generation}. In \bibinfo{booktitle}{\emph{Advances
  in Neural Information Processing Systems}}. \bibinfo{pages}{13063--13075}.
\newblock


\bibitem[\protect\citeauthoryear{Dziri, Kamalloo, Mathewson, and Zaiane}{Dziri
  et~al\mbox{.}}{2019}]%
        {dziri2019augmenting}
\bibfield{author}{\bibinfo{person}{Nouha Dziri}, \bibinfo{person}{Ehsan
  Kamalloo}, \bibinfo{person}{Kory Mathewson}, {and} \bibinfo{person}{Osmar~R
  Zaiane}.} \bibinfo{year}{2019}\natexlab{}.
\newblock \showarticletitle{Augmenting Neural Response Generation with
  Context-Aware Topical Attention}. In \bibinfo{booktitle}{\emph{Proceedings of
  the First Workshop on NLP for Conversational AI}}. \bibinfo{pages}{18--31}.
\newblock


\bibitem[\protect\citeauthoryear{Elman}{Elman}{1990}]%
        {elman1990finding}
\bibfield{author}{\bibinfo{person}{Jeffrey~L Elman}.}
  \bibinfo{year}{1990}\natexlab{}.
\newblock \showarticletitle{Finding structure in time}.
\newblock \bibinfo{journal}{\emph{Cognitive science}} \bibinfo{volume}{14},
  \bibinfo{number}{2} (\bibinfo{year}{1990}), \bibinfo{pages}{179--211}.
\newblock


\bibitem[\protect\citeauthoryear{Fan, Lewis, and Dauphin}{Fan
  et~al\mbox{.}}{2018}]%
        {fan2018hierarchical}
\bibfield{author}{\bibinfo{person}{Angela Fan}, \bibinfo{person}{Mike Lewis},
  {and} \bibinfo{person}{Yann Dauphin}.} \bibinfo{year}{2018}\natexlab{}.
\newblock \showarticletitle{Hierarchical Neural Story Generation}. In
  \bibinfo{booktitle}{\emph{Proceedings of the 56th Annual Meeting of the
  Association for Computational Linguistics (Volume 1: Long Papers)}}.
  \bibinfo{pages}{889--898}.
\newblock


\bibitem[\protect\citeauthoryear{Feng, Liu, Liu, Qin, Sun, and Liu}{Feng
  et~al\mbox{.}}{2018}]%
        {feng2018topic}
\bibfield{author}{\bibinfo{person}{Xiaocheng Feng}, \bibinfo{person}{Ming Liu},
  \bibinfo{person}{Jiahao Liu}, \bibinfo{person}{Bing Qin},
  \bibinfo{person}{Yibo Sun}, {and} \bibinfo{person}{Ting Liu}.}
  \bibinfo{year}{2018}\natexlab{}.
\newblock \showarticletitle{Topic-to-Essay Generation with Neural Networks.}.
  In \bibinfo{booktitle}{\emph{IJCAI}}. \bibinfo{pages}{4078--4084}.
\newblock


\bibitem[\protect\citeauthoryear{Feng, Ma, Liu, and Luo}{Feng
  et~al\mbox{.}}{2019}]%
        {feng2019unsupervised}
\bibfield{author}{\bibinfo{person}{Yang Feng}, \bibinfo{person}{Lin Ma},
  \bibinfo{person}{Wei Liu}, {and} \bibinfo{person}{Jiebo Luo}.}
  \bibinfo{year}{2019}\natexlab{}.
\newblock \showarticletitle{Unsupervised image captioning}. In
  \bibinfo{booktitle}{\emph{Proceedings of the IEEE Conference on Computer
  Vision and Pattern Recognition}}. \bibinfo{pages}{4125--4134}.
\newblock


\bibitem[\protect\citeauthoryear{Fu, Tan, Peng, Zhao, and Yan}{Fu
  et~al\mbox{.}}{2018}]%
        {fu2018style}
\bibfield{author}{\bibinfo{person}{Zhenxin Fu}, \bibinfo{person}{Xiaoye Tan},
  \bibinfo{person}{Nanyun Peng}, \bibinfo{person}{Dongyan Zhao}, {and}
  \bibinfo{person}{Rui Yan}.} \bibinfo{year}{2018}\natexlab{}.
\newblock \showarticletitle{Style transfer in text: Exploration and
  evaluation}. In \bibinfo{booktitle}{\emph{Thirty-Second AAAI Conference on
  Artificial Intelligence}}.
\newblock


\bibitem[\protect\citeauthoryear{Galley, Brockett, Sordoni, Ji, Auli, Quirk,
  Mitchell, Gao, and Dolan}{Galley et~al\mbox{.}}{2015}]%
        {galley2015deltableu}
\bibfield{author}{\bibinfo{person}{Michel Galley}, \bibinfo{person}{Chris
  Brockett}, \bibinfo{person}{Alessandro Sordoni}, \bibinfo{person}{Yangfeng
  Ji}, \bibinfo{person}{Michael Auli}, \bibinfo{person}{Chris Quirk},
  \bibinfo{person}{Margaret Mitchell}, \bibinfo{person}{Jianfeng Gao}, {and}
  \bibinfo{person}{Bill Dolan}.} \bibinfo{year}{2015}\natexlab{}.
\newblock \showarticletitle{deltaBLEU: A Discriminative Metric for Generation
  Tasks with Intrinsically Diverse Targets}. In
  \bibinfo{booktitle}{\emph{Proceedings of the 53rd Annual Meeting of the
  Association for Computational Linguistics and the 7th International Joint
  Conference on Natural Language Processing (Volume 2: Short Papers)}}.
  \bibinfo{pages}{445--450}.
\newblock


\bibitem[\protect\citeauthoryear{Gao and Ren}{Gao and Ren}{2019}]%
        {gao2019topic}
\bibfield{author}{\bibinfo{person}{Ce Gao} {and} \bibinfo{person}{Jiangtao
  Ren}.} \bibinfo{year}{2019}\natexlab{}.
\newblock \showarticletitle{A topic-driven language model for learning to
  generate diverse sentences}.
\newblock \bibinfo{journal}{\emph{Neurocomputing}}  \bibinfo{volume}{333}
  (\bibinfo{year}{2019}), \bibinfo{pages}{374--380}.
\newblock


\bibitem[\protect\citeauthoryear{Gao, Galley, Li, et~al\mbox{.}}{Gao
  et~al\mbox{.}}{2019}]%
        {gao2019neural}
\bibfield{author}{\bibinfo{person}{Jianfeng Gao}, \bibinfo{person}{Michel
  Galley}, \bibinfo{person}{Lihong Li}, {et~al\mbox{.}}}
  \bibinfo{year}{2019}\natexlab{}.
\newblock \showarticletitle{Neural approaches to conversational AI}.
\newblock \bibinfo{journal}{\emph{Foundations and Trends{\textregistered} in
  Information Retrieval}} \bibinfo{volume}{13}, \bibinfo{number}{2-3}
  (\bibinfo{year}{2019}), \bibinfo{pages}{127--298}.
\newblock


\bibitem[\protect\citeauthoryear{Genest and Lapalme}{Genest and
  Lapalme}{2011}]%
        {genest2011framework}
\bibfield{author}{\bibinfo{person}{Pierre-Etienne Genest} {and}
  \bibinfo{person}{Guy Lapalme}.} \bibinfo{year}{2011}\natexlab{}.
\newblock \showarticletitle{Framework for abstractive summarization using
  text-to-text generation}. In \bibinfo{booktitle}{\emph{Proceedings of the
  Workshop on Monolingual Text-To-Text Generation}}. Association for
  Computational Linguistics, \bibinfo{pages}{64--73}.
\newblock


\bibitem[\protect\citeauthoryear{Ghazvininejad, Brockett, Chang, Dolan, Gao,
  Yih, and Galley}{Ghazvininejad et~al\mbox{.}}{2018}]%
        {ghazvininejad2018knowledge}
\bibfield{author}{\bibinfo{person}{Marjan Ghazvininejad},
  \bibinfo{person}{Chris Brockett}, \bibinfo{person}{Ming-Wei Chang},
  \bibinfo{person}{Bill Dolan}, \bibinfo{person}{Jianfeng Gao},
  \bibinfo{person}{Wen-tau Yih}, {and} \bibinfo{person}{Michel Galley}.}
  \bibinfo{year}{2018}\natexlab{}.
\newblock \showarticletitle{A knowledge-grounded neural conversation model}. In
  \bibinfo{booktitle}{\emph{Thirty-Second AAAI Conference on Artificial
  Intelligence}}.
\newblock


\bibitem[\protect\citeauthoryear{Ghazvininejad, Shi, Priyadarshi, and
  Knight}{Ghazvininejad et~al\mbox{.}}{2017}]%
        {ghazvininejad2017hafez}
\bibfield{author}{\bibinfo{person}{Marjan Ghazvininejad}, \bibinfo{person}{Xing
  Shi}, \bibinfo{person}{Jay Priyadarshi}, {and} \bibinfo{person}{Kevin
  Knight}.} \bibinfo{year}{2017}\natexlab{}.
\newblock \showarticletitle{Hafez: an interactive poetry generation system}. In
  \bibinfo{booktitle}{\emph{Proceedings of ACL 2017, System Demonstrations}}.
  \bibinfo{pages}{43--48}.
\newblock


\bibitem[\protect\citeauthoryear{Goodfellow, Pouget-Abadie, Mirza, Xu,
  Warde-Farley, Ozair, Courville, and Bengio}{Goodfellow et~al\mbox{.}}{2014}]%
        {goodfellow2014generative}
\bibfield{author}{\bibinfo{person}{Ian Goodfellow}, \bibinfo{person}{Jean
  Pouget-Abadie}, \bibinfo{person}{Mehdi Mirza}, \bibinfo{person}{Bing Xu},
  \bibinfo{person}{David Warde-Farley}, \bibinfo{person}{Sherjil Ozair},
  \bibinfo{person}{Aaron Courville}, {and} \bibinfo{person}{Yoshua Bengio}.}
  \bibinfo{year}{2014}\natexlab{}.
\newblock \showarticletitle{Generative adversarial nets}. In
  \bibinfo{booktitle}{\emph{Advances in neural information processing
  systems}}. \bibinfo{pages}{2672--2680}.
\newblock


\bibitem[\protect\citeauthoryear{Guan, Wang, and Huang}{Guan
  et~al\mbox{.}}{2019}]%
        {guan2019story}
\bibfield{author}{\bibinfo{person}{Jian Guan}, \bibinfo{person}{Yansen Wang},
  {and} \bibinfo{person}{Minlie Huang}.} \bibinfo{year}{2019}\natexlab{}.
\newblock \showarticletitle{Story ending generation with incremental encoding
  and commonsense knowledge}. In \bibinfo{booktitle}{\emph{Proceedings of the
  AAAI Conference on Artificial Intelligence}}, Vol.~\bibinfo{volume}{33}.
  \bibinfo{pages}{6473--6480}.
\newblock


\bibitem[\protect\citeauthoryear{Gunel, Zhu, Zeng, and Huang}{Gunel
  et~al\mbox{.}}{2019}]%
        {gunel2019mind}
\bibfield{author}{\bibinfo{person}{Beliz Gunel}, \bibinfo{person}{Chenguang
  Zhu}, \bibinfo{person}{Michael Zeng}, {and} \bibinfo{person}{Xuedong Huang}.}
  \bibinfo{year}{2019}\natexlab{}.
\newblock \showarticletitle{Mind The Facts: Knowledge-Boosted Coherent
  Abstractive Text Summarization}. In \bibinfo{booktitle}{\emph{NeurIPS 2019}}.
\newblock


\bibitem[\protect\citeauthoryear{Guo, Lu, Cai, Zhang, Yu, and Wang}{Guo
  et~al\mbox{.}}{2018}]%
        {guo2018long}
\bibfield{author}{\bibinfo{person}{Jiaxian Guo}, \bibinfo{person}{Sidi Lu},
  \bibinfo{person}{Han Cai}, \bibinfo{person}{Weinan Zhang},
  \bibinfo{person}{Yong Yu}, {and} \bibinfo{person}{Jun Wang}.}
  \bibinfo{year}{2018}\natexlab{}.
\newblock \showarticletitle{Long text generation via adversarial training with
  leaked information}. In \bibinfo{booktitle}{\emph{Thirty-Second AAAI
  Conference on Artificial Intelligence}}.
\newblock


\bibitem[\protect\citeauthoryear{Hermann, Kocisky, Grefenstette, Espeholt, Kay,
  Suleyman, and Blunsom}{Hermann et~al\mbox{.}}{2015}]%
        {hermann2015teaching}
\bibfield{author}{\bibinfo{person}{Karl~Moritz Hermann}, \bibinfo{person}{Tomas
  Kocisky}, \bibinfo{person}{Edward Grefenstette}, \bibinfo{person}{Lasse
  Espeholt}, \bibinfo{person}{Will Kay}, \bibinfo{person}{Mustafa Suleyman},
  {and} \bibinfo{person}{Phil Blunsom}.} \bibinfo{year}{2015}\natexlab{}.
\newblock \showarticletitle{Teaching machines to read and comprehend}. In
  \bibinfo{booktitle}{\emph{Advances in neural information processing
  systems}}. \bibinfo{pages}{1693--1701}.
\newblock


\bibitem[\protect\citeauthoryear{Herzig, Shmueli-Scheuer, Sandbank, and
  Konopnicki}{Herzig et~al\mbox{.}}{2017}]%
        {herzig2017neural}
\bibfield{author}{\bibinfo{person}{Jonathan Herzig}, \bibinfo{person}{Michal
  Shmueli-Scheuer}, \bibinfo{person}{Tommy Sandbank}, {and}
  \bibinfo{person}{David Konopnicki}.} \bibinfo{year}{2017}\natexlab{}.
\newblock \showarticletitle{Neural response generation for customer service
  based on personality traits}. In \bibinfo{booktitle}{\emph{Proceedings of the
  10th International Conference on Natural Language Generation}}.
  \bibinfo{pages}{252--256}.
\newblock


\bibitem[\protect\citeauthoryear{Holtzman, Buys, Du, Forbes, and Choi}{Holtzman
  et~al\mbox{.}}{2019}]%
        {holtzman2019curious}
\bibfield{author}{\bibinfo{person}{Ari Holtzman}, \bibinfo{person}{Jan Buys},
  \bibinfo{person}{Li Du}, \bibinfo{person}{Maxwell Forbes}, {and}
  \bibinfo{person}{Yejin Choi}.} \bibinfo{year}{2019}\natexlab{}.
\newblock \showarticletitle{The Curious Case of Neural Text Degeneration}. In
  \bibinfo{booktitle}{\emph{International Conference on Learning
  Representations}}.
\newblock


\bibitem[\protect\citeauthoryear{Holtzman, Buys, Forbes, Bosselut, Golub, and
  Choi}{Holtzman et~al\mbox{.}}{2018}]%
        {holtzman2018learning}
\bibfield{author}{\bibinfo{person}{Ari Holtzman}, \bibinfo{person}{Jan Buys},
  \bibinfo{person}{Maxwell Forbes}, \bibinfo{person}{Antoine Bosselut},
  \bibinfo{person}{David Golub}, {and} \bibinfo{person}{Yejin Choi}.}
  \bibinfo{year}{2018}\natexlab{}.
\newblock \showarticletitle{Learning to Write with Cooperative Discriminators}.
  In \bibinfo{booktitle}{\emph{Proceedings of the 56th Annual Meeting of the
  Association for Computational Linguistics (Volume 1: Long Papers)}}.
  \bibinfo{pages}{1638--1649}.
\newblock


\bibitem[\protect\citeauthoryear{Hu, Yang, Liang, Salakhutdinov, and Xing}{Hu
  et~al\mbox{.}}{2017}]%
        {hu2017toward}
\bibfield{author}{\bibinfo{person}{Zhiting Hu}, \bibinfo{person}{Zichao Yang},
  \bibinfo{person}{Xiaodan Liang}, \bibinfo{person}{Ruslan Salakhutdinov},
  {and} \bibinfo{person}{Eric~P Xing}.} \bibinfo{year}{2017}\natexlab{}.
\newblock \showarticletitle{Toward controlled generation of text}. In
  \bibinfo{booktitle}{\emph{Proceedings of the 34th International Conference on
  Machine Learning-Volume 70}}. JMLR. org, \bibinfo{pages}{1587--1596}.
\newblock


\bibitem[\protect\citeauthoryear{Jaech and Ostendorf}{Jaech and
  Ostendorf}{2018}]%
        {jaech2018low}
\bibfield{author}{\bibinfo{person}{Aaron Jaech} {and} \bibinfo{person}{Mari
  Ostendorf}.} \bibinfo{year}{2018}\natexlab{}.
\newblock \showarticletitle{Low-rank RNN adaptation for context-aware language
  modeling}.
\newblock \bibinfo{journal}{\emph{Transactions of the Association for
  Computational Linguistics}}  \bibinfo{volume}{6} (\bibinfo{year}{2018}),
  \bibinfo{pages}{497--510}.
\newblock


\bibitem[\protect\citeauthoryear{Jelinek, Mercer, Bahl, and Baker}{Jelinek
  et~al\mbox{.}}{1977}]%
        {jelinek1977perplexity}
\bibfield{author}{\bibinfo{person}{Fred Jelinek}, \bibinfo{person}{Robert~L
  Mercer}, \bibinfo{person}{Lalit~R Bahl}, {and} \bibinfo{person}{James~K
  Baker}.} \bibinfo{year}{1977}\natexlab{}.
\newblock \showarticletitle{Perplexity—a measure of the difficulty of speech
  recognition tasks}.
\newblock \bibinfo{journal}{\emph{The Journal of the Acoustical Society of
  America}} \bibinfo{volume}{62}, \bibinfo{number}{S1} (\bibinfo{year}{1977}),
  \bibinfo{pages}{S63--S63}.
\newblock


\bibitem[\protect\citeauthoryear{Kang, Zhao, Zhang, and Zong}{Kang
  et~al\mbox{.}}{2020}]%
        {kang2020dynamic}
\bibfield{author}{\bibinfo{person}{Xiaomian Kang}, \bibinfo{person}{Yang Zhao},
  \bibinfo{person}{Jiajun Zhang}, {and} \bibinfo{person}{Chengqing Zong}.}
  \bibinfo{year}{2020}\natexlab{}.
\newblock \showarticletitle{Dynamic Context Selection for Document-level Neural
  Machine Translation via Reinforcement Learning}. In
  \bibinfo{booktitle}{\emph{Proceedings of the 2020 Conference on Empirical
  Methods in Natural Language Processing (EMNLP)}}.
  \bibinfo{pages}{2242--2254}.
\newblock


\bibitem[\protect\citeauthoryear{Kao, Chen, and Tsai}{Kao
  et~al\mbox{.}}{2019}]%
        {kao2019model}
\bibfield{author}{\bibinfo{person}{Chien-Hao Kao}, \bibinfo{person}{Chih-Chieh
  Chen}, {and} \bibinfo{person}{Yu-Tza Tsai}.} \bibinfo{year}{2019}\natexlab{}.
\newblock \showarticletitle{Model of Multi-turn Dialogue in Emotional Chatbot}.
  In \bibinfo{booktitle}{\emph{2019 International Conference on Technologies
  and Applications of Artificial Intelligence (TAAI)}}. IEEE,
  \bibinfo{pages}{1--5}.
\newblock


\bibitem[\protect\citeauthoryear{Kenton and Toutanova}{Kenton and
  Toutanova}{2019}]%
        {kenton2019bert}
\bibfield{author}{\bibinfo{person}{Jacob Devlin Ming-Wei~Chang Kenton} {and}
  \bibinfo{person}{Lee~Kristina Toutanova}.} \bibinfo{year}{2019}\natexlab{}.
\newblock \showarticletitle{BERT: Pre-training of Deep Bidirectional
  Transformers for Language Understanding}. In
  \bibinfo{booktitle}{\emph{Proceedings of NAACL-HLT}}.
  \bibinfo{pages}{4171--4186}.
\newblock


\bibitem[\protect\citeauthoryear{Keskar, McCann, Varshney, Xiong, and
  Socher}{Keskar et~al\mbox{.}}{2019}]%
        {keskar2019ctrl}
\bibfield{author}{\bibinfo{person}{Nitish~Shirish Keskar},
  \bibinfo{person}{Bryan McCann}, \bibinfo{person}{Lav~R Varshney},
  \bibinfo{person}{Caiming Xiong}, {and} \bibinfo{person}{Richard Socher}.}
  \bibinfo{year}{2019}\natexlab{}.
\newblock \showarticletitle{Ctrl: A conditional transformer language model for
  controllable generation}.
\newblock \bibinfo{journal}{\emph{arXiv preprint arXiv:1909.05858}}
  (\bibinfo{year}{2019}).
\newblock


\bibitem[\protect\citeauthoryear{Kim, Ahn, and Kim}{Kim et~al\mbox{.}}{2020}]%
        {Kim2020Sequential}
\bibfield{author}{\bibinfo{person}{Byeongchang Kim}, \bibinfo{person}{Jaewoo
  Ahn}, {and} \bibinfo{person}{Gunhee Kim}.} \bibinfo{year}{2020}\natexlab{}.
\newblock \showarticletitle{Sequential Latent Knowledge Selection for
  Knowledge-Grounded Dialogue}. In \bibinfo{booktitle}{\emph{International
  Conference on Learning Representations}}.
\newblock
\urldef\tempurl%
\url{https://openreview.net/forum?id=Hke0K1HKwr}
\showURL{%
\tempurl}


\bibitem[\protect\citeauthoryear{Kingma and Welling}{Kingma and
  Welling}{2013}]%
        {kingma2013auto}
\bibfield{author}{\bibinfo{person}{Diederik~P Kingma} {and}
  \bibinfo{person}{Max Welling}.} \bibinfo{year}{2013}\natexlab{}.
\newblock \showarticletitle{Auto-encoding variational bayes}.
\newblock \bibinfo{journal}{\emph{arXiv preprint arXiv:1312.6114}}
  (\bibinfo{year}{2013}).
\newblock


\bibitem[\protect\citeauthoryear{Kipf and Welling}{Kipf and Welling}{2016}]%
        {kipf2016semi}
\bibfield{author}{\bibinfo{person}{Thomas~N Kipf} {and} \bibinfo{person}{Max
  Welling}.} \bibinfo{year}{2016}\natexlab{}.
\newblock \showarticletitle{Semi-supervised classification with graph
  convolutional networks}.
\newblock \bibinfo{journal}{\emph{arXiv preprint arXiv:1609.02907}}
  (\bibinfo{year}{2016}).
\newblock


\bibitem[\protect\citeauthoryear{Koncel-Kedziorski, Bekal, Luan, Lapata, and
  Hajishirzi}{Koncel-Kedziorski et~al\mbox{.}}{2019}]%
        {koncel2019text}
\bibfield{author}{\bibinfo{person}{Rik Koncel-Kedziorski},
  \bibinfo{person}{Dhanush Bekal}, \bibinfo{person}{Yi Luan},
  \bibinfo{person}{Mirella Lapata}, {and} \bibinfo{person}{Hannaneh
  Hajishirzi}.} \bibinfo{year}{2019}\natexlab{}.
\newblock \showarticletitle{Text Generation from Knowledge Graphs with Graph
  Transformers}. In \bibinfo{booktitle}{\emph{Proceedings of NAACL-HLT}}.
  \bibinfo{pages}{2284--2293}.
\newblock


\bibitem[\protect\citeauthoryear{Kong, Li, Neubig, Hovy, and Yang}{Kong
  et~al\mbox{.}}{2019}]%
        {kong2019adversarial}
\bibfield{author}{\bibinfo{person}{Xiang Kong}, \bibinfo{person}{Bohan Li},
  \bibinfo{person}{Graham Neubig}, \bibinfo{person}{Eduard Hovy}, {and}
  \bibinfo{person}{Yiming Yang}.} \bibinfo{year}{2019}\natexlab{}.
\newblock \showarticletitle{An Adversarial Approach to High-Quality,
  Sentiment-Controlled Neural Dialogue Generation}.
\newblock \bibinfo{journal}{\emph{arXiv preprint arXiv:1901.07129}}
  (\bibinfo{year}{2019}).
\newblock


\bibitem[\protect\citeauthoryear{Krishna and Srinivasan}{Krishna and
  Srinivasan}{2018}]%
        {krishna2018generating}
\bibfield{author}{\bibinfo{person}{Kundan Krishna} {and}
  \bibinfo{person}{Balaji~Vasan Srinivasan}.} \bibinfo{year}{2018}\natexlab{}.
\newblock \showarticletitle{Generating topic-oriented summaries using neural
  attention}. In \bibinfo{booktitle}{\emph{Proceedings of the 2018 Conference
  of the North American Chapter of the Association for Computational
  Linguistics: Human Language Technologies, Volume 1 (Long Papers)}}.
  \bibinfo{pages}{1697--1705}.
\newblock


\bibitem[\protect\citeauthoryear{Lewis, Liu, Goyal, Ghazvininejad, Mohamed,
  Levy, Stoyanov, and Zettlemoyer}{Lewis et~al\mbox{.}}{2019}]%
        {lewis2019bart}
\bibfield{author}{\bibinfo{person}{Mike Lewis}, \bibinfo{person}{Yinhan Liu},
  \bibinfo{person}{Naman Goyal}, \bibinfo{person}{Marjan Ghazvininejad},
  \bibinfo{person}{Abdelrahman Mohamed}, \bibinfo{person}{Omer Levy},
  \bibinfo{person}{Ves Stoyanov}, {and} \bibinfo{person}{Luke Zettlemoyer}.}
  \bibinfo{year}{2019}\natexlab{}.
\newblock \showarticletitle{Bart: Denoising sequence-to-sequence pre-training
  for natural language generation, translation, and comprehension}.
\newblock \bibinfo{journal}{\emph{arXiv preprint arXiv:1910.13461}}
  (\bibinfo{year}{2019}).
\newblock


\bibitem[\protect\citeauthoryear{Li, Gao, Li, Li, Peng, Zhang, and Gao}{Li
  et~al\mbox{.}}{2020}]%
        {li2020optimus}
\bibfield{author}{\bibinfo{person}{Chunyuan Li}, \bibinfo{person}{Xiang Gao},
  \bibinfo{person}{Yuan Li}, \bibinfo{person}{Xiujun Li},
  \bibinfo{person}{Baolin Peng}, \bibinfo{person}{Yizhe Zhang}, {and}
  \bibinfo{person}{Jianfeng Gao}.} \bibinfo{year}{2020}\natexlab{}.
\newblock \showarticletitle{Optimus: Organizing sentences via pre-trained
  modeling of a latent space}.
\newblock \bibinfo{journal}{\emph{arXiv preprint arXiv:2004.04092}}
  (\bibinfo{year}{2020}).
\newblock


\bibitem[\protect\citeauthoryear{Li, Wang, Shen, and Hengel}{Li
  et~al\mbox{.}}{2019g}]%
        {li2019visual}
\bibfield{author}{\bibinfo{person}{Hui Li}, \bibinfo{person}{Peng Wang},
  \bibinfo{person}{Chunhua Shen}, {and} \bibinfo{person}{Anton van~den
  Hengel}.} \bibinfo{year}{2019}\natexlab{g}.
\newblock \showarticletitle{Visual Question Answering as Reading
  Comprehension}. In \bibinfo{booktitle}{\emph{Proceedings of the IEEE
  Conference on Computer Vision and Pattern Recognition}}.
  \bibinfo{pages}{6319--6328}.
\newblock


\bibitem[\protect\citeauthoryear{Li, Galley, Brockett, Spithourakis, Gao, and
  Dolan}{Li et~al\mbox{.}}{2016a}]%
        {li2016persona}
\bibfield{author}{\bibinfo{person}{Jiwei Li}, \bibinfo{person}{Michel Galley},
  \bibinfo{person}{Chris Brockett}, \bibinfo{person}{Georgios Spithourakis},
  \bibinfo{person}{Jianfeng Gao}, {and} \bibinfo{person}{Bill Dolan}.}
  \bibinfo{year}{2016}\natexlab{a}.
\newblock \showarticletitle{A Persona-Based Neural Conversation Model}. In
  \bibinfo{booktitle}{\emph{Proceedings of the 54th Annual Meeting of the
  Association for Computational Linguistics (Volume 1: Long Papers)}}.
  \bibinfo{pages}{994--1003}.
\newblock


\bibitem[\protect\citeauthoryear{Li, Li, and Zong}{Li et~al\mbox{.}}{2019b}]%
        {li2019towards}
\bibfield{author}{\bibinfo{person}{Junjie Li}, \bibinfo{person}{Haoran Li},
  {and} \bibinfo{person}{Chengqing Zong}.} \bibinfo{year}{2019}\natexlab{b}.
\newblock \showarticletitle{Towards Personalized Review Summarization via
  User-Aware Sequence Network}. In \bibinfo{booktitle}{\emph{Proceedings of the
  AAAI Conference on Artificial Intelligence}}, Vol.~\bibinfo{volume}{33}.
  \bibinfo{pages}{6690--6697}.
\newblock


\bibitem[\protect\citeauthoryear{Li, Monroe, and Jurafsky}{Li
  et~al\mbox{.}}{2016b}]%
        {li2016simple}
\bibfield{author}{\bibinfo{person}{Jiwei Li}, \bibinfo{person}{Will Monroe},
  {and} \bibinfo{person}{Dan Jurafsky}.} \bibinfo{year}{2016}\natexlab{b}.
\newblock \showarticletitle{A simple, fast diverse decoding algorithm for
  neural generation}.
\newblock \bibinfo{journal}{\emph{arXiv preprint arXiv:1611.08562}}
  (\bibinfo{year}{2016}).
\newblock


\bibitem[\protect\citeauthoryear{Li, Sun, Wei, Li, and Tao}{Li
  et~al\mbox{.}}{2019e}]%
        {li2019reinforcement}
\bibfield{author}{\bibinfo{person}{Jia Li}, \bibinfo{person}{Xiao Sun},
  \bibinfo{person}{Xing Wei}, \bibinfo{person}{Changliang Li}, {and}
  \bibinfo{person}{Jianhua Tao}.} \bibinfo{year}{2019}\natexlab{e}.
\newblock \showarticletitle{Reinforcement Learning Based Emotional Editing
  Constraint Conversation Generation}.
\newblock \bibinfo{journal}{\emph{arXiv preprint arXiv:1904.08061}}
  (\bibinfo{year}{2019}).
\newblock


\bibitem[\protect\citeauthoryear{Li, Jin, et~al\mbox{.}}{Li
  et~al\mbox{.}}{2019a}]%
        {li2019topic}
\bibfield{author}{\bibinfo{person}{Miao Li}, \bibinfo{person}{Beihong Jin},
  {et~al\mbox{.}}} \bibinfo{year}{2019}\natexlab{a}.
\newblock \showarticletitle{A Topic Augmented Text Generation Model: Joint
  Learning of Semantics and Structural Features}. In
  \bibinfo{booktitle}{\emph{Proceedings of the 2019 Conference on Empirical
  Methods in Natural Language Processing and the 9th International Joint
  Conference on Natural Language Processing (EMNLP-IJCNLP)}}.
  \bibinfo{pages}{5093--5102}.
\newblock


\bibitem[\protect\citeauthoryear{Li, Roller, Kulikov, Welleck, Boureau, Cho,
  and Weston}{Li et~al\mbox{.}}{2019d}]%
        {li2019don}
\bibfield{author}{\bibinfo{person}{Margaret Li}, \bibinfo{person}{Stephen
  Roller}, \bibinfo{person}{Ilia Kulikov}, \bibinfo{person}{Sean Welleck},
  \bibinfo{person}{Y-Lan Boureau}, \bibinfo{person}{Kyunghyun Cho}, {and}
  \bibinfo{person}{Jason Weston}.} \bibinfo{year}{2019}\natexlab{d}.
\newblock \showarticletitle{Don't Say That! Making Inconsistent Dialogue
  Unlikely with Unlikelihood Training}.
\newblock \bibinfo{journal}{\emph{arXiv preprint arXiv:1911.03860}}
  (\bibinfo{year}{2019}).
\newblock


\bibitem[\protect\citeauthoryear{Li, Wang, Bing, and Lam}{Li
  et~al\mbox{.}}{2019f}]%
        {li2019persona}
\bibfield{author}{\bibinfo{person}{Piji Li}, \bibinfo{person}{Zihao Wang},
  \bibinfo{person}{Lidong Bing}, {and} \bibinfo{person}{Wai Lam}.}
  \bibinfo{year}{2019}\natexlab{f}.
\newblock \showarticletitle{Persona-Aware Tips Generation?}. In
  \bibinfo{booktitle}{\emph{The World Wide Web Conference}}.
  \bibinfo{pages}{1006--1016}.
\newblock


\bibitem[\protect\citeauthoryear{Li, Niu, Meng, Feng, Li, and Zhou}{Li
  et~al\mbox{.}}{2019c}]%
        {li2019incremental}
\bibfield{author}{\bibinfo{person}{Zekang Li}, \bibinfo{person}{Cheng Niu},
  \bibinfo{person}{Fandong Meng}, \bibinfo{person}{Yang Feng},
  \bibinfo{person}{Qian Li}, {and} \bibinfo{person}{Jie Zhou}.}
  \bibinfo{year}{2019}\natexlab{c}.
\newblock \showarticletitle{Incremental Transformer with Deliberation Decoder
  for Document Grounded Conversations}. In
  \bibinfo{booktitle}{\emph{Proceedings of the 57th Annual Meeting of the
  Association for Computational Linguistics}}. \bibinfo{pages}{12--21}.
\newblock


\bibitem[\protect\citeauthoryear{Lin and Och}{Lin and Och}{2004a}]%
        {lin2004looking}
\bibfield{author}{\bibinfo{person}{Chin-Yew Lin} {and} \bibinfo{person}{FJ
  Och}.} \bibinfo{year}{2004}\natexlab{a}.
\newblock \showarticletitle{Looking for a few good metrics: ROUGE and its
  evaluation}. In \bibinfo{booktitle}{\emph{Ntcir Workshop}}.
\newblock


\bibitem[\protect\citeauthoryear{Lin and Och}{Lin and Och}{2004b}]%
        {lin2004orange}
\bibfield{author}{\bibinfo{person}{Chin-Yew Lin} {and}
  \bibinfo{person}{Franz~Josef Och}.} \bibinfo{year}{2004}\natexlab{b}.
\newblock \showarticletitle{Orange: a method for evaluating automatic
  evaluation metrics for machine translation}. In
  \bibinfo{booktitle}{\emph{COLING 2004: Proceedings of the 20th International
  Conference on Computational Linguistics}}. \bibinfo{pages}{501--507}.
\newblock


\bibitem[\protect\citeauthoryear{Liu, Ott, Goyal, Du, Joshi, Chen, Levy, Lewis,
  Zettlemoyer, and Stoyanov}{Liu et~al\mbox{.}}{2019}]%
        {liu2019roberta}
\bibfield{author}{\bibinfo{person}{Yinhan Liu}, \bibinfo{person}{Myle Ott},
  \bibinfo{person}{Naman Goyal}, \bibinfo{person}{Jingfei Du},
  \bibinfo{person}{Mandar Joshi}, \bibinfo{person}{Danqi Chen},
  \bibinfo{person}{Omer Levy}, \bibinfo{person}{Mike Lewis},
  \bibinfo{person}{Luke Zettlemoyer}, {and} \bibinfo{person}{Veselin
  Stoyanov}.} \bibinfo{year}{2019}\natexlab{}.
\newblock \showarticletitle{Roberta: A robustly optimized bert pretraining
  approach}.
\newblock \bibinfo{journal}{\emph{arXiv preprint arXiv:1907.11692}}
  (\bibinfo{year}{2019}).
\newblock


\bibitem[\protect\citeauthoryear{Lowe, Pow, Serban, and Pineau}{Lowe
  et~al\mbox{.}}{2015}]%
        {lowe2015ubuntu}
\bibfield{author}{\bibinfo{person}{Ryan Lowe}, \bibinfo{person}{Nissan Pow},
  \bibinfo{person}{Iulian Serban}, {and} \bibinfo{person}{Joelle Pineau}.}
  \bibinfo{year}{2015}\natexlab{}.
\newblock \showarticletitle{The ubuntu dialogue corpus: A large dataset for
  research in unstructured multi-turn dialogue systems}.
\newblock \bibinfo{journal}{\emph{arXiv preprint arXiv:1506.08909}}
  (\bibinfo{year}{2015}).
\newblock


\bibitem[\protect\citeauthoryear{Lu, Xiong, Parikh, and Socher}{Lu
  et~al\mbox{.}}{2017}]%
        {lu2017knowing}
\bibfield{author}{\bibinfo{person}{Jiasen Lu}, \bibinfo{person}{Caiming Xiong},
  \bibinfo{person}{Devi Parikh}, {and} \bibinfo{person}{Richard Socher}.}
  \bibinfo{year}{2017}\natexlab{}.
\newblock \showarticletitle{Knowing when to look: Adaptive attention via a
  visual sentinel for image captioning}. In
  \bibinfo{booktitle}{\emph{Proceedings of the IEEE conference on computer
  vision and pattern recognition}}. \bibinfo{pages}{375--383}.
\newblock


\bibitem[\protect\citeauthoryear{Luan, Brockett, Dolan, Gao, and Galley}{Luan
  et~al\mbox{.}}{2017}]%
        {luan2017multi}
\bibfield{author}{\bibinfo{person}{Yi Luan}, \bibinfo{person}{Chris Brockett},
  \bibinfo{person}{Bill Dolan}, \bibinfo{person}{Jianfeng Gao}, {and}
  \bibinfo{person}{Michel Galley}.} \bibinfo{year}{2017}\natexlab{}.
\newblock \showarticletitle{Multi-Task Learning for Speaker-Role Adaptation in
  Neural Conversation Models}. In \bibinfo{booktitle}{\emph{Proceedings of the
  Eighth International Joint Conference on Natural Language Processing (Volume
  1: Long Papers)}}. \bibinfo{pages}{605--614}.
\newblock


\bibitem[\protect\citeauthoryear{Luo, Li, Yang, Zhou, Tan, Chang, Sui, and
  Sun}{Luo et~al\mbox{.}}{2019b}]%
        {luo2019towards}
\bibfield{author}{\bibinfo{person}{Fuli Luo}, \bibinfo{person}{Peng Li},
  \bibinfo{person}{Pengcheng Yang}, \bibinfo{person}{Jie Zhou},
  \bibinfo{person}{Yutong Tan}, \bibinfo{person}{Baobao Chang},
  \bibinfo{person}{Zhifang Sui}, {and} \bibinfo{person}{Xu Sun}.}
  \bibinfo{year}{2019}\natexlab{b}.
\newblock \showarticletitle{Towards fine-grained text sentiment transfer}. In
  \bibinfo{booktitle}{\emph{Proceedings of the 57th Annual Meeting of the
  Association for Computational Linguistics}}. \bibinfo{pages}{2013--2022}.
\newblock


\bibitem[\protect\citeauthoryear{Luo, Huang, Zeng, Nie, and Sun}{Luo
  et~al\mbox{.}}{2019a}]%
        {luo2019learning}
\bibfield{author}{\bibinfo{person}{Liangchen Luo}, \bibinfo{person}{Wenhao
  Huang}, \bibinfo{person}{Qi Zeng}, \bibinfo{person}{Zaiqing Nie}, {and}
  \bibinfo{person}{Xu Sun}.} \bibinfo{year}{2019}\natexlab{a}.
\newblock \showarticletitle{Learning personalized end-to-end goal-oriented
  dialog}. In \bibinfo{booktitle}{\emph{Proceedings of the AAAI Conference on
  Artificial Intelligence}}, Vol.~\bibinfo{volume}{33}.
  \bibinfo{pages}{6794--6801}.
\newblock


\bibitem[\protect\citeauthoryear{Madotto, Lin, Bang, and Fung}{Madotto
  et~al\mbox{.}}{2020}]%
        {madotto2020adapter}
\bibfield{author}{\bibinfo{person}{Andrea Madotto}, \bibinfo{person}{Zhaojiang
  Lin}, \bibinfo{person}{Yejin Bang}, {and} \bibinfo{person}{Pascale Fung}.}
  \bibinfo{year}{2020}\natexlab{}.
\newblock \showarticletitle{The Adapter-Bot: All-In-One Controllable
  Conversational Model}.
\newblock \bibinfo{journal}{\emph{arXiv preprint arXiv:2008.12579}}
  (\bibinfo{year}{2020}).
\newblock


\bibitem[\protect\citeauthoryear{Majumder, Hong, Peng, Lu, Ghosal, Gelbukh,
  Mihalcea, and Poria}{Majumder et~al\mbox{.}}{2020}]%
        {majumder2020mime}
\bibfield{author}{\bibinfo{person}{Navonil Majumder}, \bibinfo{person}{Pengfei
  Hong}, \bibinfo{person}{Shanshan Peng}, \bibinfo{person}{Jiankun Lu},
  \bibinfo{person}{Deepanway Ghosal}, \bibinfo{person}{Alexander Gelbukh},
  \bibinfo{person}{Rada Mihalcea}, {and} \bibinfo{person}{Soujanya Poria}.}
  \bibinfo{year}{2020}\natexlab{}.
\newblock \showarticletitle{MIME: MIMicking Emotions for Empathetic Response
  Generation}. In \bibinfo{booktitle}{\emph{Proceedings of the 2020 Conference
  on Empirical Methods in Natural Language Processing (EMNLP)}}.
  \bibinfo{pages}{8968--8979}.
\newblock


\bibitem[\protect\citeauthoryear{Malinowski, Rohrbach, and Fritz}{Malinowski
  et~al\mbox{.}}{2015}]%
        {malinowski2015ask}
\bibfield{author}{\bibinfo{person}{Mateusz Malinowski}, \bibinfo{person}{Marcus
  Rohrbach}, {and} \bibinfo{person}{Mario Fritz}.}
  \bibinfo{year}{2015}\natexlab{}.
\newblock \showarticletitle{Ask your neurons: A neural-based approach to
  answering questions about images}. In \bibinfo{booktitle}{\emph{Proceedings
  of the IEEE international conference on computer vision}}.
  \bibinfo{pages}{1--9}.
\newblock


\bibitem[\protect\citeauthoryear{Mao, Majumder, McAuley, and Cottrell}{Mao
  et~al\mbox{.}}{2019}]%
        {mao2019improving}
\bibfield{author}{\bibinfo{person}{Huanru~Henry Mao},
  \bibinfo{person}{Bodhisattwa~Prasad Majumder}, \bibinfo{person}{Julian
  McAuley}, {and} \bibinfo{person}{Garrison Cottrell}.}
  \bibinfo{year}{2019}\natexlab{}.
\newblock \showarticletitle{Improving Neural Story Generation by Targeted
  Common Sense Grounding}. In \bibinfo{booktitle}{\emph{Proceedings of the 2019
  Conference on Empirical Methods in Natural Language Processing and the 9th
  International Joint Conference on Natural Language Processing
  (EMNLP-IJCNLP)}}. \bibinfo{pages}{5990--5995}.
\newblock


\bibitem[\protect\citeauthoryear{Mao, Xu, Yang, Wang, and Yuille}{Mao
  et~al\mbox{.}}{2014}]%
        {mao2014explain}
\bibfield{author}{\bibinfo{person}{Junhua Mao}, \bibinfo{person}{Wei Xu},
  \bibinfo{person}{Yi Yang}, \bibinfo{person}{Jiang Wang}, {and}
  \bibinfo{person}{Alan~L Yuille}.} \bibinfo{year}{2014}\natexlab{}.
\newblock \showarticletitle{Explain images with multimodal recurrent neural
  networks}.
\newblock \bibinfo{journal}{\emph{arXiv preprint arXiv:1410.1090}}
  (\bibinfo{year}{2014}).
\newblock


\bibitem[\protect\citeauthoryear{Mazar{\'e}, Humeau, Raison, and
  Bordes}{Mazar{\'e} et~al\mbox{.}}{2018}]%
        {mazare2018training}
\bibfield{author}{\bibinfo{person}{Pierre-Emmanuel Mazar{\'e}},
  \bibinfo{person}{Samuel Humeau}, \bibinfo{person}{Martin Raison}, {and}
  \bibinfo{person}{Antoine Bordes}.} \bibinfo{year}{2018}\natexlab{}.
\newblock \showarticletitle{Training Millions of Personalized Dialogue Agents}.
  In \bibinfo{booktitle}{\emph{EMNLP}}.
\newblock


\bibitem[\protect\citeauthoryear{Mazumder, Ma, and Liu}{Mazumder
  et~al\mbox{.}}{2018}]%
        {mazumder2018towards}
\bibfield{author}{\bibinfo{person}{Sahisnu Mazumder}, \bibinfo{person}{Nianzu
  Ma}, {and} \bibinfo{person}{Bing Liu}.} \bibinfo{year}{2018}\natexlab{}.
\newblock \showarticletitle{Towards a continuous knowledge learning engine for
  chatbots}.
\newblock \bibinfo{journal}{\emph{arXiv preprint arXiv:1802.06024}}
  (\bibinfo{year}{2018}).
\newblock


\bibitem[\protect\citeauthoryear{McCann, Bradbury, Xiong, and Socher}{McCann
  et~al\mbox{.}}{2017}]%
        {mccann2017learned}
\bibfield{author}{\bibinfo{person}{Bryan McCann}, \bibinfo{person}{James
  Bradbury}, \bibinfo{person}{Caiming Xiong}, {and} \bibinfo{person}{Richard
  Socher}.} \bibinfo{year}{2017}\natexlab{}.
\newblock \showarticletitle{Learned in translation: Contextualized word
  vectors}. In \bibinfo{booktitle}{\emph{Advances in Neural Information
  Processing Systems}}. \bibinfo{pages}{6294--6305}.
\newblock


\bibitem[\protect\citeauthoryear{McKeown}{McKeown}{1992}]%
        {mckeown1992text}
\bibfield{author}{\bibinfo{person}{Kathleen McKeown}.}
  \bibinfo{year}{1992}\natexlab{}.
\newblock \bibinfo{booktitle}{\emph{Text generation}}.
\newblock \bibinfo{publisher}{Cambridge University Press}.
\newblock


\bibitem[\protect\citeauthoryear{Mikolov, Sutskever, Chen, Corrado, and
  Dean}{Mikolov et~al\mbox{.}}{2013}]%
        {mikolov2013distributed}
\bibfield{author}{\bibinfo{person}{Tomas Mikolov}, \bibinfo{person}{Ilya
  Sutskever}, \bibinfo{person}{Kai Chen}, \bibinfo{person}{Greg~S Corrado},
  {and} \bibinfo{person}{Jeff Dean}.} \bibinfo{year}{2013}\natexlab{}.
\newblock \showarticletitle{Distributed representations of words and phrases
  and their compositionality}. In \bibinfo{booktitle}{\emph{Advances in neural
  information processing systems}}. \bibinfo{pages}{3111--3119}.
\newblock


\bibitem[\protect\citeauthoryear{Moon, Shah, Kumar, and Subba}{Moon
  et~al\mbox{.}}{2019}]%
        {moon2019opendialkg}
\bibfield{author}{\bibinfo{person}{Seungwhan Moon}, \bibinfo{person}{Pararth
  Shah}, \bibinfo{person}{Anuj Kumar}, {and} \bibinfo{person}{Rajen Subba}.}
  \bibinfo{year}{2019}\natexlab{}.
\newblock \showarticletitle{Opendialkg: Explainable conversational reasoning
  with attention-based walks over knowledge graphs}. In
  \bibinfo{booktitle}{\emph{Proceedings of the 57th Annual Meeting of the
  Association for Computational Linguistics}}. \bibinfo{pages}{845--854}.
\newblock


\bibitem[\protect\citeauthoryear{Moussallem, Ar{\v{c}}an, Ngomo, and
  Buitelaar}{Moussallem et~al\mbox{.}}{2019}]%
        {moussallem2019augmenting}
\bibfield{author}{\bibinfo{person}{Diego Moussallem}, \bibinfo{person}{Mihael
  Ar{\v{c}}an}, \bibinfo{person}{Axel-Cyrille~Ngonga Ngomo}, {and}
  \bibinfo{person}{Paul Buitelaar}.} \bibinfo{year}{2019}\natexlab{}.
\newblock \showarticletitle{Augmenting neural machine translation with
  knowledge graphs}.
\newblock \bibinfo{journal}{\emph{arXiv preprint arXiv:1902.08816}}
  (\bibinfo{year}{2019}).
\newblock


\bibitem[\protect\citeauthoryear{Nallapati, Zhou, dos Santos, glar
  Gul{\c{c}}ehre, and Xiang}{Nallapati et~al\mbox{.}}{2016}]%
        {nallapati2016abstractive}
\bibfield{author}{\bibinfo{person}{Ramesh Nallapati}, \bibinfo{person}{Bowen
  Zhou}, \bibinfo{person}{Cicero dos Santos}, \bibinfo{person}{{\c{C}}a glar
  Gul{\c{c}}ehre}, {and} \bibinfo{person}{Bing Xiang}.}
  \bibinfo{year}{2016}\natexlab{}.
\newblock \showarticletitle{Abstractive Text Summarization using
  Sequence-to-sequence RNNs and Beyond}.
\newblock \bibinfo{journal}{\emph{CoNLL 2016}} (\bibinfo{year}{2016}),
  \bibinfo{pages}{280}.
\newblock


\bibitem[\protect\citeauthoryear{Ni and McAuley}{Ni and McAuley}{2018}]%
        {ni2018personalized}
\bibfield{author}{\bibinfo{person}{Jianmo Ni} {and} \bibinfo{person}{Julian
  McAuley}.} \bibinfo{year}{2018}\natexlab{}.
\newblock \showarticletitle{Personalized Review Generation by Expanding Phrases
  and Attending on Aspect-Aware Representations}. In
  \bibinfo{booktitle}{\emph{Proceedings of the 56th Annual Meeting of the
  Association for Computational Linguistics (Volume 2: Short Papers)}}.
  \bibinfo{pages}{706--711}.
\newblock


\bibitem[\protect\citeauthoryear{Oremus}{Oremus}{2014}]%
        {oremus2014first}
\bibfield{author}{\bibinfo{person}{Will Oremus}.}
  \bibinfo{year}{2014}\natexlab{}.
\newblock \showarticletitle{The first news report on the LA earthquake was
  written by a robot}.
\newblock \bibinfo{journal}{\emph{Slate. com}}  \bibinfo{volume}{17}
  (\bibinfo{year}{2014}).
\newblock


\bibitem[\protect\citeauthoryear{Papineni, Roukos, Ward, and Zhu}{Papineni
  et~al\mbox{.}}{2002}]%
        {papineni2002bleu}
\bibfield{author}{\bibinfo{person}{Kishore Papineni}, \bibinfo{person}{Salim
  Roukos}, \bibinfo{person}{Todd Ward}, {and} \bibinfo{person}{Wei-Jing Zhu}.}
  \bibinfo{year}{2002}\natexlab{}.
\newblock \showarticletitle{BLEU: a method for automatic evaluation of machine
  translation}. In \bibinfo{booktitle}{\emph{Proceedings of the 40th annual
  meeting on association for computational linguistics}}. Association for
  Computational Linguistics, \bibinfo{pages}{311--318}.
\newblock


\bibitem[\protect\citeauthoryear{Peng, Fang, Xie, and Zhou}{Peng
  et~al\mbox{.}}{2019}]%
        {peng2019topic}
\bibfield{author}{\bibinfo{person}{Yehong Peng}, \bibinfo{person}{Yizhen Fang},
  \bibinfo{person}{Zhiwen Xie}, {and} \bibinfo{person}{Guangyou Zhou}.}
  \bibinfo{year}{2019}\natexlab{}.
\newblock \showarticletitle{Topic-enhanced emotional conversation generation
  with attention mechanism}.
\newblock \bibinfo{journal}{\emph{Knowledge-Based Systems}}
  \bibinfo{volume}{163} (\bibinfo{year}{2019}), \bibinfo{pages}{429--437}.
\newblock


\bibitem[\protect\citeauthoryear{Peters, Neumann, Iyyer, Gardner, Clark, Lee,
  and Zettlemoyer}{Peters et~al\mbox{.}}{2018}]%
        {peters2018deep}
\bibfield{author}{\bibinfo{person}{Matthew~E Peters}, \bibinfo{person}{Mark
  Neumann}, \bibinfo{person}{Mohit Iyyer}, \bibinfo{person}{Matt Gardner},
  \bibinfo{person}{Christopher Clark}, \bibinfo{person}{Kenton Lee}, {and}
  \bibinfo{person}{Luke Zettlemoyer}.} \bibinfo{year}{2018}\natexlab{}.
\newblock \showarticletitle{Deep contextualized word representations}. In
  \bibinfo{booktitle}{\emph{Proceedings of NAACL-HLT}}.
  \bibinfo{pages}{2227--2237}.
\newblock


\bibitem[\protect\citeauthoryear{Radev, Hovy, and McKeown}{Radev
  et~al\mbox{.}}{2002}]%
        {radev2002introduction}
\bibfield{author}{\bibinfo{person}{Dragomir~R Radev}, \bibinfo{person}{Eduard
  Hovy}, {and} \bibinfo{person}{Kathleen McKeown}.}
  \bibinfo{year}{2002}\natexlab{}.
\newblock \showarticletitle{Introduction to the special issue on
  summarization}.
\newblock \bibinfo{journal}{\emph{Computational linguistics}}
  \bibinfo{volume}{28}, \bibinfo{number}{4} (\bibinfo{year}{2002}),
  \bibinfo{pages}{399--408}.
\newblock


\bibitem[\protect\citeauthoryear{Radford, Narasimhan, Salimans, and
  Sutskever}{Radford et~al\mbox{.}}{2018}]%
        {radford2018improving}
\bibfield{author}{\bibinfo{person}{Alec Radford}, \bibinfo{person}{Karthik
  Narasimhan}, \bibinfo{person}{Tim Salimans}, {and} \bibinfo{person}{Ilya
  Sutskever}.} \bibinfo{year}{2018}\natexlab{}.
\newblock \showarticletitle{Improving language understanding by generative
  pre-training}.
\newblock \bibinfo{journal}{\emph{URL https://s3-us-west-2. amazonaws.
  com/openai-assets/researchcovers/languageunsupervised/language understanding
  paper. pdf}} (\bibinfo{year}{2018}).
\newblock


\bibitem[\protect\citeauthoryear{Radford, Wu, Child, Luan, Amodei, and
  Sutskever}{Radford et~al\mbox{.}}{2019}]%
        {radford2019language}
\bibfield{author}{\bibinfo{person}{Alec Radford}, \bibinfo{person}{Jeffrey Wu},
  \bibinfo{person}{Rewon Child}, \bibinfo{person}{David Luan},
  \bibinfo{person}{Dario Amodei}, {and} \bibinfo{person}{Ilya Sutskever}.}
  \bibinfo{year}{2019}\natexlab{}.
\newblock \showarticletitle{Language models are unsupervised multitask
  learners}.
\newblock \bibinfo{journal}{\emph{OpenAI Blog}} \bibinfo{volume}{1},
  \bibinfo{number}{8} (\bibinfo{year}{2019}), \bibinfo{pages}{9}.
\newblock


\bibitem[\protect\citeauthoryear{Raffel, Shazeer, Roberts, Lee, Narang, Matena,
  Zhou, Li, and Liu}{Raffel et~al\mbox{.}}{2019}]%
        {raffel2019exploring}
\bibfield{author}{\bibinfo{person}{Colin Raffel}, \bibinfo{person}{Noam
  Shazeer}, \bibinfo{person}{Adam Roberts}, \bibinfo{person}{Katherine Lee},
  \bibinfo{person}{Sharan Narang}, \bibinfo{person}{Michael Matena},
  \bibinfo{person}{Yanqi Zhou}, \bibinfo{person}{Wei Li}, {and}
  \bibinfo{person}{Peter~J Liu}.} \bibinfo{year}{2019}\natexlab{}.
\newblock \showarticletitle{Exploring the limits of transfer learning with a
  unified text-to-text transformer}.
\newblock \bibinfo{journal}{\emph{arXiv preprint arXiv:1910.10683}}
  (\bibinfo{year}{2019}).
\newblock


\bibitem[\protect\citeauthoryear{Rashkin, Smith, Li, and Boureau}{Rashkin
  et~al\mbox{.}}{2019}]%
        {rashkin2019towards}
\bibfield{author}{\bibinfo{person}{Hannah Rashkin},
  \bibinfo{person}{Eric~Michael Smith}, \bibinfo{person}{Margaret Li}, {and}
  \bibinfo{person}{Y-Lan Boureau}.} \bibinfo{year}{2019}\natexlab{}.
\newblock \showarticletitle{Towards Empathetic Open-domain Conversation Models:
  A New Benchmark and Dataset}. In \bibinfo{booktitle}{\emph{Proceedings of the
  57th Conference of the Association for Computational Linguistics}}.
  \bibinfo{pages}{5370--5381}.
\newblock


\bibitem[\protect\citeauthoryear{Ren, Chen, Monz, Ma, and de~Rijke}{Ren
  et~al\mbox{.}}{2019}]%
        {ren2019thinking}
\bibfield{author}{\bibinfo{person}{Pengjie Ren}, \bibinfo{person}{Zhumin Chen},
  \bibinfo{person}{Christof Monz}, \bibinfo{person}{Jun Ma}, {and}
  \bibinfo{person}{Maarten de Rijke}.} \bibinfo{year}{2019}\natexlab{}.
\newblock \showarticletitle{Thinking Globally, Acting Locally: Distantly
  Supervised Global-to-Local Knowledge Selection for Background Based
  Conversation}.
\newblock \bibinfo{journal}{\emph{arXiv preprint arXiv:1908.09528}}
  (\bibinfo{year}{2019}).
\newblock


\bibitem[\protect\citeauthoryear{Rush, Harvard, Chopra, and Weston}{Rush
  et~al\mbox{.}}{2017}]%
        {rush2017neural}
\bibfield{author}{\bibinfo{person}{Alexander~M Rush}, \bibinfo{person}{SEAS
  Harvard}, \bibinfo{person}{Sumit Chopra}, {and} \bibinfo{person}{Jason
  Weston}.} \bibinfo{year}{2017}\natexlab{}.
\newblock \showarticletitle{A Neural Attention Model for Sentence
  Summarization}. In \bibinfo{booktitle}{\emph{ACLWeb. Proceedings of the 2015
  Conference on Empirical Methods in Natural Language Processing}}.
\newblock


\bibitem[\protect\citeauthoryear{See, Roller, Kiela, and Weston}{See
  et~al\mbox{.}}{2019}]%
        {see2019makes}
\bibfield{author}{\bibinfo{person}{Abigail See}, \bibinfo{person}{Stephen
  Roller}, \bibinfo{person}{Douwe Kiela}, {and} \bibinfo{person}{Jason
  Weston}.} \bibinfo{year}{2019}\natexlab{}.
\newblock \showarticletitle{What makes a good conversation? How controllable
  attributes affect human judgments}. In \bibinfo{booktitle}{\emph{Proceedings
  of the 2019 Conference of the North American Chapter of the Association for
  Computational Linguistics: Human Language Technologies, Volume 1 (Long and
  Short Papers)}}. \bibinfo{pages}{1702--1723}.
\newblock


\bibitem[\protect\citeauthoryear{Serban, Sordoni, Bengio, Courville, and
  Pineau}{Serban et~al\mbox{.}}{2016}]%
        {serban2016building}
\bibfield{author}{\bibinfo{person}{Iulian~V Serban},
  \bibinfo{person}{Alessandro Sordoni}, \bibinfo{person}{Yoshua Bengio},
  \bibinfo{person}{Aaron Courville}, {and} \bibinfo{person}{Joelle Pineau}.}
  \bibinfo{year}{2016}\natexlab{}.
\newblock \showarticletitle{Building end-to-end dialogue systems using
  generative hierarchical neural network models}. In
  \bibinfo{booktitle}{\emph{Thirtieth AAAI Conference on Artificial
  Intelligence}}.
\newblock


\bibitem[\protect\citeauthoryear{Serban, Sordoni, Lowe, Charlin, Pineau,
  Courville, and Bengio}{Serban et~al\mbox{.}}{2017}]%
        {serban2017hierarchical}
\bibfield{author}{\bibinfo{person}{Iulian~Vlad Serban},
  \bibinfo{person}{Alessandro Sordoni}, \bibinfo{person}{Ryan Lowe},
  \bibinfo{person}{Laurent Charlin}, \bibinfo{person}{Joelle Pineau},
  \bibinfo{person}{Aaron Courville}, {and} \bibinfo{person}{Yoshua Bengio}.}
  \bibinfo{year}{2017}\natexlab{}.
\newblock \showarticletitle{A hierarchical latent variable encoder-decoder
  model for generating dialogues}. In \bibinfo{booktitle}{\emph{Thirty-First
  AAAI Conference on Artificial Intelligence}}.
\newblock


\bibitem[\protect\citeauthoryear{Shang, Lu, and Li}{Shang
  et~al\mbox{.}}{2015}]%
        {shang2015neural}
\bibfield{author}{\bibinfo{person}{Lifeng Shang}, \bibinfo{person}{Zhengdong
  Lu}, {and} \bibinfo{person}{Hang Li}.} \bibinfo{year}{2015}\natexlab{}.
\newblock \showarticletitle{Neural Responding Machine for Short-Text
  Conversation}. In \bibinfo{booktitle}{\emph{Proceedings of the 53rd Annual
  Meeting of the Association for Computational Linguistics and the 7th
  International Joint Conference on Natural Language Processing (Volume 1: Long
  Papers)}}. \bibinfo{pages}{1577--1586}.
\newblock


\bibitem[\protect\citeauthoryear{Shao, Huang, Wen, Xu, et~al\mbox{.}}{Shao
  et~al\mbox{.}}{2019}]%
        {shao2019long}
\bibfield{author}{\bibinfo{person}{Zhihong Shao}, \bibinfo{person}{Minlie
  Huang}, \bibinfo{person}{Jiangtao Wen}, \bibinfo{person}{Wenfei Xu},
  {et~al\mbox{.}}} \bibinfo{year}{2019}\natexlab{}.
\newblock \showarticletitle{Long and Diverse Text Generation with
  Planning-based Hierarchical Variational Model}. In
  \bibinfo{booktitle}{\emph{Proceedings of the 2019 Conference on Empirical
  Methods in Natural Language Processing and the 9th International Joint
  Conference on Natural Language Processing (EMNLP-IJCNLP)}}.
  \bibinfo{pages}{3248--3259}.
\newblock


\bibitem[\protect\citeauthoryear{Shen, Celikyilmaz, Zhang, Chen, Wang, Gao, and
  Carin}{Shen et~al\mbox{.}}{2019}]%
        {shen2019towards}
\bibfield{author}{\bibinfo{person}{Dinghan Shen}, \bibinfo{person}{Asli
  Celikyilmaz}, \bibinfo{person}{Yizhe Zhang}, \bibinfo{person}{Liqun Chen},
  \bibinfo{person}{Xin Wang}, \bibinfo{person}{Jianfeng Gao}, {and}
  \bibinfo{person}{Lawrence Carin}.} \bibinfo{year}{2019}\natexlab{}.
\newblock \showarticletitle{Towards Generating Long and Coherent Text with
  Multi-Level Latent Variable Models}. In \bibinfo{booktitle}{\emph{Proceedings
  of the 57th Annual Meeting of the Association for Computational
  Linguistics}}. \bibinfo{pages}{2079--2089}.
\newblock


\bibitem[\protect\citeauthoryear{Song, Wang, Zhang, Liu, and Liu}{Song
  et~al\mbox{.}}{2020}]%
        {song2020generate}
\bibfield{author}{\bibinfo{person}{Haoyu Song}, \bibinfo{person}{Yan Wang},
  \bibinfo{person}{Wei-Nan Zhang}, \bibinfo{person}{Xiaojiang Liu}, {and}
  \bibinfo{person}{Ting Liu}.} \bibinfo{year}{2020}\natexlab{}.
\newblock \showarticletitle{Generate, Delete and Rewrite: A Three-Stage
  Framework for Improving Persona Consistency of Dialogue Generation}.
\newblock \bibinfo{journal}{\emph{arXiv preprint arXiv:2004.07672}}
  (\bibinfo{year}{2020}).
\newblock


\bibitem[\protect\citeauthoryear{Song, Zeng, Gao, and Shen}{Song
  et~al\mbox{.}}{2018}]%
        {song2018pixels}
\bibfield{author}{\bibinfo{person}{Jingkuan Song}, \bibinfo{person}{Pengpeng
  Zeng}, \bibinfo{person}{Lianli Gao}, {and} \bibinfo{person}{Heng~Tao Shen}.}
  \bibinfo{year}{2018}\natexlab{}.
\newblock \showarticletitle{From Pixels to Objects: Cubic Visual Attention for
  Visual Question Answering.}. In \bibinfo{booktitle}{\emph{IJCAI}}.
  \bibinfo{pages}{906--912}.
\newblock


\bibitem[\protect\citeauthoryear{Song, Tan, Qin, Lu, and Liu}{Song
  et~al\mbox{.}}{2019}]%
        {song2019mass}
\bibfield{author}{\bibinfo{person}{Kaitao Song}, \bibinfo{person}{Xu Tan},
  \bibinfo{person}{Tao Qin}, \bibinfo{person}{Jianfeng Lu}, {and}
  \bibinfo{person}{Tie-Yan Liu}.} \bibinfo{year}{2019}\natexlab{}.
\newblock \showarticletitle{MASS: Masked Sequence to Sequence Pre-training for
  Language Generation}. In \bibinfo{booktitle}{\emph{International Conference
  on Machine Learning}}. \bibinfo{pages}{5926--5936}.
\newblock


\bibitem[\protect\citeauthoryear{Sordoni, Galley, Auli, Brockett, Ji, Mitchell,
  Nie, Gao, and Dolan}{Sordoni et~al\mbox{.}}{2015}]%
        {sordoni2015neural}
\bibfield{author}{\bibinfo{person}{Alessandro Sordoni}, \bibinfo{person}{Michel
  Galley}, \bibinfo{person}{Michael Auli}, \bibinfo{person}{Chris Brockett},
  \bibinfo{person}{Yangfeng Ji}, \bibinfo{person}{Margaret Mitchell},
  \bibinfo{person}{Jian-Yun Nie}, \bibinfo{person}{Jianfeng Gao}, {and}
  \bibinfo{person}{Bill Dolan}.} \bibinfo{year}{2015}\natexlab{}.
\newblock \showarticletitle{A Neural Network Approach to Context-Sensitive
  Generation of Conversational Responses}. In
  \bibinfo{booktitle}{\emph{Proceedings of the 2015 Conference of the North
  American Chapter of the Association for Computational Linguistics: Human
  Language Technologies}}. \bibinfo{pages}{196--205}.
\newblock


\bibitem[\protect\citeauthoryear{Sutskever, Vinyals, and Le}{Sutskever
  et~al\mbox{.}}{2014}]%
        {sutskever2014sequence}
\bibfield{author}{\bibinfo{person}{Ilya Sutskever}, \bibinfo{person}{Oriol
  Vinyals}, {and} \bibinfo{person}{Quoc~V Le}.}
  \bibinfo{year}{2014}\natexlab{}.
\newblock \showarticletitle{Sequence to sequence learning with neural
  networks}. In \bibinfo{booktitle}{\emph{Advances in neural information
  processing systems}}. \bibinfo{pages}{3104--3112}.
\newblock


\bibitem[\protect\citeauthoryear{Tang, Yang, Carton, Zhang, and Mei}{Tang
  et~al\mbox{.}}{2016}]%
        {tang2016context}
\bibfield{author}{\bibinfo{person}{Jian Tang}, \bibinfo{person}{Yifan Yang},
  \bibinfo{person}{Sam Carton}, \bibinfo{person}{Ming Zhang}, {and}
  \bibinfo{person}{Qiaozhu Mei}.} \bibinfo{year}{2016}\natexlab{}.
\newblock \showarticletitle{Context-aware natural language generation with
  recurrent neural networks}.
\newblock \bibinfo{journal}{\emph{arXiv preprint arXiv:1611.09900}}
  (\bibinfo{year}{2016}).
\newblock


\bibitem[\protect\citeauthoryear{Tian, Yan, Mou, Song, Feng, and Zhao}{Tian
  et~al\mbox{.}}{2017}]%
        {tian2017make}
\bibfield{author}{\bibinfo{person}{Zhiliang Tian}, \bibinfo{person}{Rui Yan},
  \bibinfo{person}{Lili Mou}, \bibinfo{person}{Yiping Song},
  \bibinfo{person}{Yansong Feng}, {and} \bibinfo{person}{Dongyan Zhao}.}
  \bibinfo{year}{2017}\natexlab{}.
\newblock \showarticletitle{How to make context more useful? an empirical study
  on context-aware neural conversational models}. In
  \bibinfo{booktitle}{\emph{Proceedings of the 55th Annual Meeting of the
  Association for Computational Linguistics (Volume 2: Short Papers)}}.
  \bibinfo{pages}{231--236}.
\newblock


\bibitem[\protect\citeauthoryear{Van Der~Lee, Gatt, Van~Miltenburg, Wubben, and
  Krahmer}{Van Der~Lee et~al\mbox{.}}{2019}]%
        {van2019best}
\bibfield{author}{\bibinfo{person}{Chris Van Der~Lee}, \bibinfo{person}{Albert
  Gatt}, \bibinfo{person}{Emiel Van~Miltenburg}, \bibinfo{person}{Sander
  Wubben}, {and} \bibinfo{person}{Emiel Krahmer}.}
  \bibinfo{year}{2019}\natexlab{}.
\newblock \showarticletitle{Best practices for the human evaluation of
  automatically generated text}. In \bibinfo{booktitle}{\emph{Proceedings of
  the 12th International Conference on Natural Language Generation}}.
  \bibinfo{pages}{355--368}.
\newblock


\bibitem[\protect\citeauthoryear{Vaswani, Shazeer, Parmar, Uszkoreit, Jones,
  Gomez, Kaiser, and Polosukhin}{Vaswani et~al\mbox{.}}{2017}]%
        {vaswani2017attention}
\bibfield{author}{\bibinfo{person}{Ashish Vaswani}, \bibinfo{person}{Noam
  Shazeer}, \bibinfo{person}{Niki Parmar}, \bibinfo{person}{Jakob Uszkoreit},
  \bibinfo{person}{Llion Jones}, \bibinfo{person}{Aidan~N Gomez},
  \bibinfo{person}{{\L}ukasz Kaiser}, {and} \bibinfo{person}{Illia
  Polosukhin}.} \bibinfo{year}{2017}\natexlab{}.
\newblock \showarticletitle{Attention is all you need}. In
  \bibinfo{booktitle}{\emph{Advances in neural information processing
  systems}}. \bibinfo{pages}{5998--6008}.
\newblock


\bibitem[\protect\citeauthoryear{Vedantam, Lawrence~Zitnick, and
  Parikh}{Vedantam et~al\mbox{.}}{2015}]%
        {vedantam2015cider}
\bibfield{author}{\bibinfo{person}{Ramakrishna Vedantam}, \bibinfo{person}{C
  Lawrence~Zitnick}, {and} \bibinfo{person}{Devi Parikh}.}
  \bibinfo{year}{2015}\natexlab{}.
\newblock \showarticletitle{Cider: Consensus-based image description
  evaluation}. In \bibinfo{booktitle}{\emph{Proceedings of the IEEE conference
  on computer vision and pattern recognition}}. \bibinfo{pages}{4566--4575}.
\newblock


\bibitem[\protect\citeauthoryear{Vijayakumar, Cogswell, Selvaraju, Sun, Lee,
  Crandall, and Batra}{Vijayakumar et~al\mbox{.}}{2016}]%
        {vijayakumar2016diverse}
\bibfield{author}{\bibinfo{person}{Ashwin~K Vijayakumar},
  \bibinfo{person}{Michael Cogswell}, \bibinfo{person}{Ramprasath~R Selvaraju},
  \bibinfo{person}{Qing Sun}, \bibinfo{person}{Stefan Lee},
  \bibinfo{person}{David Crandall}, {and} \bibinfo{person}{Dhruv Batra}.}
  \bibinfo{year}{2016}\natexlab{}.
\newblock \showarticletitle{Diverse beam search: Decoding diverse solutions
  from neural sequence models}.
\newblock \bibinfo{journal}{\emph{arXiv preprint arXiv:1610.02424}}
  (\bibinfo{year}{2016}).
\newblock


\bibitem[\protect\citeauthoryear{Vinyals and Le}{Vinyals and Le}{2015}]%
        {vinyals2015neural}
\bibfield{author}{\bibinfo{person}{Oriol Vinyals} {and} \bibinfo{person}{Quoc
  Le}.} \bibinfo{year}{2015}\natexlab{}.
\newblock \showarticletitle{A neural conversational model}.
\newblock \bibinfo{journal}{\emph{arXiv preprint arXiv:1506.05869}}
  (\bibinfo{year}{2015}).
\newblock


\bibitem[\protect\citeauthoryear{Vinyals, Toshev, Bengio, and Erhan}{Vinyals
  et~al\mbox{.}}{2015}]%
        {vinyals2015show}
\bibfield{author}{\bibinfo{person}{Oriol Vinyals}, \bibinfo{person}{Alexander
  Toshev}, \bibinfo{person}{Samy Bengio}, {and} \bibinfo{person}{Dumitru
  Erhan}.} \bibinfo{year}{2015}\natexlab{}.
\newblock \showarticletitle{Show and tell: A neural image caption generator}.
  In \bibinfo{booktitle}{\emph{Proceedings of the IEEE conference on computer
  vision and pattern recognition}}. \bibinfo{pages}{3156--3164}.
\newblock


\bibitem[\protect\citeauthoryear{Voita, Serdyukov, Sennrich, and Titov}{Voita
  et~al\mbox{.}}{2018}]%
        {voita2018context}
\bibfield{author}{\bibinfo{person}{Elena Voita}, \bibinfo{person}{Pavel
  Serdyukov}, \bibinfo{person}{Rico Sennrich}, {and} \bibinfo{person}{Ivan
  Titov}.} \bibinfo{year}{2018}\natexlab{}.
\newblock \showarticletitle{Context-Aware Neural Machine Translation Learns
  Anaphora Resolution}. In \bibinfo{booktitle}{\emph{Proceedings of the 56th
  Annual Meeting of the Association for Computational Linguistics (Volume 1:
  Long Papers)}}. \bibinfo{pages}{1264--1274}.
\newblock


\bibitem[\protect\citeauthoryear{Wang and Wan}{Wang and Wan}{2018}]%
        {wang2018sentigan}
\bibfield{author}{\bibinfo{person}{Ke Wang} {and} \bibinfo{person}{Xiaojun
  Wan}.} \bibinfo{year}{2018}\natexlab{}.
\newblock \showarticletitle{SentiGAN: Generating Sentimental Texts via Mixture
  Adversarial Networks.}. In \bibinfo{booktitle}{\emph{IJCAI}}.
  \bibinfo{pages}{4446--4452}.
\newblock


\bibitem[\protect\citeauthoryear{Wang, Yao, Tao, Zhong, Liu, and Du}{Wang
  et~al\mbox{.}}{2018}]%
        {wang2018reinforced}
\bibfield{author}{\bibinfo{person}{Li Wang}, \bibinfo{person}{Junlin Yao},
  \bibinfo{person}{Yunzhe Tao}, \bibinfo{person}{Li Zhong},
  \bibinfo{person}{Wei Liu}, {and} \bibinfo{person}{Qiang Du}.}
  \bibinfo{year}{2018}\natexlab{}.
\newblock \showarticletitle{A Reinforced Topic-Aware Convolutional
  Sequence-to-Sequence Model for Abstractive Text Summarization}. In
  \bibinfo{booktitle}{\emph{International Joint Conference on Artificial
  Intelligence}}.
\newblock


\bibitem[\protect\citeauthoryear{Wang, Ling, and Hu}{Wang
  et~al\mbox{.}}{2019b}]%
        {wang2019knowledge}
\bibfield{author}{\bibinfo{person}{Run-Ze Wang}, \bibinfo{person}{Zhen-Hua
  Ling}, {and} \bibinfo{person}{Yu Hu}.} \bibinfo{year}{2019}\natexlab{b}.
\newblock \showarticletitle{Knowledge base question answering with attentive
  pooling for question representation}.
\newblock \bibinfo{journal}{\emph{IEEE Access}}  \bibinfo{volume}{7}
  (\bibinfo{year}{2019}), \bibinfo{pages}{46773--46784}.
\newblock


\bibitem[\protect\citeauthoryear{Wang, Gan, Xu, Zhang, Wang, Shen, Chen, and
  Carin}{Wang et~al\mbox{.}}{2019a}]%
        {wang2019topic}
\bibfield{author}{\bibinfo{person}{Wenlin Wang}, \bibinfo{person}{Zhe Gan},
  \bibinfo{person}{Hongteng Xu}, \bibinfo{person}{Ruiyi Zhang},
  \bibinfo{person}{Guoyin Wang}, \bibinfo{person}{Dinghan Shen},
  \bibinfo{person}{Changyou Chen}, {and} \bibinfo{person}{Lawrence Carin}.}
  \bibinfo{year}{2019}\natexlab{a}.
\newblock \showarticletitle{Topic-Guided Variational Auto-Encoder for Text
  Generation}. In \bibinfo{booktitle}{\emph{Proceedings of the 2019 Conference
  of the North American Chapter of the Association for Computational
  Linguistics: Human Language Technologies, Volume 1 (Long and Short Papers)}}.
  \bibinfo{pages}{166--177}.
\newblock


\bibitem[\protect\citeauthoryear{Wang, Rong, Ouyang, and Xiong}{Wang
  et~al\mbox{.}}{2019c}]%
        {wang2019augmenting}
\bibfield{author}{\bibinfo{person}{Yanmeng Wang}, \bibinfo{person}{Wenge Rong},
  \bibinfo{person}{Yuanxin Ouyang}, {and} \bibinfo{person}{Zhang Xiong}.}
  \bibinfo{year}{2019}\natexlab{c}.
\newblock \showarticletitle{Augmenting Dialogue Response Generation With
  Unstructured Textual Knowledge}.
\newblock \bibinfo{journal}{\emph{IEEE Access}}  \bibinfo{volume}{7}
  (\bibinfo{year}{2019}), \bibinfo{pages}{34954--34963}.
\newblock


\bibitem[\protect\citeauthoryear{Welleck, Kulikov, Roller, Dinan, Cho, and
  Weston}{Welleck et~al\mbox{.}}{2019}]%
        {welleck2019neural}
\bibfield{author}{\bibinfo{person}{Sean Welleck}, \bibinfo{person}{Ilia
  Kulikov}, \bibinfo{person}{Stephen Roller}, \bibinfo{person}{Emily Dinan},
  \bibinfo{person}{Kyunghyun Cho}, {and} \bibinfo{person}{Jason Weston}.}
  \bibinfo{year}{2019}\natexlab{}.
\newblock \showarticletitle{Neural Text Generation With Unlikelihood Training}.
  In \bibinfo{booktitle}{\emph{International Conference on Learning
  Representations}}.
\newblock


\bibitem[\protect\citeauthoryear{Wu, Li, Zhang, Zhou, and Wu}{Wu
  et~al\mbox{.}}{2020}]%
        {wu2020diverse}
\bibfield{author}{\bibinfo{person}{Sixing Wu}, \bibinfo{person}{Ying Li},
  \bibinfo{person}{Dawei Zhang}, \bibinfo{person}{Yang Zhou}, {and}
  \bibinfo{person}{Zhonghai Wu}.} \bibinfo{year}{2020}\natexlab{}.
\newblock \showarticletitle{Diverse and informative dialogue generation with
  context-specific commonsense knowledge awareness}. In
  \bibinfo{booktitle}{\emph{Proceedings of the 58th Annual Meeting of the
  Association for Computational Linguistics}}. \bibinfo{pages}{5811--5820}.
\newblock


\bibitem[\protect\citeauthoryear{Wu, Schuster, Chen, Le, Norouzi, Macherey,
  Krikun, Cao, Gao, Macherey, et~al\mbox{.}}{Wu et~al\mbox{.}}{2016}]%
        {wu2016google}
\bibfield{author}{\bibinfo{person}{Yonghui Wu}, \bibinfo{person}{Mike
  Schuster}, \bibinfo{person}{Zhifeng Chen}, \bibinfo{person}{Quoc~V Le},
  \bibinfo{person}{Mohammad Norouzi}, \bibinfo{person}{Wolfgang Macherey},
  \bibinfo{person}{Maxim Krikun}, \bibinfo{person}{Yuan Cao},
  \bibinfo{person}{Qin Gao}, \bibinfo{person}{Klaus Macherey}, {et~al\mbox{.}}}
  \bibinfo{year}{2016}\natexlab{}.
\newblock \showarticletitle{Google's neural machine translation system:
  Bridging the gap between human and machine translation}.
\newblock \bibinfo{journal}{\emph{arXiv preprint arXiv:1609.08144}}
  (\bibinfo{year}{2016}).
\newblock


\bibitem[\protect\citeauthoryear{Xing, Wu, Wu, Liu, Huang, Zhou, and Ma}{Xing
  et~al\mbox{.}}{2017}]%
        {xing2017topic}
\bibfield{author}{\bibinfo{person}{Chen Xing}, \bibinfo{person}{Wei Wu},
  \bibinfo{person}{Yu Wu}, \bibinfo{person}{Jie Liu}, \bibinfo{person}{Yalou
  Huang}, \bibinfo{person}{Ming Zhou}, {and} \bibinfo{person}{Wei-Ying Ma}.}
  \bibinfo{year}{2017}\natexlab{}.
\newblock \showarticletitle{Topic aware neural response generation}. In
  \bibinfo{booktitle}{\emph{Thirty-First AAAI Conference on Artificial
  Intelligence}}.
\newblock


\bibitem[\protect\citeauthoryear{Xing, Wu, Wu, Huang, and Zhou}{Xing
  et~al\mbox{.}}{2018}]%
        {xing2018hierarchical}
\bibfield{author}{\bibinfo{person}{Chen Xing}, \bibinfo{person}{Yu Wu},
  \bibinfo{person}{Wei Wu}, \bibinfo{person}{Yalou Huang}, {and}
  \bibinfo{person}{Ming Zhou}.} \bibinfo{year}{2018}\natexlab{}.
\newblock \showarticletitle{Hierarchical recurrent attention network for
  response generation}. In \bibinfo{booktitle}{\emph{Thirty-Second AAAI
  Conference on Artificial Intelligence}}.
\newblock


\bibitem[\protect\citeauthoryear{Xu, Ba, Kiros, Cho, Courville, Salakhudinov,
  Zemel, and Bengio}{Xu et~al\mbox{.}}{2015}]%
        {xu2015show}
\bibfield{author}{\bibinfo{person}{Kelvin Xu}, \bibinfo{person}{Jimmy Ba},
  \bibinfo{person}{Ryan Kiros}, \bibinfo{person}{Kyunghyun Cho},
  \bibinfo{person}{Aaron Courville}, \bibinfo{person}{Ruslan Salakhudinov},
  \bibinfo{person}{Rich Zemel}, {and} \bibinfo{person}{Yoshua Bengio}.}
  \bibinfo{year}{2015}\natexlab{}.
\newblock \showarticletitle{Show, attend and tell: Neural image caption
  generation with visual attention}. In \bibinfo{booktitle}{\emph{International
  conference on machine learning}}. \bibinfo{pages}{2048--2057}.
\newblock


\bibitem[\protect\citeauthoryear{Yang, Wang, Liu, Liu, Lyu, Wu, She, and
  Li}{Yang et~al\mbox{.}}{2019b}]%
        {yang2019enhancing}
\bibfield{author}{\bibinfo{person}{An Yang}, \bibinfo{person}{Quan Wang},
  \bibinfo{person}{Jing Liu}, \bibinfo{person}{Kai Liu},
  \bibinfo{person}{Yajuan Lyu}, \bibinfo{person}{Hua Wu},
  \bibinfo{person}{Qiaoqiao She}, {and} \bibinfo{person}{Sujian Li}.}
  \bibinfo{year}{2019}\natexlab{b}.
\newblock \showarticletitle{Enhancing pre-trained language representations with
  rich knowledge for machine reading comprehension}. In
  \bibinfo{booktitle}{\emph{Proceedings of the 57th Annual Meeting of the
  Association for Computational Linguistics}}. \bibinfo{pages}{2346--2357}.
\newblock


\bibitem[\protect\citeauthoryear{Yang, Qu, Lei, Zhu, Zhao, Chen, and
  Huang}{Yang et~al\mbox{.}}{2018}]%
        {yang2018investigating}
\bibfield{author}{\bibinfo{person}{Min Yang}, \bibinfo{person}{Qiang Qu},
  \bibinfo{person}{Kai Lei}, \bibinfo{person}{Jia Zhu}, \bibinfo{person}{Zhou
  Zhao}, \bibinfo{person}{Xiaojun Chen}, {and} \bibinfo{person}{Joshua~Z
  Huang}.} \bibinfo{year}{2018}\natexlab{}.
\newblock \showarticletitle{Investigating deep reinforcement learning
  techniques in personalized dialogue generation}. In
  \bibinfo{booktitle}{\emph{Proceedings of the 2018 SIAM International
  Conference on Data Mining}}. SIAM, \bibinfo{pages}{630--638}.
\newblock


\bibitem[\protect\citeauthoryear{Yang, Zhao, Zhao, Chen, Zhu, Zhou, and
  Cao}{Yang et~al\mbox{.}}{2017}]%
        {yang2017personalized}
\bibfield{author}{\bibinfo{person}{Min Yang}, \bibinfo{person}{Zhou Zhao},
  \bibinfo{person}{Wei Zhao}, \bibinfo{person}{Xiaojun Chen},
  \bibinfo{person}{Jia Zhu}, \bibinfo{person}{Lianqiang Zhou}, {and}
  \bibinfo{person}{Zigang Cao}.} \bibinfo{year}{2017}\natexlab{}.
\newblock \showarticletitle{Personalized response generation via domain
  adaptation}. In \bibinfo{booktitle}{\emph{Proceedings of the 40th
  International ACM SIGIR Conference on Research and Development in Information
  Retrieval}}. ACM, \bibinfo{pages}{1021--1024}.
\newblock


\bibitem[\protect\citeauthoryear{Yang, Dai, Yang, Carbonell, Salakhutdinov, and
  Le}{Yang et~al\mbox{.}}{2019a}]%
        {yang2019xlnet}
\bibfield{author}{\bibinfo{person}{Zhilin Yang}, \bibinfo{person}{Zihang Dai},
  \bibinfo{person}{Yiming Yang}, \bibinfo{person}{Jaime Carbonell},
  \bibinfo{person}{Russ~R Salakhutdinov}, {and} \bibinfo{person}{Quoc~V Le}.}
  \bibinfo{year}{2019}\natexlab{a}.
\newblock \showarticletitle{Xlnet: Generalized autoregressive pretraining for
  language understanding}. In \bibinfo{booktitle}{\emph{Advances in neural
  information processing systems}}. \bibinfo{pages}{5754--5764}.
\newblock


\bibitem[\protect\citeauthoryear{Young, Cambria, Chaturvedi, Zhou, Biswas, and
  Huang}{Young et~al\mbox{.}}{2018}]%
        {young2018augmenting}
\bibfield{author}{\bibinfo{person}{Tom Young}, \bibinfo{person}{Erik Cambria},
  \bibinfo{person}{Iti Chaturvedi}, \bibinfo{person}{Hao Zhou},
  \bibinfo{person}{Subham Biswas}, {and} \bibinfo{person}{Minlie Huang}.}
  \bibinfo{year}{2018}\natexlab{}.
\newblock \showarticletitle{Augmenting end-to-end dialogue systems with
  commonsense knowledge}. In \bibinfo{booktitle}{\emph{Thirty-Second AAAI
  Conference on Artificial Intelligence}}.
\newblock


\bibitem[\protect\citeauthoryear{Yu, Zhang, Wang, and Yu}{Yu
  et~al\mbox{.}}{2017}]%
        {yu2017seqgan}
\bibfield{author}{\bibinfo{person}{Lantao Yu}, \bibinfo{person}{Weinan Zhang},
  \bibinfo{person}{Jun Wang}, {and} \bibinfo{person}{Yong Yu}.}
  \bibinfo{year}{2017}\natexlab{}.
\newblock \showarticletitle{Seqgan: Sequence generative adversarial nets with
  policy gradient}. In \bibinfo{booktitle}{\emph{Thirty-First AAAI Conference
  on Artificial Intelligence}}.
\newblock


\bibitem[\protect\citeauthoryear{Yuan and Huang}{Yuan and Huang}{2019}]%
        {yuan2019personalized}
\bibfield{author}{\bibinfo{person}{Chenhan Yuan} {and} \bibinfo{person}{Yi-Chin
  Huang}.} \bibinfo{year}{2019}\natexlab{}.
\newblock \showarticletitle{Personalized sentence generation using generative
  adversarial networks with author-specific word usage}.
\newblock \bibinfo{journal}{\emph{arXiv preprint arXiv:1904.09442}}
  (\bibinfo{year}{2019}).
\newblock


\bibitem[\protect\citeauthoryear{Zhang, Chan, Song, Zhao, and Yan}{Zhang
  et~al\mbox{.}}{2018a}]%
        {zhang2018less}
\bibfield{author}{\bibinfo{person}{Haisong Zhang}, \bibinfo{person}{Zhangming
  Chan}, \bibinfo{person}{Yan Song}, \bibinfo{person}{Dongyan Zhao}, {and}
  \bibinfo{person}{Rui Yan}.} \bibinfo{year}{2018}\natexlab{a}.
\newblock \showarticletitle{When Less Is More: Using Less Context Information
  to Generate Better Utterances in Group Conversations}. In
  \bibinfo{booktitle}{\emph{CCF International Conference on Natural Language
  Processing and Chinese Computing}}. Springer, \bibinfo{pages}{76--84}.
\newblock


\bibitem[\protect\citeauthoryear{Zhang, Dinan, Urbanek, Szlam, Kiela, and
  Weston}{Zhang et~al\mbox{.}}{2018b}]%
        {zhang2018personalizing}
\bibfield{author}{\bibinfo{person}{Saizheng Zhang}, \bibinfo{person}{Emily
  Dinan}, \bibinfo{person}{Jack Urbanek}, \bibinfo{person}{Arthur Szlam},
  \bibinfo{person}{Douwe Kiela}, {and} \bibinfo{person}{Jason Weston}.}
  \bibinfo{year}{2018}\natexlab{b}.
\newblock \showarticletitle{Personalizing Dialogue Agents: I have a dog, do you
  have pets too?}. In \bibinfo{booktitle}{\emph{Proceedings of the 56th Annual
  Meeting of the Association for Computational Linguistics (Volume 1: Long
  Papers)}}. \bibinfo{pages}{2204--2213}.
\newblock


\bibitem[\protect\citeauthoryear{Zhang, Zhu, Wang, Zhao, and Liu}{Zhang
  et~al\mbox{.}}{2019b}]%
        {zhang2019neural}
\bibfield{author}{\bibinfo{person}{Wei-Nan Zhang}, \bibinfo{person}{Qingfu
  Zhu}, \bibinfo{person}{Yifa Wang}, \bibinfo{person}{Yanyan Zhao}, {and}
  \bibinfo{person}{Ting Liu}.} \bibinfo{year}{2019}\natexlab{b}.
\newblock \showarticletitle{Neural personalized response generation as domain
  adaptation}.
\newblock \bibinfo{journal}{\emph{World Wide Web}} \bibinfo{volume}{22},
  \bibinfo{number}{4} (\bibinfo{year}{2019}), \bibinfo{pages}{1427--1446}.
\newblock


\bibitem[\protect\citeauthoryear{Zhang, Gan, and Carin}{Zhang
  et~al\mbox{.}}{2016}]%
        {zhang2016generating}
\bibfield{author}{\bibinfo{person}{Yizhe Zhang}, \bibinfo{person}{Zhe Gan},
  {and} \bibinfo{person}{Lawrence Carin}.} \bibinfo{year}{2016}\natexlab{}.
\newblock \showarticletitle{Generating text via adversarial training}. In
  \bibinfo{booktitle}{\emph{NIPS workshop on Adversarial Training}},
  Vol.~\bibinfo{volume}{21}.
\newblock


\bibitem[\protect\citeauthoryear{Zhang, Sun, Galley, Chen, Brockett, Gao, Gao,
  Liu, and Dolan}{Zhang et~al\mbox{.}}{2019a}]%
        {zhang2019dialogpt}
\bibfield{author}{\bibinfo{person}{Yizhe Zhang}, \bibinfo{person}{Siqi Sun},
  \bibinfo{person}{Michel Galley}, \bibinfo{person}{Yen-Chun Chen},
  \bibinfo{person}{Chris Brockett}, \bibinfo{person}{Xiang Gao},
  \bibinfo{person}{Jianfeng Gao}, \bibinfo{person}{Jingjing Liu}, {and}
  \bibinfo{person}{Bill Dolan}.} \bibinfo{year}{2019}\natexlab{a}.
\newblock \showarticletitle{Dialogpt: Large-scale generative pre-training for
  conversational response generation}.
\newblock \bibinfo{journal}{\emph{arXiv preprint arXiv:1911.00536}}
  (\bibinfo{year}{2019}).
\newblock


\bibitem[\protect\citeauthoryear{Zhang, Wang, Li, Gan, Brockett, and
  Dolan}{Zhang et~al\mbox{.}}{2020}]%
        {zhang2020pointer}
\bibfield{author}{\bibinfo{person}{Yizhe Zhang}, \bibinfo{person}{Guoyin Wang},
  \bibinfo{person}{Chunyuan Li}, \bibinfo{person}{Zhe Gan},
  \bibinfo{person}{Chris Brockett}, {and} \bibinfo{person}{Bill Dolan}.}
  \bibinfo{year}{2020}\natexlab{}.
\newblock \showarticletitle{POINTER: Constrained Text Generation via
  Insertion-based Generative Pre-training}.
\newblock \bibinfo{journal}{\emph{arXiv preprint arXiv:2005.00558}}
  (\bibinfo{year}{2020}).
\newblock


\bibitem[\protect\citeauthoryear{Zhao, Tao, Wu, Xu, Zhao, and Yan}{Zhao
  et~al\mbox{.}}{2019}]%
        {zhao2019document}
\bibfield{author}{\bibinfo{person}{Xueliang Zhao}, \bibinfo{person}{Chongyang
  Tao}, \bibinfo{person}{Wei Wu}, \bibinfo{person}{Can Xu},
  \bibinfo{person}{Dongyan Zhao}, {and} \bibinfo{person}{Rui Yan}.}
  \bibinfo{year}{2019}\natexlab{}.
\newblock \showarticletitle{A document-grounded matching network for response
  selection in retrieval-based chatbots}. In
  \bibinfo{booktitle}{\emph{Proceedings of the 28th International Joint
  Conference on Artificial Intelligence}}. AAAI Press,
  \bibinfo{pages}{5443--5449}.
\newblock


\bibitem[\protect\citeauthoryear{Zhao, Wu, Tao, Xu, Zhao, and Yan}{Zhao
  et~al\mbox{.}}{2020a}]%
        {Zhao2020Low-Resource}
\bibfield{author}{\bibinfo{person}{Xueliang Zhao}, \bibinfo{person}{Wei Wu},
  \bibinfo{person}{Chongyang Tao}, \bibinfo{person}{Can Xu},
  \bibinfo{person}{Dongyan Zhao}, {and} \bibinfo{person}{Rui Yan}.}
  \bibinfo{year}{2020}\natexlab{a}.
\newblock \showarticletitle{Low-Resource Knowledge-Grounded Dialogue
  Generation}. In \bibinfo{booktitle}{\emph{International Conference on
  Learning Representations}}.
\newblock
\urldef\tempurl%
\url{https://openreview.net/forum?id=rJeIcTNtvS}
\showURL{%
\tempurl}


\bibitem[\protect\citeauthoryear{Zhao, Wu, Xu, Tao, Zhao, and Yan}{Zhao
  et~al\mbox{.}}{2020b}]%
        {zhao2020knowledge}
\bibfield{author}{\bibinfo{person}{Xueliang Zhao}, \bibinfo{person}{Wei Wu},
  \bibinfo{person}{Can Xu}, \bibinfo{person}{Chongyang Tao},
  \bibinfo{person}{Dongyan Zhao}, {and} \bibinfo{person}{Rui Yan}.}
  \bibinfo{year}{2020}\natexlab{b}.
\newblock \showarticletitle{Knowledge-Grounded Dialogue Generation with
  Pre-trained Language Models}. In \bibinfo{booktitle}{\emph{Proceedings of the
  2020 Conference on Empirical Methods in Natural Language Processing
  (EMNLP)}}. \bibinfo{pages}{3377--3390}.
\newblock


\bibitem[\protect\citeauthoryear{Zheng, Chen, Huang, Liu, and Zhu}{Zheng
  et~al\mbox{.}}{2019}]%
        {zheng2019personalized}
\bibfield{author}{\bibinfo{person}{Yinhe Zheng}, \bibinfo{person}{Guanyi Chen},
  \bibinfo{person}{Minlie Huang}, \bibinfo{person}{Song Liu}, {and}
  \bibinfo{person}{Xuan Zhu}.} \bibinfo{year}{2019}\natexlab{}.
\newblock \showarticletitle{Personalized dialogue generation with diversified
  traits}.
\newblock \bibinfo{journal}{\emph{arXiv preprint arXiv:1901.09672}}
  (\bibinfo{year}{2019}).
\newblock


\bibitem[\protect\citeauthoryear{Zhong, Wang, and Miao}{Zhong
  et~al\mbox{.}}{2019}]%
        {zhong2019knowledge}
\bibfield{author}{\bibinfo{person}{Peixiang Zhong}, \bibinfo{person}{Di Wang},
  {and} \bibinfo{person}{Chunyan Miao}.} \bibinfo{year}{2019}\natexlab{}.
\newblock \showarticletitle{Knowledge-Enriched Transformer for Emotion
  Detection in Textual Conversations}. In \bibinfo{booktitle}{\emph{Proceedings
  of the 2019 Conference on Empirical Methods in Natural Language Processing
  and the 9th International Joint Conference on Natural Language Processing
  (EMNLP-IJCNLP)}}. \bibinfo{pages}{165--176}.
\newblock


\bibitem[\protect\citeauthoryear{Zhong, Zhang, Wang, Liu, and Miao}{Zhong
  et~al\mbox{.}}{2020}]%
        {zhong2020towards}
\bibfield{author}{\bibinfo{person}{Peixiang Zhong}, \bibinfo{person}{Chen
  Zhang}, \bibinfo{person}{Hao Wang}, \bibinfo{person}{Yong Liu}, {and}
  \bibinfo{person}{Chunyan Miao}.} \bibinfo{year}{2020}\natexlab{}.
\newblock \showarticletitle{Towards Persona-Based Empathetic Conversational
  Models}. In \bibinfo{booktitle}{\emph{Proceedings of the 2020 Conference on
  Empirical Methods in Natural Language Processing (EMNLP)}}.
  \bibinfo{pages}{6556--6566}.
\newblock


\bibitem[\protect\citeauthoryear{Zhou, Huang, Zhang, Zhu, and Liu}{Zhou
  et~al\mbox{.}}{2018a}]%
        {zhou2018emotional}
\bibfield{author}{\bibinfo{person}{Hao Zhou}, \bibinfo{person}{Minlie Huang},
  \bibinfo{person}{Tianyang Zhang}, \bibinfo{person}{Xiaoyan Zhu}, {and}
  \bibinfo{person}{Bing Liu}.} \bibinfo{year}{2018}\natexlab{a}.
\newblock \showarticletitle{Emotional chatting machine: Emotional conversation
  generation with internal and external memory}. In
  \bibinfo{booktitle}{\emph{Thirty-Second AAAI Conference on Artificial
  Intelligence}}.
\newblock


\bibitem[\protect\citeauthoryear{Zhou, Young, Huang, Zhao, Xu, and Zhu}{Zhou
  et~al\mbox{.}}{2018c}]%
        {zhou2018commonsense}
\bibfield{author}{\bibinfo{person}{Hao Zhou}, \bibinfo{person}{Tom Young},
  \bibinfo{person}{Minlie Huang}, \bibinfo{person}{Haizhou Zhao},
  \bibinfo{person}{Jingfang Xu}, {and} \bibinfo{person}{Xiaoyan Zhu}.}
  \bibinfo{year}{2018}\natexlab{c}.
\newblock \showarticletitle{Commonsense Knowledge Aware Conversation Generation
  with Graph Attention.}. In \bibinfo{booktitle}{\emph{IJCAI}}.
  \bibinfo{pages}{4623--4629}.
\newblock


\bibitem[\protect\citeauthoryear{Zhou, Prabhumoye, and Black}{Zhou
  et~al\mbox{.}}{2018b}]%
        {zhou2018dataset}
\bibfield{author}{\bibinfo{person}{Kangyan Zhou}, \bibinfo{person}{Shrimai
  Prabhumoye}, {and} \bibinfo{person}{Alan~W Black}.}
  \bibinfo{year}{2018}\natexlab{b}.
\newblock \showarticletitle{A Dataset for Document Grounded Conversations}. In
  \bibinfo{booktitle}{\emph{Proceedings of the 2018 Conference on Empirical
  Methods in Natural Language Processing}}. \bibinfo{pages}{708--713}.
\newblock


\bibitem[\protect\citeauthoryear{Zhou, Gao, Li, and Shum}{Zhou
  et~al\mbox{.}}{[n.d.]}]%
        {zhou2018design}
\bibfield{author}{\bibinfo{person}{Li Zhou}, \bibinfo{person}{Jianfeng Gao},
  \bibinfo{person}{Di Li}, {and} \bibinfo{person}{Heung-Yeung Shum}.}
  \bibinfo{year}{[n.d.]}\natexlab{}.
\newblock \showarticletitle{The design and implementation of XiaoIce, an
  empathetic social chatbot}.
\newblock \bibinfo{journal}{\emph{Computational Linguistics}}
  \bibinfo{number}{Just Accepted} (\bibinfo{year}{[n.\,d.]}),
  \bibinfo{pages}{1--62}.
\newblock


\bibitem[\protect\citeauthoryear{Zhu, Groth, Bernstein, and Fei-Fei}{Zhu
  et~al\mbox{.}}{2016}]%
        {zhu2016visual7w}
\bibfield{author}{\bibinfo{person}{Yuke Zhu}, \bibinfo{person}{Oliver Groth},
  \bibinfo{person}{Michael Bernstein}, {and} \bibinfo{person}{Li Fei-Fei}.}
  \bibinfo{year}{2016}\natexlab{}.
\newblock \showarticletitle{Visual7w: Grounded question answering in images}.
  In \bibinfo{booktitle}{\emph{Proceedings of the IEEE conference on computer
  vision and pattern recognition}}. \bibinfo{pages}{4995--5004}.
\newblock


\bibitem[\protect\citeauthoryear{Ziegler, Melas-Kyriazi, Gehrmann, and
  Rush}{Ziegler et~al\mbox{.}}{2019}]%
        {ziegler2019encoder}
\bibfield{author}{\bibinfo{person}{Zachary~M Ziegler}, \bibinfo{person}{Luke
  Melas-Kyriazi}, \bibinfo{person}{Sebastian Gehrmann}, {and}
  \bibinfo{person}{Alexander~M Rush}.} \bibinfo{year}{2019}\natexlab{}.
\newblock \showarticletitle{Encoder-agnostic adaptation for conditional
  language generation}.
\newblock \bibinfo{journal}{\emph{arXiv preprint arXiv:1908.06938}}
  (\bibinfo{year}{2019}).
\newblock


\end{thebibliography}

\end{document}